\documentclass[11pt]{article}

\usepackage[final]{acl}

\usepackage{times}
\usepackage{latexsym}
\usepackage{booktabs}

\usepackage[T1]{fontenc}

\usepackage[utf8]{inputenc}

\usepackage{microtype}

\usepackage{inconsolata}

\usepackage{graphicx}

\usepackage{amsmath}
\usepackage{multirow}
\usepackage{rotating}
\usepackage{comment}
\usepackage{fancyvrb}
\usepackage{fvextra}
\usepackage{amssymb}

\title{Understanding LLM Performance Degradation in Multi-Instance Processing: The Roles of Instance Count and Context Length}

\author{Jingxuan Chen \quad \quad Mohammad Taher Pilehvar \quad \quad Jose Camacho-Collados  \\
        School of Computer Science and Informatics, Cardiff University \\
        \texttt{\{ChenJ192,PilehvarMT,CamachoColladosJ\}@cardiff.ac.uk}}

\newcommand{\jose}[1]{{\color{black}#1}}
\newcommand{\feedback}[1]{{\color{black}#1}}
\newcommand{\camera}[1]{{\color{black}#1}}
\newcommand{\update}[1]{{\color{black}#1}}

\begin{document}
\maketitle
\begin{abstract}
Users often rely on Large Language Models (LLMs) for processing \update{multiple} documents or performing analysis over a number of instances. For \camera{example}, analysing the overall sentiment of a number of movie reviews requires an LLM to process the sentiment of each review individually in order to provide a final aggregated answer. 
While LLM performance on such individual tasks is generally high, there has been little research on how LLMs perform when dealing with multi-instance inputs. In this paper, we perform a
\update{comprehensive} evaluation of the \jose{multi-instance processing (MIP)} ability of LLMs for tasks in which they excel individually. The results show that \update{all} LLMs follow a pattern of slight performance degradation for small numbers of instances ($\approx$20–100), followed by a performance collapse on larger instance counts. Crucially, our analysis shows that while context length is 
\update{associated with}
this degradation, the number of instances has a stronger effect on the final results. This finding suggests that when optimising LLM performance for \jose{MIP}, attention should be paid to both context length and, in particular, instance count.\footnote{\jose{Data} and code are available at 
\update{\url{https://github.com/jingxuanchen916/mip-understanding}}.}

\end{abstract}

\section{Introduction}
LLMs have demonstrated remarkable capabilities across a wide range of \update{natural language processing} tasks and beyond~\citep{wang2024survey,zhao2023survey}. However, these capabilities have been predominantly evaluated in settings where a single instance is provided to the model at a time, which we refer to as single-instance processing (SIP). In contrast, many real-world applications, such as data analytics, document analysis, and large-scale information processing, require multi-instance processing (MIP), where the model generates individual predictions for multiple instances and subsequently aggregates them into a single, cohesive final prediction. \jose{The ability of LLMs to process multiple instances has been studied in recent literature in different contexts such as data \feedback{analytics}}~\citep{chen2025large,rahman2025llm,sun2025survey} and long\feedback{-}context settings~\citep{bertsch2025oolong,liu2025comprehensive,qiu-etal-2024-clongeval,shaham-etal-2023-zeroscrolls,wolfson2026monaco}, \jose{which we unify and formalise under the MIP umbrella in this work.}  

Compared to SIP, MIP poses additional challenges due to its long-context nature and the need to perform repeated reasoning and aggregation over multiple instances, making it substantially more demanding for current LLMs~\citep{bertsch2025oolong}. 
Ensuring reliable performance in such settings therefore requires a careful understanding of model failure modes, in order to inform the development of effective mitigation strategies for MIP.

\begin{figure}[t!]
  \includegraphics[width=\columnwidth]{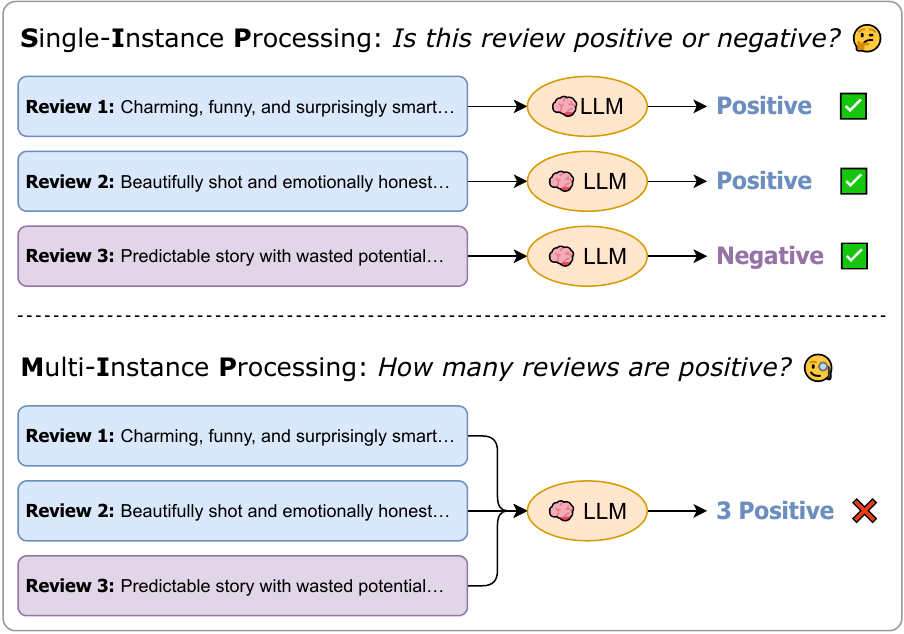}
  \caption{A toy example of SIP and MIP settings for sentiment analysis, where an LLM succeeds under SIP but fails under MIP given the same instances.}
  \label{fig:intro}
\end{figure}

As a motivating example, consider a non-expert user who inputs multiple movie reviews and wishes to \camera{determine how many of them are positive.}
Figure~\ref{fig:intro} illustrates a toy comparison between SIP and MIP: while an LLM can correctly classify the sentiment of each review in isolation, it may fail when required to process and aggregate all instances within a single prompt. 
Given that the model is capable of accurately handling individual instances, it is 
crucial to understand the nature of the errors that arise when multiple instances must be processed jointly. Although alternative solutions, such as agentic designs that process instances separately or require users to manually batch instances or write code, are possible, these approaches are often impractical for non-expert users in real-world settings.

Existing work has extensively examined the challenges posed by long-context inputs, showing that model performance often degrades as context length increases, even when inputs remain within the model's nominal context window~\citep{andoes2025,liu2024lost,moon2025needlechain}. However, in most long-context benchmarks, increasing input length is accompanied by a simultaneous increase in task complexity. This coupling makes it difficult to disentangle whether observed performance degradation arises from longer inputs per se or from increased semantic and reasoning demands.
In parallel, multi-instance or batch processing settings have been explored in prior work, but primarily from an efficiency or cost-reduction perspective, and typically with a limited number of instances~\citep{cheng-etal-2023-batch,linbatchprompt2024}. As a result, the effect of scaling the number of instances itself on model performance remains underexplored.

Given these gaps, we propose two research questions (RQs) to better understand LLM behaviour in MIP settings: %

\vspace*{-1.2mm}
\begin{enumerate}
    \item How does LLM performance change in MIP settings, and what failure behaviours emerge \jose{as the number of instances increases}?
    \vspace*{-1.2mm}
    \item 
    \jose{What are the primary drivers of} performance degradation in MIP? \jose{Do} 
    \update{instance count }\jose{and} context length \jose{have similar effects}?
\end{enumerate}
\vspace*{-1.2mm}

To answer these RQs, we evaluate 
\update{sixteen}
LLMs across a diverse set of eight tasks, including calculation, token-level and sentence-level classification, for which they are known to perform well when only individual instances are provided. 
We analyse a broad range of open-weight and closed-source LLMs, and systematically examine the limits and factors that influence their performance in MIP.
Our results show that model performance typically degrades gradually for small instance counts before collapsing at larger scales, and that this degradation is
\camera{more strongly associated with} the number of instances than with context length alone, even when the latter is substantially increased.

\section{Related Work}
\label{sec:related_work}

The practical demand for MIP is typically driven by the use of LLM-based agents for complex data analytics, such as data wrangling, exploratory analysis, and automated machine learning workflows~\citep{guo2024ds,hong-etal-2025-data,nam2025ds}. 
In these workflows, data processing is a foundational step and often requires transforming heterogeneous inputs into intermediate representations or executable code~\citep{lin2024towards,shankar2024docetl}. 
While current approaches typically adopt modular pipelines to improve reliability through task decomposition~\citep{nam2025ds,shankar2024docetl}, these systems are ultimately bounded by the LLM's ability to handle high-density information. 
Understanding their limits therefore requires examining LLM performance along two dimensions: the capacity to maintain coherence over long contexts and the ability to 
\update{process multiple inputs efficiently}
through batch processing.

\paragraph{Long Context.} 
Recent benchmarks have examined LLM performance degradation in long-context settings, spanning retrieval-based evaluations~\citep{hsiehruler2024,yangcontrollable}, long-context reasoning over dispersed evidence~\citep{kuratov2024babilong,vodrahalli2024michelangelo}, and broader multitask suites~\citep{bai2024longbench,bai2025longbench,chen2026longbench,yen2025helmet,zhang2024infty}.  
Complementary work also studies specialised regimes, including long procedural generation~\citep{ye2025longproc}, scalable mathematical reasoning~\citep{zhou2025gsm}, narrative understanding~\citep{hamilton2025too}, and long-term conversational memory~\citep{wulongmemeval}. 
Prior work has also isolated specific factors, such as input length~\citep{levy2024same} and document count~\citep{levy-etal-2025-documents}. Closest to our work,~\citet{bertsch2025oolong} evaluate long-context aggregation via many in-context instances but primarily vary context length rather than isolating instance count.

\paragraph{Batch Processing.} 
Recent works study whether LLMs can answer multiple questions within a single prompt, typically motivated by reducing inference cost and balancing capacity limits~\citep{cheng-etal-2023-batch,ji2025optimized,linbatchprompt2024,son-etal-2024-multi-task}. 
These studies consistently observe that only a small number of instances can be processed reliably before accuracy degrades. 
However, they do not investigate scaling behaviour beyond this regime, which arises in many practical settings where non-expert users may directly pass all instances to an LLM.
Moreover, their evaluations primarily focus on classification, reasoning, or heterogeneous multitask problems~\citep{wang-etal-2025-exploring,wang2025evaluating,gozzi2024comparative}, rather than controlled aggregation settings where the instance count itself is the primary source of difficulty.

\section{Multi-Instance Processing}
We study MIP, where an LLM is required to reason over multiple input instances within a single prompt. In contrast to retrieval-augmented generation (RAG), which typically uses only a subset of retrieved inputs to produce the final answer, MIP requires processing all provided instances. As shown in Figure~\ref{fig:intro}, LLMs must iterate over all instances individually in MIP to produce intermediate results, which are then aggregated into a final answer.

An instance is defined as a single data entry $x$ (e.g., a movie review, a user post, a sentence, or a number). Let $\mathcal{X}$ denote the set of all available instances. 
The model $\feedback{M}$ is provided with a subset $\mathcal{X}' \subseteq \mathcal{X}$\update{, where the}
number of instances $n = |\mathcal{X}'|$ may vary across inputs. SIP is treated as a special case of MIP where $n = 1$.

\subsection{Formulation}
\label{sec:formulation}

Given a task instruction prompt $\update{\tau}$ and an input instance set
$\mathcal{X}' = \{x_1, \dots, x_n\} \subseteq \mathcal{X}$, the model \feedback{$M$}\feedback{, parameterised by $\theta$}\update{,}
generates an output
\[
o \sim p_\theta(\cdot \mid \update{\tau}, \mathcal{X}')\feedback{,}
\]
\feedback{where $p_\theta$ is the model's conditional distribution.}

A valid model output typically contains an aggregated prediction
$y^{\mathrm{agg}}$ together with a natural language explanation $r$. Let
$y^{\mathrm{agg}\star}$ denote the corresponding ground-truth aggregated label. An output is considered correct when the aggregated prediction matches the ground truth, i.e., $y^{\mathrm{agg}} = y^{\mathrm{agg}\star}$. Otherwise, it is considered a wrong output and belongs to the set $\mathcal{W}$. 
The model may also produce an invalid output, in which case it fails to generate a well-formed prediction. Such outputs belong to the set $\mathcal{I}$. We define both wrong answers (i.e., valid but incorrect outputs) and invalid outputs as failures:
\[
\mathcal{E}_{\text{fail}} = \mathcal{W} \cup \mathcal{I}.
\]
We write $o \in \mathcal{E}_{\text{fail}}$ to denote that a model output is a failure.

\subsection{Filtering for Controlled Difficulty}
\label{sec:filtering}
An important aspect of our methodology is ensuring that the evaluated tasks are simple enough for LLMs to solve when individual instances are provided (i.e., in SIP \jose{settings}). Therefore, to control instance-level task difficulty when evaluating MIP, we construct inputs based on SIP outcomes. 

Let $\mathcal{X}_{\mathrm{SIP}} \subseteq \mathcal{X}$ denote the subset of instances for which all \jose{comparison} models \update{can} produce \update{the} correct prediction under the SIP setting. MIP inputs are then formed by uniformly sampling subsets $\mathcal{X}' \subseteq \mathcal{X}_{\mathrm{SIP}}$ using fixed random seeds. 
To further ensure reliable evaluation, we retain only models whose average SIP task success rate exceeds 95\% and whose per-task SIP success rate exceeds 90\%. We also keep only tasks for which agreement among all retained models exceeds 85\%, measured as the proportion of instances that all \update{comparison} models answer correctly prior to filtering. 

\update{Our filtering procedure ensures that failures observed in the MIP setting are not attributable to intrinsic instance difficulty or ambiguity, but instead reflect the model's ability to reason over and aggregate multiple instances, which is the primary focus of this work.}

\subsection{Evaluation Metrics}

We define an experiment as a specific evaluation configuration represented as a tuple $e = (\feedback{M}, \tau, \mathcal{X}')$, 
\update{which} produces a model output $o_e$.
We define accuracy as a binary metric\feedback{. Let $y^{\mathrm{agg}}\feedback{_e}$ denote the aggregated prediction extracted from the model output $o_e$. Then}:
\[
\mathrm{Acc}(e) =
\begin{cases}
1, & \text{if } y^{\mathrm{agg}}\feedback{_e} = y^{\mathrm{agg}\star}\feedback{_e}, \\
0, & \text{otherwise}.
\end{cases}
\]

Let $\mathcal{D}$ denote the set of all evaluated experiments. The success rate (SR) is defined as the average accuracy across experiments:
\[
\mathrm{SR} = \frac{1}{|\mathcal{D}|} \sum_{e \in \mathcal{D}} \mathrm{Acc}(e).
\]

The invalid rate (IR) measures the fraction of experiments in which the model produces an invalid output:
\[
\mathrm{IR} = \frac{|\{\, e \in \mathcal{D} \mid o_e \in \mathcal{I} \,\}|}{|\mathcal{D}|}.
\]

\section{Experimental Setting}

In this section, we describe our general experimental setting. 

\subsection{Individual Tasks}
\label{sub:tasks}
We consider eight heterogeneous tasks\footnote{We removed three additional tasks whose SIP performance fell below our requirements.} for our analysis, as summarised in Table~\ref{tab:task-overview}. Each task is chosen such that it can be solved individually in the SIP setting by standard LLMs. When multiple instances are provided in the MIP setting, the model is tasked with additionally aggregating outputs across all instances (e.g., counting how many movie reviews are classified as positive in sentiment analysis). Detailed task descriptions and examples are provided in Appendix~\ref{appendix:data}.

\begin{table}
  \centering
   \setlength{\tabcolsep}{12pt}
  \scalebox{0.9}{
    \begin{tabular}{ll}
      \toprule
      \textbf{Name} &
      \textbf{Task} \\
      \midrule
      \emph{Arithmetic} &
      \begin{tabular}[c]{@{}l@{}}
        Solve arithmetic problems \\
        ~~~~~\& Sum of answers
      \end{tabular} \\
      \emph{Category} &
      \begin{tabular}[c]{@{}l@{}}
        Classify news category \\
        ~~~~~\& Aggregate class counts
      \end{tabular} \\
      \emph{Language} &
      \begin{tabular}[c]{@{}l@{}}
        Identify language \\
        ~~~~~\& Aggregate class counts
      \end{tabular} \\
      \emph{NER} &
      \begin{tabular}[c]{@{}l@{}}
        Count ``person'' entities \\
        ~~~~~\& Aggregate total counts
      \end{tabular} \\
      \emph{Parity} &
      \begin{tabular}[c]{@{}l@{}}
        Detect odd or even number \\
        ~~~~~\& Aggregate counts
      \end{tabular} \\
      \emph{Sentiment} &
      \begin{tabular}[c]{@{}l@{}}
        Detect sentiment polarity \\
        ~~~~~\& Aggregate counts
      \end{tabular} \\
      \emph{Word} &
      \begin{tabular}[c]{@{}l@{}}
        Count target word ``women'' \\
        ~~~~~\& Aggregate total counts
      \end{tabular} \\
      \emph{WSD} &
      \begin{tabular}[c]{@{}l@{}}
        Identify ``apple'' word sense \\
        ~~~~~\& Aggregate counts
      \end{tabular} \\
      \bottomrule
    \end{tabular}
  }
  \caption{Overview of the selected tasks and their aggregation logic in the MIP setting. \update{More details can be found in Appendix~\ref{appendix:data}.}}
  \label{tab:task-overview}
\end{table}

\subsection{Models and Prompting}
\label{sub:models}
We \update{use OpenRouter\footnote{\url{https://openrouter.ai}} to} evaluate 
\update{sixteen} LLMs,
including \camera{nine} open-weight models (\camera{\emph{DeepSeek R1}, }\emph{DeepSeek V3}, \emph{gpt-oss-120b}, \emph{gpt-oss-20b}, \emph{Llama 3.3}, \emph{Llama 4 Maverick}, \camera{\emph{MiniMax M2.5}, }\emph{Qwen3-Instruct} \camera{and \emph{Qwen3-Thinking}}) and \update{seven} closed-source models (\update{\emph{Claude Sonnet 4.6},} \emph{Gemini 2.5 Flash}, \update{\emph{Gemini 3.1 Pro\footnote{We used the preview version, which was the only version available at the time of experimentation (March 2026).}},} \update{\emph{GPT-5},} \emph{GPT-5 Nano}, \update{\emph{Grok 4}} and \emph{Grok 4 Fast}).\footnote{As with the task filtering, we removed two open-weight LLMs\update{,} \feedback{\emph{Llama 4 Scout}
and \emph{Mistral NeMo},} whose SIP performance did not meet our criteria.}

For prompting, we use a temperature of 0 and a maximum output length of 20K tokens for consistency across models. To allow limited tolerance to formatting errors, we permit up to three retries when a model produces an invalid output belonging to $\mathcal{I}$. The full set of prompting templates is provided in Appendix~\ref{appendix:prompt}.

\subsection{Single-Instance Filtering}
\label{sub:filtering}
As described in Section~\ref{sec:filtering}, we ensure that each instance can be successfully solved in the SIP setting. To this end, we conduct SIP experiments on 2{,}500 instances for each task.\footnote{The original dataset of \emph{Category} contains fewer than 2{,}500 instances, and each instance is substantially longer. We therefore use 250 instances instead of 2{,}500 for this task. Correspondingly, the maximum MIP sample size for \emph{Category} is also ten times smaller than for the other tasks.} We report each LLM's SIP performance for each task in Appendix~\ref{appendix:single_indi}.
\jose{Moreover,} Table~\ref{tab:single-filtering} in Appendix~\ref{appendix:single_final} reports the percentage of instances retained (i.e., agreement) for each task (from 89\% to nearly 100\%), and the corresponding maximum and minimum SIP success rates across models, all exceeding 93\%. \jose{Finally, as 
\update{described in} Section \ref{sec:filtering},} this filtering retains only instances \jose{
\update{for which} all \update{comparison} models agreed on the correct answer, thereby excluding potentially ambiguous instances and} annotation errors.

\subsection{MIP Sampling}

After single-instance filtering, for each task $\tau$ we construct MIP inputs by sampling instances from $\mathcal{X}_{\mathrm{SIP}}$ using five different random seeds ($s \in \{1,2,3,4,5\}$). We evaluate ten MIP sample sizes
$n \in \mathcal{N} = \{2, 5, 10, 20, 50, 100, 200, 500, 1000, 2000\}$.
For each $(\tau, n, s)$, we prompt each model $\feedback{M}$ with the corresponding instance set, retaining only instances for which all models are correct in the SIP setting.

\section{RQ1: Performance and Failure Behaviours}
\label{sec:RQ1}

Our main goal is to evaluate LLM performance and failure behaviours in MIP settings, particularly as the number of instances increases.

\subsection{Performance Analysis}

\begin{figure*}[t]
\centering
  \includegraphics[width=0.92\textwidth]{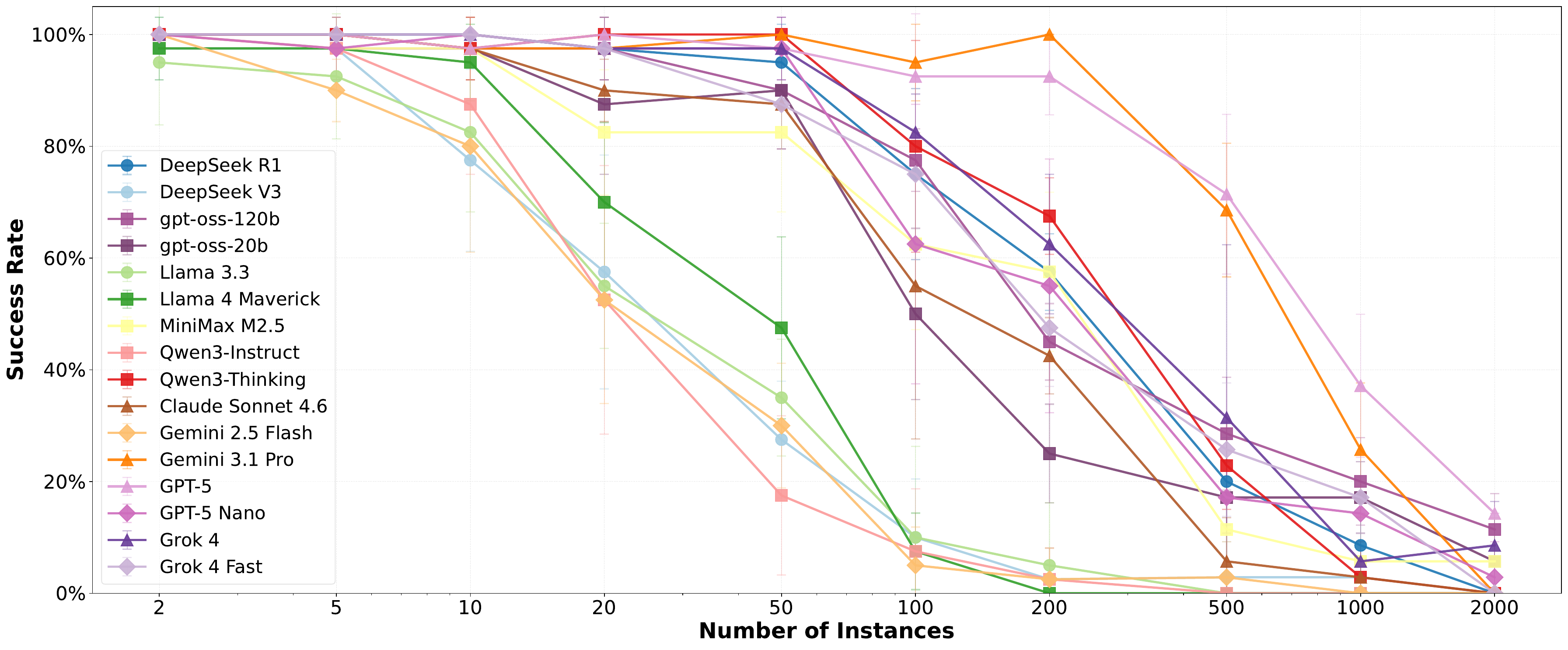}
  \caption{Model success rates (averaged across all tasks) as a function of the number of instances. Error bars indicate standard deviation across five random seeds. \update{LLMs from the same company share the same colour family, while markers denote categories: $\bullet$ (open-weight, $\geq$37B active parameters), $\blacksquare$ (open-weight, $\leq$22B active parameters), $\blacktriangle$ (frontier closed-source), and $\blacklozenge$ (lightweight closed-source).}}
  \label{fig:aggregated_model}
\end{figure*}

\begin{figure}[t]
  \includegraphics[width=\columnwidth]{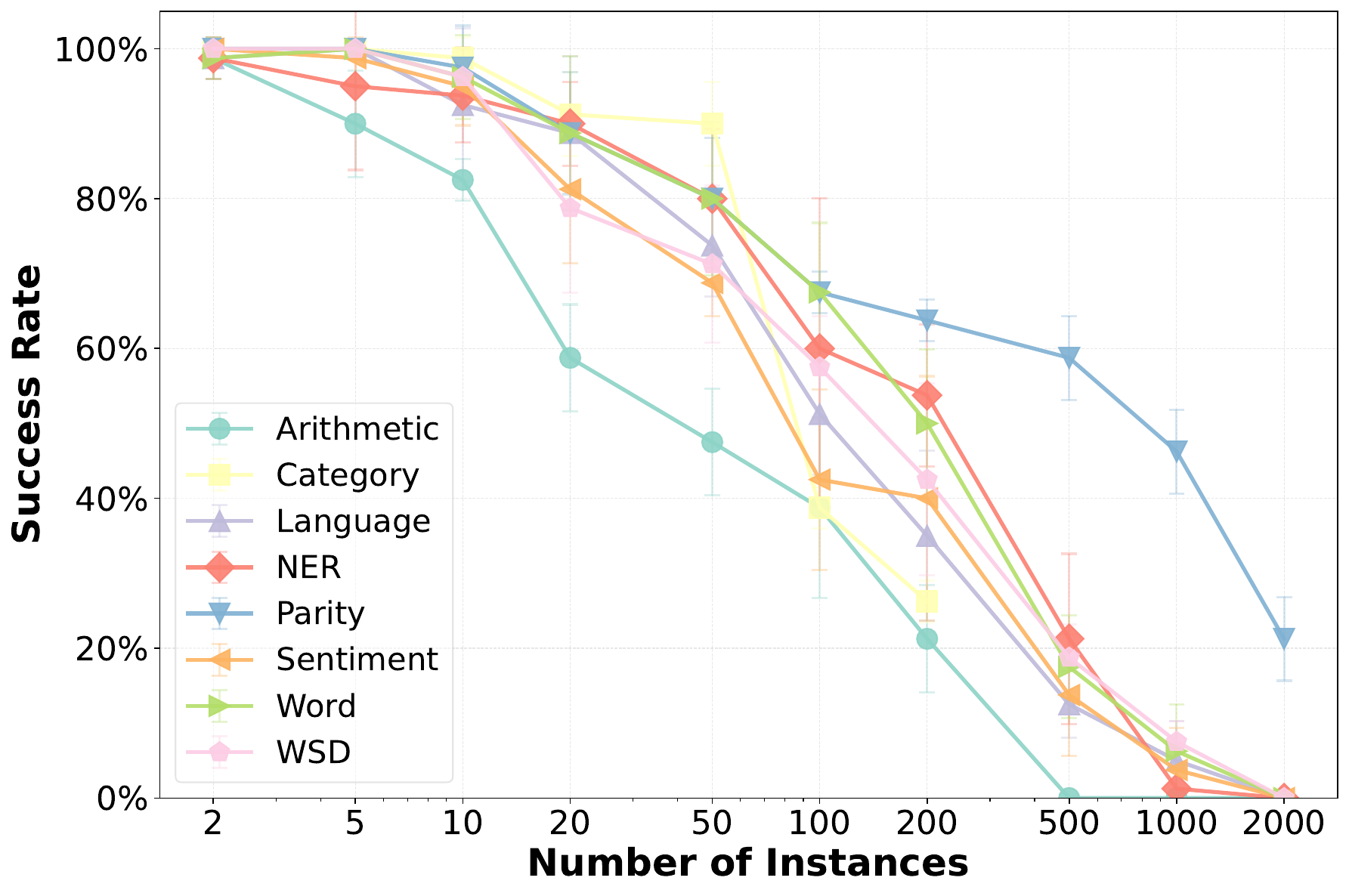}
  \caption{Task success rates (averaged across all LLMs) as a function of the number of instances. Error bars indicate standard deviation across five random seeds.}
  \label{fig:aggregated_task}
\end{figure}

\paragraph{\update{Gradual success rate degradation followed by collapse} \color{black}as the number of instances increases.}
\update{Figure~\ref{fig:aggregated_model}}
reports success rates aggregated across tasks for each model. We observe a consistent performance degradation as the number of instances increases. In particular, all models 
\update{show noticeable drops above 200} instances \feedback{and \update{near-}collapse beyond \update{1,0}00 instances}, \update{with} success rates fall\update{ing} below \update{2}0\% \update{at 2,000 instances}. 
\update{Figure~\ref{fig:aggregated_task}}
shows success rates aggregated across models for each task and reveals a similar downward trend.
With the exception of \emph{Arithmetic}, all tasks achieve success rates above \update{6}0\% when fewer than 50 instances\update{.}
Performance then deteriorates steadily as the instance count grows. Complete results by model and task are reported in Appendix~\ref{appendix:success_rate_details}.

\paragraph{LLM comparison.}
Table~\ref{tab:model-results-full} reports the success rate and invalid rate for all models. 
Overall, closed-source LLMs do not \update{consistently} exhibit superior performance.
\update{While frontier proprietary models (\emph{GPT-5}, \emph{Gemini 3.1 Pro} and \emph{Grok 4}) achieve the highest success rates, this advantage comes at substantially higher cost.}\footnote{\update{Given the high cost of frontier models and practical budget constraints, we restrict subsequent experiments to only lightweight closed-source and open-weight models.}}
\update{Beyond them, }\camera{\emph{Qwen3-Thinking},} \emph{gpt-oss-120b}, \camera{\emph{DeepSeek R1},} \emph{Grok 4 Fast} and \emph{GPT-5 Nano} achieve the highest success rates (above 65\%). Notably, only \update{\emph{Llama 4 Maverick} and} \emph{Grok 4 Fast} produce no invalid outputs, indicating greater robustness. 
While all models achieve success rates above 35\% on average, \camera{
\update{Figure~\ref{fig:aggregated_model}}
indicates}
that most successful cases occur when 
\camera{fewer than 500 instances are processed}.

\begin{table}[t]
  \centering
  \scalebox{0.9}{
  \begin{tabular}{lccc}
    \toprule
    \multirow{2}{*}{\textbf{Model}} &
    \textbf{Size\update{~/}} &
    \textbf{Success} &
    \textbf{Invalid} \\
    &
    \textbf{\update{Cost}}&
    \textbf{(\%)} &
    \textbf{(\%)} \\
    \midrule
\camera{DeepSeek R1}          & \camera{A37B}      & \camera{67.5{\footnotesize$\pm$2.6}} & \camera{2.9{\footnotesize$\pm$0.6}} \\
DeepSeek V3          & A37B      & 39.0{\footnotesize$\pm$3.6} & 2.9{\footnotesize$\pm$0.6} \\
gpt-oss-120b         & A5.1B      & 68.3{\footnotesize$\pm$2.8} & 3.6{\footnotesize$\pm$1.1} \\
gpt-oss-20b          & A3.6B       & 60.8{\footnotesize$\pm$2.5} & 4.9{\footnotesize$\pm$0.6} \\
Llama 3.3            & 70B       & 39.0{\footnotesize$\pm$3.8} & 2.9{\footnotesize$\pm$0.6} \\
Llama 4 Maverick     & A17B       & 43.1{\footnotesize$\pm$1.1} & \textbf{0.0}{\footnotesize$\pm$0.0} \\
\camera{MiniMax M2.5}       & \camera{A10B}      & \camera{62.3{\footnotesize$\pm$1.8}} & \camera{16.1{\footnotesize$\pm$2.7}} \\
Qwen3-Instruct       & A22B      & 37.9{\footnotesize$\pm$3.6} & 1.3{\footnotesize$\pm$0.0} \\
\camera{Qwen3-Thinking}       & \camera{A22B}      & \camera{69.4{\footnotesize$\pm$2.4}} & \camera{3.9{\footnotesize$\pm$1.6}} \\
\midrule
\update{Claude Sonnet 4.6}          & \update{\$ 4.68}      & \update{60.3{\footnotesize$\pm$2.7}} & \update{0.3{\footnotesize$\pm$0.6}} \\
Gemini 2.5 Flash     & \update{\$ 0.40}         & 37.7{\footnotesize$\pm$3.9} & 4.2{\footnotesize$\pm$3.2} \\
\update{Gemini 3.1 Pro} %
& \update{\$ 6.28}      & \update{80.3{\footnotesize$\pm$1.4}} & \update{2.6{\footnotesize$\pm$0.9}} \\
\update{GPT-5}          & \update{\$ 2.64}      & \update{\textbf{81.8}{\footnotesize$\pm$2.6}} & \update{1.8{\footnotesize$\pm$0.7}} \\
GPT-5 Nano           & \update{\$ 0.13}         & 66.5{\footnotesize$\pm$3.8} & 7.5{\footnotesize$\pm$0.6} \\
\update{Grok 4}          & \update{\$ 5.54}      & \update{70.6{\footnotesize$\pm$1.7}} & \update{1.3{\footnotesize$\pm$0.0}} \\
Grok 4 Fast          & \update{\$ 0.26}         & 67.0{\footnotesize$\pm$2.8} & \textbf{0.0}{\footnotesize$\pm$0.0} \\
\bottomrule
  \end{tabular}
  }
  \caption{Model success rate and invalid rate (mean{\footnotesize$\pm$std}), averaged across all tasks and instance counts. Standard deviation is computed over five random seeds. The top \camera{nine} models are open-weight LLMs, for which we report model size, with ``A'' denoting active parameters for mixture-of-experts models (total sizes 21B–671B). The bottom \update{seven} are closed-source LLMs with undisclosed sizes, for which we report average cost per task across five runs.}
  \label{tab:model-results-full}
\end{table}

\begin{figure}[t]
  \includegraphics[width=\columnwidth]{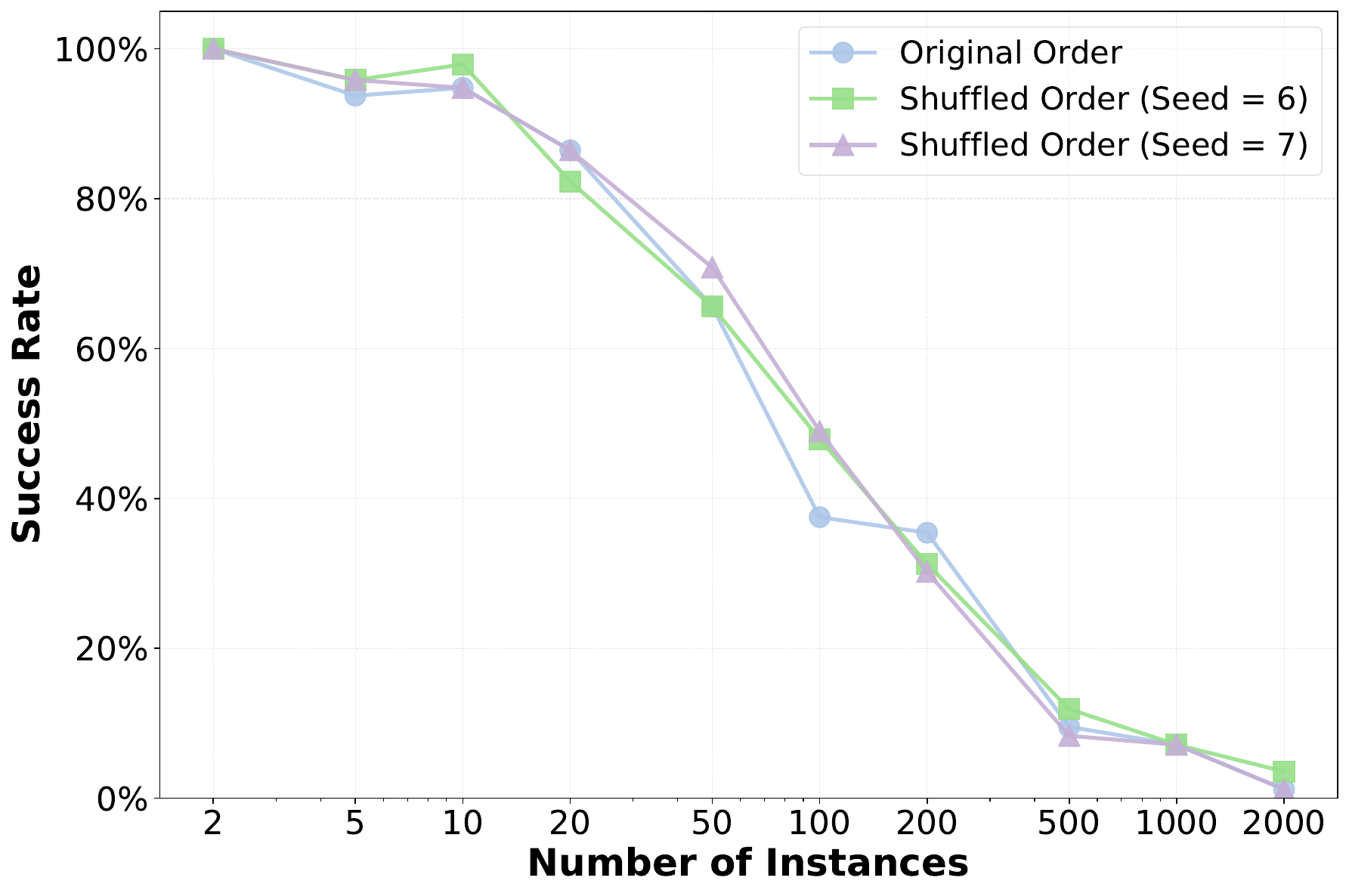}
  \caption{\camera{Success rates as a function of the number of instances for the original instance order and two shuffled variants constructed from the same instance sets.}}
  \label{fig:e3_compare}
\end{figure}

\paragraph{\jose{Robustness \feedback{to instance order}.}}
To examine whether instance order affects LLM performance, we conduct an additional robustness experiment. For all experiments whose instance sets were originally sampled with \jose{random seed} $s = 1$, we randomly shuffle the instance order twice (\jose{using $s = 6$ and $s = 7$}) and rerun the evaluation. Figure~\ref{fig:e3_compare} reports the resulting success rates as the number of instances increases. The degradation patterns remain highly consistent across different orderings of the same instance sets, suggesting that instance order has little effect on overall performance.

\subsection{Failure Behaviours}
\label{sec:failure_behaviours}
Beyond our default setting, which requires only an aggregated answer, we introduce an additional variant for more fine-grained analysis. In this variant, models are required to produce instance-level predictions $\{y_i\}_{i=1}^n$ before providing the aggregated answer. Even with such instance-level predictions, which provide 
\update{explicit intermediate reasoning}, the relative performance of models remains similar to the aggregated-only setting.

As described in Section~\ref{sec:formulation}, we consider two broad categories of failures: wrong answers and invalid outputs. Wrong answers include errors at the individual-instance level, the aggregation level, or both. Invalid outputs include (1) parsing errors, where the model output cannot be reliably parsed into the expected structured format, and (2) overlong input errors, where the input exceeds the model's allowable context length. To analyse failure behaviours in greater detail, we use the instance-level variant in the following experiments.

\begin{figure*}[t]
\centering
  \includegraphics[width=0.9\textwidth]{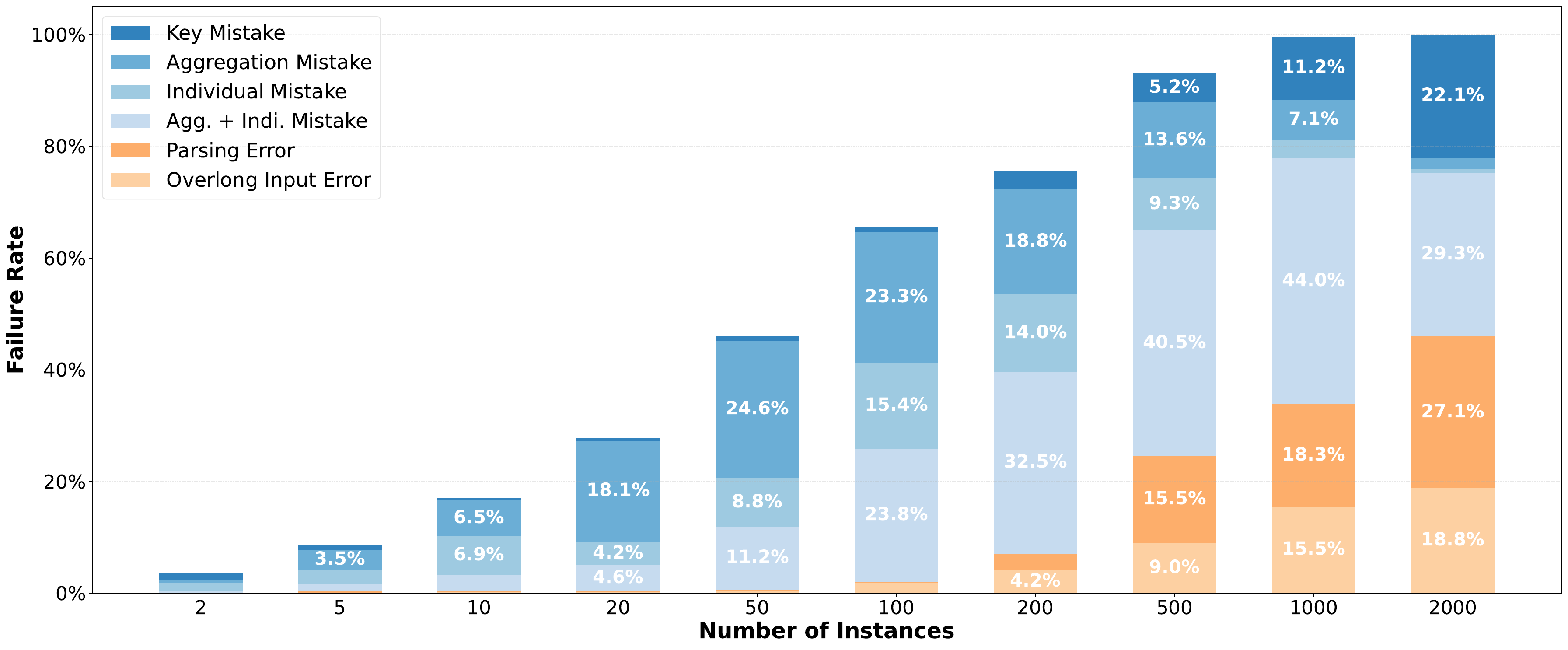}
  \caption{Breakdown of failure types. Key mistakes, aggregation mistakes, individual mistakes, and combined mistakes (Agg\camera{.}+Indi.) are categorised as wrong answers (blue), while parsing errors and overlong input errors are categorised as invalid outputs (orange).}
  \label{fig:error_id}
\end{figure*}

\paragraph{Different failure types emerge as the number of instances increases.}
Figure~\ref{fig:error_id} presents a stacked bar plot showing the contribution of each failure type as the instance count increases. Blue bars correspond to wrong answers, while orange bars correspond to invalid outputs. Wrong answers can occur even with as few as two instances. When the instance count exceeds 200, parsing errors increase substantially and reach nearly 30\% at 2{,}000 instances. Overlong input errors emerge \feedback{mainly} beyond \feedback{2}00 instances, primarily due to the \emph{Language} task, as non-English inputs typically require more prompt tokens~\citep{petrov2023language}.

\paragraph{Mistakes in individual instances and aggregation.}
To further characterise wrong answers, we 
\feedback{categorise them into}
four error types: key mistakes (keys do not span from 1 to $n$), individual mistakes (at least one instance-level prediction is incorrect), aggregation mistakes (all instance-level predictions are correct but the aggregated answer is incorrect), and combined mistakes (both instance-level and aggregation errors occur). As shown in Figure~\ref{fig:error_id}, when the instance count exceeds \feedback{100}, \camera{combined} mistakes increase markedly, accounting for approximately 25\% to 45\% of failures. Moreover, when aggregation mistakes and combined mistakes are considered together, aggregation remains challenging for LLMs regardless of instance-level correctness or instance count. Complete results by model and task are reported in Appendix~\ref{appendix:failure_breakdown_details}.

\paragraph{Model differences in making mistakes.}
Table~\ref{tab:incorrect-counts} compares models in terms of how individual-instance mistakes are distributed across experiments. Some models tend to concentrate many individual mistakes within a small number of failed experiments, while others exhibit more frequent but sparser errors. Focusing on individual mistakes, we observe that each failed experiment of \emph{Grok 4 Fast} typically contains many incorrect instance-level predictions. In contrast, \emph{Gemini 2.5 Flash} more often produces failed experiments with only a small number of individual mistakes\camera{, and \emph{MiniMax M2.5} shows low values in both metrics.}

\begin{table}[ht]
  \centering
  \scalebox{0.9}{
  \begin{tabular}{lcc}
  \toprule
  \textbf{Model} &
  \begin{tabular}{c}\textbf{Wrong}\\\textbf{Exp.}\end{tabular} &
  \begin{tabular}{c}\textbf{Wrong Indv.}\\\textbf{Per Exp.}\end{tabular} \\
  \midrule
\camera{DeepSeek R1}        & \camera{21.0{\footnotesize$\pm$3.7}} & \camera{49.5{\footnotesize$\pm$9.9}} \\
DeepSeek V3        & 29.6{\footnotesize$\pm$2.7} & 34.8{\footnotesize$\pm$13.1} \\
gpt-oss-120b       & 16.2{\footnotesize$\pm$2.9} & 37.5{\footnotesize$\pm$17.4} \\
gpt-oss-20b        & 15.8{\footnotesize$\pm$2.4} & 11.8{\footnotesize$\pm$5.0} \\
Llama 3.3          & 35.6{\footnotesize$\pm$2.2} & 53.8{\footnotesize$\pm$26.7} \\
Llama 4 Maverick   & 33.6{\footnotesize$\pm$4.2} & 38.4{\footnotesize$\pm$4.6} \\
\camera{MiniMax M2.5}        & ~~\camera{8.2{\footnotesize$\pm$1.5}} & ~~\camera{5.8{\footnotesize$\pm$5.5}} \\
Qwen3-Instruct     & 29.4{\footnotesize$\pm$2.1} & 20.8{\footnotesize$\pm$3.1} \\
\camera{Qwen3-Thinking}        & \camera{16.0{\footnotesize$\pm$2.9}} & \camera{15.4{\footnotesize$\pm$4.0}} \\
\midrule
Gemini 2.5 Flash   & 13.4{\footnotesize$\pm$1.3} & 5.5{\footnotesize$\pm$3.2} \\
GPT-5 Nano         & 11.2{\footnotesize$\pm$3.1} & 10.3{\footnotesize$\pm$15.3} \\
Grok 4 Fast        & 22.4{\footnotesize$\pm$1.7} & 117.3{\footnotesize$\pm$8.0} \\
\bottomrule
\end{tabular}
  }
  \caption{Percentage of experiments with at least one incorrect instance-level prediction (Wrong Exp.) and the average number of incorrect instance-level predictions per failed experiment (mean{\footnotesize$\pm$std}). Standard deviation is computed over five random seeds. %
  }
  \label{tab:incorrect-counts}
\end{table}

\paragraph{Self-awareness of limitations.}
Ideally, an LLM should recognise its own capability boundaries. A desirable behaviour is for the model to explicitly acknowledge in its reasoning $r$ when it cannot handle a large number of instances (e.g., by suggesting batch-wise processing or stating that the instance count exceeds its capacity). Such behaviour can be reflected by the model producing predictions for only the first few instances and omitting the remaining ones\feedback{, which corresponds to a key mistake (missing key)}. After inspecting the model outputs, we found that only \camera{171} out of \camera{4{,}620} experiments exhibit this omission, almost exclusively at instance counts of 500 or more, as expected. However, a manual analysis shows that only 27 out of these \camera{171} experiments explicitly suggest batch-wise processing, and most models do not warn users about such limitations. \emph{GPT-5 Nano} demonstrates this behaviour most frequently, explicitly indicating difficulty in 19 out of 28 cases. In contrast, \emph{DeepSeek V3}, \emph{gpt-oss-120b}, and \emph{gpt-oss-20b} do so in fewer than 20\% of their own omission cases, while no such behaviour is observed for the remaining models. Notably, although \emph{Qwen3-Instruct} and \emph{Gemini 2.5 Flash} do not acknowledge limitations, they also produce almost no cases of this omission.

\subsection{Discussion}

Overall, our findings answer RQ1 by showing that as the 
\update{instance count} increases, all LLMs experience degraded performance, characterised by lower success rates and more frequent failures. In particular, when the instance count reaches \update{1,0}00 or higher, no LLM achieves a success rate above 40\%.

Among the evaluated LLMs, \update{\emph{GPT-5}, \emph{Gemini 3.1 Pro}, \emph{Grok 4},}
\emph{gpt-oss-120b} \update{and} \camera{\emph{Qwen3-Thinking}}
exhibit the strongest overall performance. \update{Focusing on the lightweight and open-weight models,} although \emph{Gemini 2.5 Flash} achieves relatively low success rates, its failures typically involve 
\update{a relatively small number of} incorrect instances. At the same time, invalid outputs become increasingly common as the instance count grows. Notably, \emph{GPT-5 Nano} is the only 
model that 
\update{consistently identifies} when a task exceeds its capacity.

\section{RQ2: Context Length vs Number of Instances}

In the previous section, we analysed LLM behaviour in MIP settings as the number of instances increases. The results revealed consistent performance degradation across all models, albeit at different rates and to different extents. A natural question is whether this degradation is driven by increased context length, as has been observed in prior work across a range of tasks and settings (see Section~\ref{sec:related_work}). In this section, we analyse the effects of context length and instance count jointly, and 
\camera{examine }their respective contributions.

\subsection{Context Length Augmentation}
\label{sub:augmentation}

To \camera{study} the impact of context length, we design a setting in which the length of each individual instance is artificially increased without altering its original content.\footnote{We exclude \emph{Parity} from this experiment because each instance contains only a single number. For consistency with context length constraints, we also cap the maximum sample size at $n=1000$ in this setting \camera{($n=100$ for \textit{Category})}, as discussed below.}
For each sampled instance $x$, we construct a perturbed instance
$x' = x + \epsilon$, where $\epsilon$ denotes injected noise.
Following~\citet{hsiehruler2024}, we define $\epsilon$ as the string \camera{``- IRRELEVANT CONTEXT START - ''} followed by seven repetitions of the sentence ``The grass is green. The sky is blue. The sun is yellow. Here we go. There and back again.''\camera{, plus `` - IRRELEVANT CONTEXT END -''}. This choice is motivated by prior work showing that even irrelevant context can degrade model performance~\citep{shi2023large,yang-etal-2025-llm-reasoning}.
After noise injection, the average length of each instance (measured in the number of prompt tokens) more than doubles, increasing from approximately 136 tokens in the default setting to \camera{326} tokens in the artificially augmented setting.
 
\begin{figure}[t]
  \includegraphics[width=\columnwidth]{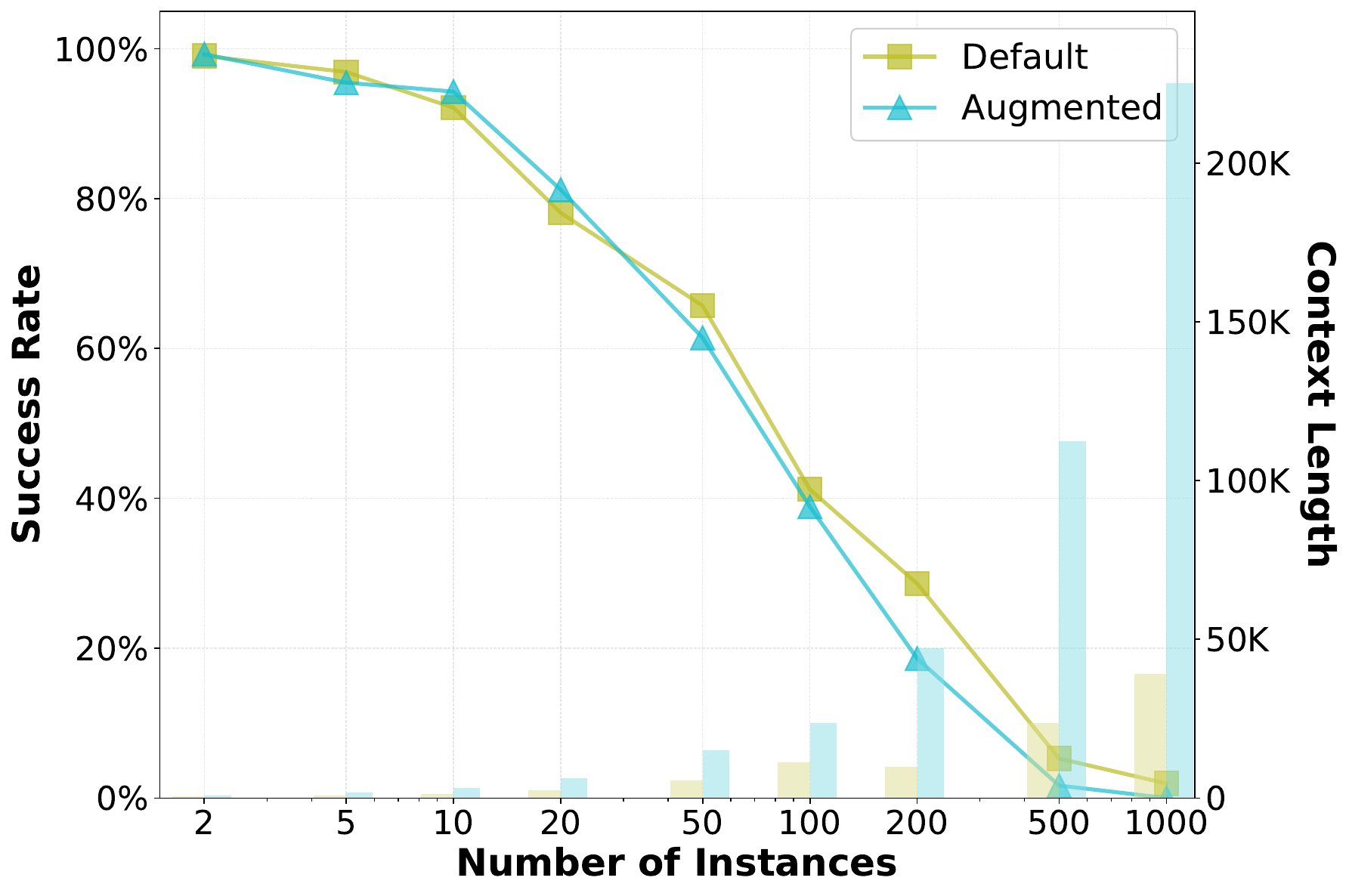}
  \caption{Success rate (lines) and total prompt token length (bars) in the artificial length setting as the number of instances increases. Error bars indicate standard deviation across five random seeds.}
  \label{fig:artificial_overall}
\end{figure}

\paragraph{Results.}
Figure~\ref{fig:artificial_overall} compares average performance across all tasks and models between the default and artificially augmented settings\camera{, as the number of instances increases}.
When the number of instances is held constant, the success rates of the two settings remain broadly similar, despite the average context length being more than twice as large in the augmented setting.
This indicates that artificially increasing the length of individual instances does not substantially impact performance, suggesting that context length alone is 
\camera{unlikely to fully explain the} performance degradation 
\camera{observed} in MIP settings.
\camera{Complete results by model and task are reported in Appendix~\ref{appendix:context_length}.}

\begin{figure}[t]
  \includegraphics[width=\columnwidth]{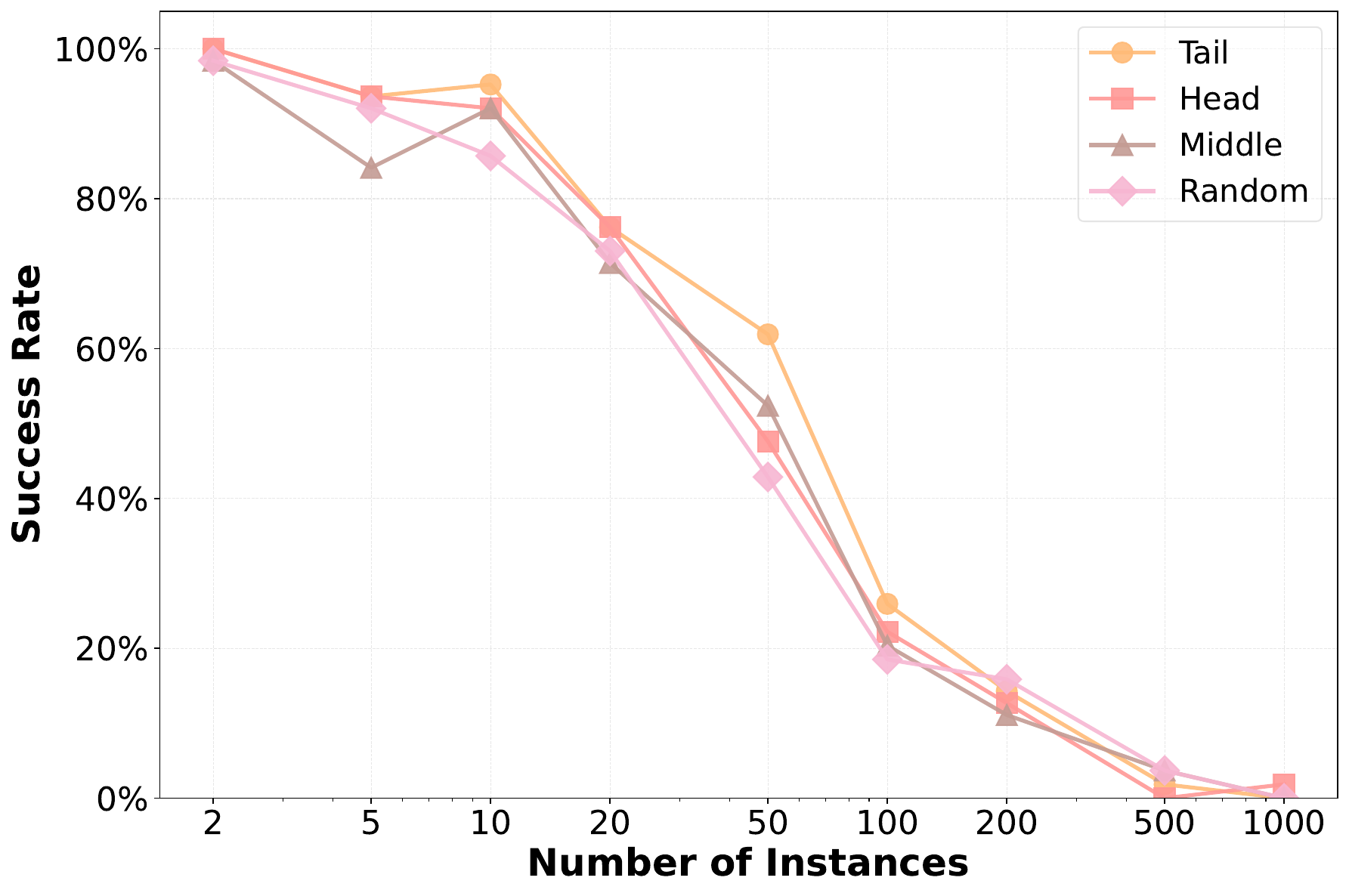}
  \caption{\camera{Success rates as a function of the number of instances for different injected-noise positions constructed from the same instance sets.}}
  \label{fig:e4_compare}
\end{figure}

\paragraph{\jose{Robustness \feedback{to noise position}.}} 
To further examine whether the position of injected noise affects model performance, we vary where the noise is inserted for the \update{same} instance set 
sampled \feedback{with $s = 1$.}
In addition to the tail setting used above, we also insert the noise at the head, in the middle, and at random positions within each instance.
Figure~\ref{fig:e4_compare} shows that the performance remains broadly similar across these variants, indicating that the position of injected noise does not lead to substantial performance differences.

\subsection{Correlation Analysis}
Motivated by the above findings, we conduct a correlation analysis to 
\camera{further examine the 
relationships}
of instance count and context length to overall model performance. Recall that in our default setting, instance samples are constructed using five random seeds. While this random sampling introduces some variation in individual instance lengths, the average prompt length tends to be similar across samples, especially when the number of instances is high. To obtain a wider range of context lengths for correlation analysis, we augment the default samples with two additional variants drawn from $\mathcal{X}_{\mathrm{SIP}}$: a \emph{long} set, consisting of the longest instances in each dataset, and a \emph{short} set, consisting of the shortest instances. This design enables a clearer 
\camera{comparison across different}
context lengths by introducing greater variation in total input length. %

\paragraph{Correlation with instance count and total context length.}
As an initial analysis, we compute the \jose{Spearman} correlation between success rate and each factor independently: the number of instances and the total context length. 
Both factors exhibit strong negative correlations, indicating that performance decreases as either quantity increases. However, the number of instances shows a notably stronger relationship, with Spearman correlations of \camera{$-0.61$}, compared to that of \camera{$-0.37$} for the context length. %
In both cases, the corresponding $p$-values are below $0.001$, indicating that the correlations are highly unlikely to arise by chance.

\paragraph{Correlation conditioned on the number of instances.}
The two factors are inherently related, as total context length grows with the number of instances. To 
\camera{better understand}
their effects, we additionally compute correlations while holding the number of instances fixed and examining variation only in context length. As shown in Table~\ref{tab:instance-correlation}, the resulting correlations between context length and success rate are substantially weaker, with values ranging between $-0.1\camera{5}$ and $0.1\camera{5}$, and $p$-values consistently above $0.1$. 
These results indicate that context length alone has limited explanatory power in this setting and 
suggest that the performance degradation observed in Section~\ref{sub:augmentation} is 
\camera{more strongly associated with}
the number of instances 
than total input length.

\begin{table}
  \centering
   \scalebox{0.9}{
  \begin{tabular}{ccc}
    \toprule
    \multicolumn{1}{c}{\multirow{2}{*}{\textbf{\begin{tabular}[c]{@{}c@{}}Number of\\ Instances\end{tabular}}}} &
    \multicolumn{2}{c}{\textbf{Success rate and Context length}} \\
    \cmidrule(lr){2-3}
    
    \multicolumn{1}{c}{} &
    \textbf{Correlation} & \textbf{P-Value} \\
    \midrule
    \camera{2}    & \camera{~0.043}  & \update{0.498} \\
    \camera{5}    & \camera{~0.044}  & \update{0.491} \\
    \camera{10}   & \camera{~0.044}  & \update{0.490} \\
    \camera{20}   & \camera{-0.042}  & \update{0.510} \\
    \camera{50}   & \camera{-0.033}  & \update{0.602} \\
    \camera{100}  & \camera{-0.014}  & \update{0.823} \\
    \camera{200}  & \camera{~0.056}  & \update{0.420} \\
    \camera{500}  & \camera{~0.117}  & \update{0.102} \\
    \camera{1000} & \camera{-0.053}  & \update{0.475} \\
    \bottomrule
  \end{tabular}
   }
  \caption{Correlation between success rate and prompt token length when the number of instances is fixed.}
  \label{tab:instance-correlation}
\end{table}

\subsection{Discussion}

Based on the above results, we 
\camera{find}
that the number of instances
\camera{plays a stronger role}
than context length in determining model success rates. When the number of instances is held fixed, the effect of context length appears to be 
\camera{comparatively limited}
according to our correlation analysis. LLMs have been shown to struggle when required to perform many repeated operations~\citep{fu2024large}, which is precisely what MIP entails as the instance count increases. This behaviour contrasts with RAG settings, where models primarily need to identify relevant contexts rather than process all inputs exhaustively. In MIP settings, LLMs must process each instance individually and aggregate the resulting outputs. While prior work has shown that LLMs can handle increasingly long contexts~\citep{liu2025comprehensive}, our findings suggest that improving reliability in MIP settings may \camera{also} require training strategies that explicitly target multi-instance reasoning and aggregation. %

\section{Conclusion}
In this paper, we presented a comprehensive evaluation of LLMs in MIP settings, that is, tasks that require aggregation of information from multiple instances to produce a final answer. The results show that LLMs are generally able to solve tasks involving a small number of instances, but begin to make mistakes as the number of instances increases. 
While the errors are initially small, this has important implications 
\update{for user trust}, since models are able to consistently solve the task when only a single instance is given. \jose{Moreover,} as the number of instances further increases, models' \jose{performance starts collapsing, in most cases without any warning to the user that this may happen.} %

Crucially, our experiments on context length highlight the importance of reasoning at the instance-count level, rather than focusing solely on context length as is commonly done. This has implications for how context should be processed in batches, for example when developing data science agents. Instead of relying only on context length for batching decisions in MIP settings, \jose{we should also consider} the number of instances \jose{as a relevant factor}. More generally, \jose{our} results 
suggest that models may benefit from training strategies that better support multi-instance reasoning and aggregation, especially in settings where accuracy \jose{and user trustworthiness are} paramount.

\section*{Limitations}

This work focuses on diagnosing failure modes of LLMs in multi-instance processing (MIP) settings, rather than proposing or validating concrete solutions. While our controlled evaluation reveals
performance degradation as the number of instances increases, we do not evaluate mitigation strategies such as task decomposition, external tool use, verification, or agentic designs. As a result, this paper should be interpreted primarily as an empirical characterisation of model behaviour rather than as a prescription for improving MIP reliability.

Our experiments emphasise exact aggregation tasks (e.g., counting, summation, exact class frequencies). Although these tasks are common in analytics-style applications, they may overemphasise brittleness in settings where approximate or semantic aggregation would suffice. The extent to which our findings generalise to softer aggregation objectives (e.g., majority voting, summarisation, or trend identification) therefore remains an open question. \update{In addition, our experiments rely on a fixed prompt template across tasks, which may not fully capture the variability of prompt formulations encountered in real-world applications.}

Despite our efforts to 
\camera{examine the roles of} instance count 
\update{and}
context length, these two factors are not entirely independent in practice. While correlation and controlled noise-injection analyses suggest that instance count 
\camera{has a stronger relationship with} degradation, more fine-grained causal analyses (e.g., controlled computational complexity or attention-level diagnostics) are left for future work. \jose{Moreover, we did not stress-test LLMs with extremely large context inputs, which might have given different results and perhaps a stronger context length effect.}

Finally, our study does not include model-internal interpretability analyses such as attention patterns or hidden-state dynamics, nor does it evaluate training-time interventions. In addition, our experiments are English-centric and limited to a set of selected LLMs, due to time and budget constraints. Future work could explore cross-lingual MIP behaviour and architectural or training modifications explicitly designed for multi-instance reasoning.

\section*{Ethical Considerations}

This work presents an empirical evaluation of LLMs in multi-instance processing settings and does not involve the collection of new data, interaction with human subjects, or the deployment of models in real-world decision-making systems. All datasets used in our experiments are publicly available and open-source, and were accessed and processed in accordance with their original licences and intended research use. Also, we do not claim that any particular model or provider is inherently unsafe or unsuitable. Rather, our results reflect general limitations of current LLMs under specific experimental conditions. We hope that this work contributes to more responsible deployment of LLM-based systems by encouraging practitioners to consider instance-level reliability, aggregation strategies, and potential failure modes when designing real-world applications.

\section*{\jose{Acknowledgments}}

\jose{Jose Camacho-Collados was supported by a UKRI Future Leaders Fellowship.}

\bibliography{bibexport}

\appendix

\section{Task and Data Source}
\label{appendix:data}
Here we introduce the details for each task and their data source.

\subsection{Arithmetic}
Th\update{e} task is about \textbf{arithmetic} calculation, which uses questions from Mathematics Dataset~\citep{saxtonanalysing2019}, including only easy addition or subtraction questions from its training split. The average word count for this task is 4.7.

\paragraph{Example Input:}
\begin{enumerate}
    \item What is the difference between -2 and 251860?
    \item -9,259,432 + 1
    \item What is 1,141.09 less than 1?
\end{enumerate}

\paragraph{SIP Ground Truth Output:}
\begin{enumerate}
    \item 251,862
    \item -9,259,431
    \item -1,140.09
\end{enumerate}

\paragraph{MIP Question:} Solve all the provided arithmetic questions and calculate the sum of all answers.

\paragraph{MIP Ground Truth Output:} -9,008,709.09

\subsection{Category}
Th\update{e} task is about news \textbf{category} classification, where a news \update{article} can belong to ``tech'' or ``business'' or ``entertainment'' or ``politics'' or ``sport''. The dataset is BBC News~\citep{greene2006practical}, where we use its training split. The average word count for this task is 371.4.

\paragraph{Example \update{I}nput:}
\begin{enumerate}
    \item german business confidence slides german business confidence fell in february knocking hopes of a speedy recovery in europe s largest economy...
    \item bbc poll indicates economic gloom citizens in a majority of nations surveyed in a bbc world service poll believe the world economy is worsening...
    \item lifestyle  governs mobile choice  faster  better or funkier hardware alone is not going to help phone firms sell more handsets  research suggests...
\end{enumerate}

\paragraph{SIP Ground Truth Output:}
\begin{enumerate}
    \item business
    \item business 
    \item tech
\end{enumerate}

\paragraph{MIP Question:} Count how many of the provided news articles belong to the ``tech'' category.

\paragraph{MIP Ground Truth Output:} 1

\subsection{Language}
Th\update{e} task is about \textbf{language} identification, where a paragraph can belong to ``English'' or ``Chinese'' or ``Persian'' or ``Spanish''. The dataset is WiLI-2018~\citep{thoma2018wili}. The average word count for this task is 55.8.

\paragraph{Example Input:}
\begin{enumerate}
    \item Nordahl Road is a station served by North County Transit District's SPRINTER light rail line...
    \item En Navidad de 1974, poco después de que interpretó la canción en francés película Papillon (Toi qui Regarde la Mer)...
    \item A talk by Takis Fotopoulos about the Internationalization of the Capitalist Market Economy and the project of Inclusive Democracy...
\end{enumerate}

\paragraph{SIP Ground Truth Output:}
\begin{enumerate}
    \item English
    \item Spanish
    \item English
\end{enumerate}

\paragraph{MIP Question:} Count how many paragraphs are in English.

\paragraph{MIP Ground Truth Output:} 2

\subsection{NER}
The task is about \textbf{n}amed \textbf{e}ntity \textbf{r}ecognition, which uses data from UNER~\citep{mayhew-etal-2024-universal}.
The average word count for this task is 16.0.

\paragraph{Example Input:}
\begin{enumerate}
    \item we love everything about the fence .
    \item i want to hook up with that girl \underline{paige} in the brown leather jacket .
    \item in addition , there is a reduction of 22,101 mmbtu which is the difference between the scada values ( best available ) that \underline{anita} showed on the february 29th storage sheet and the " official " february 29th values that \underline{gary wilson} received from mips .
\end{enumerate}

\paragraph{SIP Ground Truth Output:}
\begin{enumerate}
    \item 0
    \item 1
    \item 2
\end{enumerate}

\paragraph{MIP Question:} Count occurrences of the entity 'PERSON' in all sentences.

\paragraph{MIP Ground Truth Output:} 3

\subsection{Parity}
The task is about \textbf{parity} classification (i.e., identify if a number is ``odd'' or ``even''), where we use synthetic data generated by ourselves. The average word count for this task is 1 since only a single number is provided.

\paragraph{Example Input:}
\begin{enumerate}
    \item 18010
    \item 10160
    \item 89449
\end{enumerate}

\paragraph{SIP Ground Truth Output:}
\begin{enumerate}
    \item even
    \item even
    \item odd
\end{enumerate}

\paragraph{MIP Question:} Count how many of the provided numbers are odd.

\paragraph{MIP Ground Truth Output:} 1

\subsection{Sentiment}
The task is about \textbf{sentiment} analysis, where a movie review can belong to ``positive'' or ``negative''. The dataset is Sentiment Treebank~\citep{socher-etal-2013-recursive}, where we only use the ``most'' positive and negative reviews to avoid ambiguity. The average word count for this task is 18.7.

\paragraph{Example Input:}
\begin{enumerate}
    \item High Crimes is a cinematic misdemeanor , a routine crime thriller remarkable only for its lack of logic and misuse of two fine actors , Morgan Freeman and Ashley Judd .
    \item One of the worst movies of the year .
    \item A mix of gritty realism , crisp storytelling and radiant compassion that effortlessly draws you in .
\end{enumerate}

\paragraph{SIP Ground Truth Output:}
\begin{enumerate}
    \item negative
    \item negative
    \item positive
\end{enumerate}

\paragraph{MIP Question:} Count how many of the provided movie reviews are positive.

\paragraph{MIP Ground Truth Output:} 1

\subsection{Word}
The task is about tweets \textbf{word} occurrence (i.e., count a target word's occurrences in given tweets). The dataset is TweetEval~\citep{barbieri-etal-2020-tweeteval}, where we use its stance detection subset~\citep{mohammad-etal-2016-semeval}. The average word count for this task is 17.3.

\paragraph{Example Input:}
\begin{enumerate}
    \item IF FEMINISTS WERE HONEST   ``I want a worldwide matriarchal dictatorship with all men enslaved to \underline{women}''   \#GamerGate \#SemST 
    \item What the fuck do \underline{women} even do? I mean seriously they're just useless other than sex.  \#\underline{women}srights \#Feminist \#SemST 
    \item DEAR FEMINISTS  Start asking for accountability from man-haters instead of shielding them for convenient concealment.  \#SemST 
\end{enumerate}

\paragraph{SIP Ground Truth Output:}
\begin{enumerate}
    \item 1
    \item 2
    \item 0
\end{enumerate}

\paragraph{MIP Question:} Count occurrences of the word ``women'' in all tweets.

\paragraph{MIP Ground Truth Output:} 3

\subsection{WSD}
The task is about \textbf{w}ord \textbf{s}ense \textbf{d}isambiguation, where a target word ``apple'' is required to be distinguished as meaning either ``company'' or ``fruit'' based on its context. The dataset is CoarseWSD-20~\citep{loureiro-etal-2021-analysis}, where we use its ``apple'' subset. The average word count for this task is 31.4.

\paragraph{Example Input:}
\begin{enumerate}
    \item both seasons are available for download from apple 's itunes store .
    \item in klayman ii , the plaintiffs sued the same government defendants and in addition , facebook , yahoo! , google , microsoft , youtube , aol , paltalk , skype , sprint , at\&t , apple again alleging the bulk metadata collection violates the first , fourth and fifth amendment and constitutes divulgence of communication records in violation of section 2702 of stored communications act .
    \item description alongside dried pears the filling also contains raisin , walnut and other dried fruit such as apple or figs .

\end{enumerate}

\paragraph{SIP Ground Truth Output:}
\begin{enumerate}
    \item company
    \item company
    \item fruit
\end{enumerate}

\paragraph{MIP Question:} Count how many paragraphs contain the word "apple" referring to the company (Apple Inc.), not the fruit.

\paragraph{MIP Ground Truth Output:} 2

\subsection{Excluded Tasks}
Beyond the tasks mentioned above, we have three additional tasks that have been filtered out due to unsatisfactory SIP performance:

\begin{itemize}
        \item \textbf{B}igram \textbf{Shift} detection: from SentEval~\citep{conneau-kiela-2018-senteval}, which checks whether a bigram in a sentence has been shifted, with binary outcomes (i.e., shifted or not).
        \item \textbf{Subj}ect \textbf{Num}ber identification: from SentEval~\citep{conneau-kiela-2018-senteval}, which checks whether the subject of a sentence is ``plural'' or ``singular''.
        \item \textbf{Voice} classification: from Universal Dependencies~\citep{silveira-etal-2014-gold}, which checks whether a sentence is in the ``active'' or ``passive'' voice. Since no ground-truth labels are available, we use rule-based approaches to annotate the dataset.
\end{itemize}

\section{Prompt Template}
\label{appendix:prompt}
Here we present the prompt templates we use in our experiments. We \update{use the} example inputs in Appendix~\ref{appendix:data} for illustration.
\subsection{Default Setting}

\subsubsection{Arithmetic}
\begin{Verbatim}[breaklines=true, breakanywhere=true]
Task: Solve all the provided arithmetic questions and calculate the sum of all answers.
      
Instructions:
- Calculate the answer for each arithmetic operation
- Sum all individual answers
- Provide exact values without unnecessary trailing zeros (e.g., "5" not "5.0")
- Prefix negatives with '-' (e.g., "-42")
- For decimal results, keep only necessary decimal places (e.g., "3.14" not "3.140")
- If you only receive one question, the sum is just its answer

Response format:
Return a JSON object with:
- "reasoning": briefly explain your calculation process
- "answer": sum of all answers as a string

Example:
{"reasoning": "your approach here", "answer": "42"}

=== Here are the arithmetic questions ===
Question 1: What is the difference between -2 and 251860?
Question 2: -9259432 + 1
Question 3: What is 1141.09 less than 1?
...
=== End of arithmetic questions ===
\end{Verbatim}

\subsubsection{Category}
\begin{Verbatim}[breaklines=true, breakanywhere=true]
Task: Count how many of the provided news articles belong to the 'tech' category.
      
Background:
- Each news article belongs to one of 5 categories:
* business
* entertainment
* politics
* sport
* tech
- You need to classify each article based on its content and context

Instructions:
- Read each news article carefully
- Identify which category each article belongs to based on the content
- Count how many articles belong to the 'tech' category
- Do not count articles from other categories (business, entertainment, politics, sport)
- If you only receive one article, return 1 if it's tech, else return 0

Response format:
Return a JSON object with:
- "reasoning": briefly explain your approach to classifying news categories and how you counted
- "answer": integer count of tech news articles

Example:
{"reasoning": "your approach here", "answer": 42}
      
=== Here are the news articles ===
Article 1: german business confidence slides german business confidence fell in february knocking hopes of a speedy recovery in europe s largest economy...
Article 2: bbc poll indicates economic gloom citizens in a majority of nations surveyed in a bbc world service poll believe the world economy is worsening...
Article 3: lifestyle  governs mobile choice  faster  better or funkier hardware alone is not going to help phone firms sell more handsets  research suggests...
...
=== End of news articles ===
\end{Verbatim}

\subsubsection{Language}
\begin{Verbatim}[breaklines=true, breakanywhere=true]
Task: Count how many of the provided paragraphs are written in English.
      
Background:
The paragraphs are written in one of four languages:
- English (label 0)
- Chinese (label 1)
- Persian (label 2)
- Spanish (label 3)

Instructions:
- Read each paragraph carefully
- Identify the language of each paragraph
- Count how many paragraphs are written in English
- Do not count paragraphs in other languages
- If you only receive one paragraph, return 1 if it's English, else return 0

Response format:
Return a JSON object with:
- "reasoning": briefly explain your approach to identifying English paragraphs
- "answer": integer count of English paragraphs

Example:
{"reasoning": "your approach here", "answer": 5}
      
=== Here are the paragraphs ===
Paragraph 1: Nordahl Road is a station served by North County Transit District's SPRINTER light rail line...
Paragraph 2: En Navidad de 1974, poco después de que interpretó la canción en francés película Papillon (Toi qui Regarde la Mer)...
Paragraph 3: A talk by Takis Fotopoulos about the Internationalization of the Capitalist Market Economy and the project of Inclusive Democracy...
...
=== End of paragraphs ===

\end{Verbatim}

\subsubsection{NER}
\begin{Verbatim}[breaklines=true, breakanywhere=true]
Task: Count how many times the entity "PERSON" appears across all provided sentences.
      
Background:
- An entity may consist of multiple words that form a contiguous fragment in the text
- You need to first identify entities in the sentence (named entity recognition), then count them
- Two entities may appear consecutively without punctuation or words between them
- No entity overlaps occur (each word belongs to at most one entity)
- Some words do not belong to any entity

Entity Definition:
- PERSON: names of people, real or fictional, but not nominals

Instructions:
- Identify all PERSON entities across all sentences
- Count the total number of PERSON entity mentions (not unique entities, but total occurrences)
- Each distinct mention counts as one occurrence, even if it refers to the same person

Response format:
Return a JSON object with:
- "reasoning": briefly explain how you identified the entities and counted them
- "answer": integer count of PERSON entities

Example:
{"reasoning": "your approach here", "answer": 5}
      
=== Here are the sentences ===
Sentence 1: we love everything about the fence .
Sentence 2: i want to hook up with that girl paige in the brown leather jacket .
Sentence 3: in addition , there is a reduction of 22,101 mmbtu which is the difference between the scada values ( best available ) that anita showed on the february 29th storage sheet and the " official " february 29th values that gary wilson received from mips .
...
=== End of sentences ===
\end{Verbatim}

\subsubsection{Parity}
\begin{Verbatim}[breaklines=true, breakanywhere=true]
Task: Count how many of the provided numbers are odd.
      
Background:
- An odd number is an integer that is not evenly divisible by 2
- Odd numbers end in 1, 3, 5, 7, or 9
- An even number is an integer that is evenly divisible by 2
- Even numbers end in 0, 2, 4, 6, or 8

Instructions:
- Check each number to determine if it is odd or even
- Count how many numbers are odd
- Do not count even numbers
- If you only receive one number, return 1 if it's odd, else return 0

Response format:
Return a JSON object with:
- "reasoning": briefly explain your approach to identifying odd numbers
- "answer": integer count of odd numbers

Example:
{"reasoning": "your approach here", "answer": 3}

=== Here are the numbers ===
Number 1: 18010
Number 2: 10160
Number 3: 89449
...
=== End of numbers ===
\end{Verbatim}

\subsubsection{Sentiment}
\begin{Verbatim}[breaklines=true, breakanywhere=true]
Task: Count how many of the provided movie reviews are positive.
      
Instructions:
- Each review has a sentiment: positive or negative
- Count only the reviews with positive sentiment
- Return the total count of positive reviews
- If you only receive one review, return 1 if it's positive, else return 0

Response format:
Return a JSON object with:
- "reasoning": briefly explain how you solved this task
- "answer": integer count of positive reviews

Example:
{"reasoning": "your approach here", "answer": 42}

=== Here are the movie reviews ===
Review 1: High Crimes is a cinematic misdemeanor , a routine crime thriller remarkable only for its lack of logic and misuse of two fine actors , Morgan Freeman and Ashley Judd .
Review 2: One of the worst movies of the year .
Review 3: A mix of gritty realism , crisp storytelling and radiant compassion that effortlessly draws you in .
...
=== End of movie reviews ===
\end{Verbatim}

\subsubsection{Word}
\begin{Verbatim}[breaklines=true, breakanywhere=true]
Task: Count how many times the word "women" appears in the provided tweets.
      
Instructions:
- Search is case-insensitive (e.g., "women", "Women", "WOMEN" all count)
- Count occurrences that include the substring "women" (e.g., "women", "womens", "women's", "womenfolk")
- Do not count forms that lack the substring "women" (e.g., "woman", "womankind")
- Count all occurrences across all tweets, not just unique tweets
- If you only receive one tweet, just return the occurrence for that tweet

Response format:
Return a JSON object with:
- "reasoning": briefly explain how you counted the matches
- "answer": integer total count

Example:
{"reasoning": "your approach here", "answer": 42}

=== Here are the tweets ===
Tweet 1: IF FEMINISTS WERE HONEST   "I want a worldwide matriarchal dictatorship with all men enslaved to women"   \#GamerGate \#SemST 
Tweet 2: What the fuck do women even do? I mean seriously they're just useless other than sex.  \#womensrights \#Feminist \#SemST 
Tweet 3: DEAR FEMINISTS  Start asking for accountability from man-haters instead of shielding them for convenient concealment.  \#SemST 
...
=== End of tweets ===
\end{Verbatim}

\subsubsection{WSD}
\begin{Verbatim}[breaklines=true, breakanywhere=true]
Task: Count how many paragraphs contain the word "apple" referring to the company (Apple Inc.), not the fruit.
      
Background:
- Each paragraph contains exactly one occurrence of the word "apple" (case-insensitive, as a complete word, not as part of another word)
- This "apple" can mean either:
* The company: Apple Inc., the technology company
* The fruit: the edible fruit that grows on apple trees
- You need to determine the meaning of "apple" in each paragraph based on context

Instructions:
- Read each paragraph carefully
- Identify whether "apple" refers to the company or the fruit based on contextual clues
- Count how many paragraphs where "apple" means the company (Apple Inc.)
- Do not count paragraphs where "apple" means the fruit
- If you only receive one paragraph, return 1 if it means the company, else return 0

Response format:
Return a JSON object with:
- "reasoning": briefly explain your approach to disambiguating the word sense and how you counted
- "answer": integer count of paragraphs where "apple" means the company

Example:
{"reasoning": "your approach here", "answer": 42}

=== Here are the paragraphs ===
Paragraph 1: both seasons are available for download from apple 's itunes store .
Paragraph 2: in klayman ii , the plaintiffs sued the same government defendants and in addition , facebook , yahoo! , google , microsoft , youtube , aol , paltalk , skype , sprint , at\&t , apple again alleging the bulk metadata collection violates the first , fourth and fifth amendment and constitutes divulgence of communication records in violation of section 2702 of stored communications act .
Paragraph 3: description alongside dried pears the filling also contains raisin , walnut and other dried fruit such as apple or figs .
...
=== End of paragraphs ===
\end{Verbatim}

\subsection{Instance-Level Setting}
\subsubsection{Arithmetic}
\begin{Verbatim}[breaklines=true, breakanywhere=true]
Task: Solve each of the provided arithmetic questions and calculate the sum of all answers.
      
Instructions:
- Calculate the answer for each arithmetic question
- Use 1-based indexing for question numbers ("1", "2", "3", …)
- Provide exact values without unnecessary trailing zeros (e.g., "5" not "5.0")
- Prefix negatives with '-' (e.g., "-42")
- For decimal results, keep only necessary decimal places (e.g., "3.14" not "3.140")
- Sum all individual answers

Response format:
Return a JSON object with:
- One key per question: "1", "2", "3", …, mapping to the answer as a string
- "sum": sum of all answers as a string
- "reasoning": brief explanation of your approach

Example for 3 questions:
{"1": "8", "2": "-30", "3": "5.6", "sum": "-16.4", "reasoning": "your approach here"}
      
=== Here are the arithmetic questions ===
Question 1: What is the difference between -2 and 251860?
Question 2: -9259432 + 1
Question 3: What is 1141.09 less than 1?
...
=== End of arithmetic questions ===
\end{Verbatim}

\subsubsection{Category}
\begin{Verbatim}[breaklines=true, breakanywhere=true]
Task: For each news article, classify it into one of 5 categories, then provide total counts for each category.
      
Background:
- Each news article belongs to one of 5 categories:
* business (label 0)
* entertainment (label 1)
* politics (label 2)
* sport (label 3)
* tech (label 4)
- You need to classify each article based on its content and context

Instructions:
- Read each news article carefully and classify it into the most appropriate category
- For each article, assign:
* 0 if it's business news
* 1 if it's entertainment news
* 2 if it's politics news
* 3 if it's sport news
* 4 if it's tech news
- Count the total number of articles for each category
- Provide classification for each article along with summary counts

Response format:
Return a JSON object with:
- One key per article: "1", "2", "3", …, mapping to the category label (0-4)
- "business": integer count of articles classified as business
- "entertainment": integer count of articles classified as entertainment
- "politics": integer count of articles classified as politics
- "sport": integer count of articles classified as sport
- "tech": integer count of articles classified as tech
- "reasoning": brief explanation of your approach to news classification

Example for 3 articles:
{"1": 4, "2": 0, "3": 3, "business": 1, "entertainment": 0, "politics": 0, "sport": 1, "tech": 1, "reasoning": "your approach here"}
    
=== Here are the news articles ===
Article 1: german business confidence slides german business confidence fell in february knocking hopes of a speedy recovery in europe s largest economy...
Article 2: bbc poll indicates economic gloom citizens in a majority of nations surveyed in a bbc world service poll believe the world economy is worsening...
Article 3: lifestyle  governs mobile choice  faster  better or funkier hardware alone is not going to help phone firms sell more handsets  research suggests...
...
=== End of news articles ===
\end{Verbatim}

\subsubsection{Language}
\begin{Verbatim}[breaklines=true, breakanywhere=true]
Task: For each paragraph, identify which language it is written in, then provide summary counts for all categories.
      
Background:
The paragraphs are written in one of four languages:
- English (label 0)
- Chinese (label 1)
- Persian (label 2)
- Spanish (label 3)

Instructions:
- Read each paragraph carefully and identify its language
- Classify each paragraph using the labels:
* 0 = English
* 1 = Chinese
* 2 = Persian
* 3 = Spanish
- Provide the classification for each individual paragraph
- Also provide summary counts for all four language categories

Response format:
Return a JSON object with:
- One key per paragraph: "1", "2", "3", …, mapping to the classification (0, 1, 2, or 3)
- "english": integer count of paragraphs classified as English
- "chinese": integer count of paragraphs classified as Chinese
- "persian": integer count of paragraphs classified as Persian
- "spanish": integer count of paragraphs classified as Spanish
- "reasoning": brief explanation of your approach

Example for 5 paragraphs:
{"1": 0, "2": 1, "3": 2, "4": 3, "5": 0, "english": 2, "chinese": 1, "persian": 1, "spanish": 1, "reasoning": "your approach here"}
      
=== Here are the paragraphs ===
Paragraph 1: Nordahl Road is a station served by North County Transit District's SPRINTER light rail line...
Paragraph 2: En Navidad de 1974, poco después de que interpretó la canción en francés película Papillon (Toi qui Regarde la Mer)...
Paragraph 3: A talk by Takis Fotopoulos about the Internationalization of the Capitalist Market Economy and the project of Inclusive Democracy...
...
=== End of paragraphs ===
\end{Verbatim}

\subsubsection{NER}
\begin{Verbatim}[breaklines=true, breakanywhere=true]
Task: Count how many times the entity "PERSON" appears in each sentence and provide the total count.
      
Background:
- An entity may consist of multiple words that form a contiguous fragment in the text
- You need to first identify entities in each sentence (named entity recognition), then count them
- Two entities may appear consecutively without punctuation or words between them
- No entity overlaps occur (each word belongs to at most one entity)
- Some words do not belong to any entity

Entity Definition:
- PERSON: names of people, real or fictional, but not nominals

Instructions:
- Identify all PERSON entities in each sentence
- Count the number of PERSON entity mentions in each sentence separately
- Provide the count for each sentence along with the total count across all sentences
- Each distinct mention counts as one occurrence, even if it refers to the same person

Response format:
Return a JSON object with:
- One key per sentence: "1", "2", "3", …, mapping to the integer count of PERSON mentions in that sentence
- "total": total count of PERSON entities across all sentences
- "reasoning": brief explanation of how you identified and counted the PERSON entities

Example:
{"1": 0, "2": 2, "3": 1, "total": 3, "reasoning": "your approach here"}
      
=== Here are the sentences ===
Sentence 1: we love everything about the fence .
Sentence 2: i want to hook up with that girl paige in the brown leather jacket .
Sentence 3: in addition , there is a reduction of 22,101 mmbtu which is the difference between the scada values ( best available ) that anita showed on the february 29th storage sheet and the " official " february 29th values that gary wilson received from mips .
...
=== End of sentences ===
\end{Verbatim}

\subsubsection{Parity}
\begin{Verbatim}[breaklines=true, breakanywhere=true]
Task: For each number, identify whether it is odd or even, then provide summary counts for both categories.
      
Background:
- An odd number is an integer that is not evenly divisible by 2
- Odd numbers end in 1, 3, 5, 7, or 9 (label 1)
- An even number is an integer that is evenly divisible by 2
- Even numbers end in 0, 2, 4, 6, or 8 (label 0)

Instructions:
- Check each number to determine if it is odd or even
- Classify each number using the labels:
* 1 = odd
* 0 = even
- Provide the classification for each individual number
- Also provide summary counts for both odd and even categories

Response format:
Return a JSON object with:
- One key per number: "1", "2", "3", …, mapping to the classification (0 or 1)
- "odd": integer count of numbers classified as odd
- "even": integer count of numbers classified as even
- "reasoning": brief explanation of your approach

Example for 5 numbers:
{"1": 0, "2": 1, "3": 0, "4": 1, "5": 1, "odd": 3, "even": 2, "reasoning": "your approach here"}
      
=== Here are the numbers ===
Number 1: 18010
Number 2: 10160
Number 3: 89449
...
=== End of numbers ===
\end{Verbatim}

\subsubsection{Sentiment}
\begin{Verbatim}[breaklines=true, breakanywhere=true]
Task: For each movie review, classify whether it is positive or negative, then provide summary counts.
      
Instructions:
- Classify each review as either:
- 0 = negative sentiment
- 1 = positive sentiment
- Provide the classification for each individual review
- Also provide summary counts for negative and positive reviews

Response format:
Return a JSON object with:
- One key per review: "1", "2", "3", …, mapping to the classification (0 or 1)
- "negative": integer count of reviews classified as negative
- "positive": integer count of reviews classified as positive
- "reasoning": brief explanation of your approach to classification

Example for 3 reviews:
{"1": 1, "2": 0, "3": 1, "negative": 1, "positive": 2, "reasoning": "your approach here"}
      
=== Here are the movie reviews ===
Review 1: High Crimes is a cinematic misdemeanor , a routine crime thriller remarkable only for its lack of logic and misuse of two fine actors , Morgan Freeman and Ashley Judd .
Review 2: One of the worst movies of the year .
Review 3: A mix of gritty realism , crisp storytelling and radiant compassion that effortlessly draws you in .
...
=== End of movie reviews ===
\end{Verbatim}

\subsubsection{Word}
\begin{Verbatim}[breaklines=true, breakanywhere=true]
Task: For each tweet, count how many times the word "women" appears, then provide the total count.
      
Instructions:
- Search is case-insensitive (e.g., "women", "Women", "WOMEN" all count)
- Count occurrences that include the substring "women" (e.g., "women", "womens", "women's", "womenfolk")
- Do not count forms that lack the substring "women" (e.g., "woman", "womankind")
- Count occurrences in each tweet separately
- Provide the per-tweet counts plus the overall total

Response format:
Return a JSON object with:
- One key per tweet: "1", "2", "3", …, mapping to the count of "women" in that tweet (integer)
- "total": integer representing the total count of "women" across all tweets
- "reasoning": brief explanation of your counting approach

Example for 3 tweets:
{"1": 2, "2": 0, "3": 1, "total": 3, "reasoning": "your approach here"}
      
=== Here are the tweets ===
Tweet 1: IF FEMINISTS WERE HONEST   "I want a worldwide matriarchal dictatorship with all men enslaved to women"   \#GamerGate \#SemST 
Tweet 2: What the fuck do women even do? I mean seriously they're just useless other than sex.  \#womensrights \#Feminist \#SemST 
Tweet 3: DEAR FEMINISTS  Start asking for accountability from man-haters instead of shielding them for convenient concealment.  \#SemST 
...
=== End of tweets ===
\end{Verbatim}

\subsubsection{WSD}
\begin{Verbatim}[breaklines=true, breakanywhere=true]
Task: For each paragraph, identify whether the word "apple" refers to the company or the fruit, then provide total counts.
      
Background:
- Each paragraph contains exactly one occurrence of the word "apple" (case-insensitive, as a complete word, not as part of another word)
- This "apple" can mean either:
* The company (label 0): Apple Inc., the technology company
* The fruit (label 1): the edible fruit that grows on apple trees
- You need to determine the meaning of "apple" in each paragraph based on context

Instructions:
- Read each paragraph carefully and classify the meaning of "apple"
- For each paragraph, assign:
* 0 if "apple" means the company (Apple Inc.)
* 1 if "apple" means the fruit
- Count the total number of paragraphs for each category
- Provide classification for each paragraph along with summary counts

Response format:
Return a JSON object with:
- One key per paragraph: "1", "2", "3", …, mapping to either 0 (company) or 1 (fruit)
- "company": integer count of paragraphs where "apple" means the company
- "fruit": integer count of paragraphs where "apple" means the fruit
- "reasoning": brief explanation of your approach to word sense disambiguation

Example for 3 paragraphs:
{"1": 0, "2": 1, "3": 0, "company": 2, "fruit": 1, "reasoning": "your approach here"}

=== Here are the paragraphs ===
Paragraph 1: both seasons are available for download from apple 's itunes store .
Paragraph 2: in klayman ii , the plaintiffs sued the same government defendants and in addition , facebook , yahoo! , google , microsoft , youtube , aol , paltalk , skype , sprint , at\&t , apple again alleging the bulk metadata collection violates the first , fourth and fifth amendment and constitutes divulgence of communication records in violation of section 2702 of stored communications act .
Paragraph 3: description alongside dried pears the filling also contains raisin , walnut and other dried fruit such as apple or figs .
...
=== End of paragraphs ===
\end{Verbatim}

\section{Single Instance Filtering Results}
\label{appendix:single}

\update{
\camera{W}e note that not all models in our evaluation are included in this SIP filtering analysis, as the filtering procedure is defined based on a fixed subset of \camera{comparison} models; however, the excluded models \jose{generally} exhibit \jose{stronger} performance in the MIP setting, \jose{and are 
\camera{likely to perform similarly under the}
SIP settings 
(e.g., for this filtering, we used the smaller versions of the closed-source LLMs considered)}.} %

\subsection{Individual Results}
\label{appendix:single_indi}
Table~\ref{tab:single-filtering-removed} reports the agreements for the tasks that are removed, while Table~\ref{tab:taskwise-accuracy-rotated} shows the average SIP success rate for each model and task. Although \emph{BShift} achieves a relatively high SIP success rate for the selected models, its agreement across models is low.

\begin{table}
  \centering
  \begin{tabular}{lc}
    \hline
    \textbf{Task} & \textbf{Agreement (\%)} \\
    \hline
    BShift     & 69.3 \\
    SubjNum    & 82.5 \\
    Voice      & 78.3 \\
    \hline
  \end{tabular}
  \caption{Task filtering across tasks that have been removed.}
  \label{tab:single-filtering-removed}
\end{table}

\begin{sidewaystable}
  \centering
  \scalebox{0.95}{
  \begin{tabular}{lcccccccccccc}
    \hline
    \textbf{Model} &
    \textbf{Arithmetic} &
    \textbf{Language} &
    \textbf{NER} &
    \textbf{News} &
    \textbf{Parity} &
    \textbf{BShift} &
    \textbf{SubjNum} &
    \textbf{Sentiment} &
    \textbf{Tweets} &
    \textbf{Voice} &
    \textbf{WSD} &
    \textbf{Average} \\
    \hline
    DeepSeek V3 &
    98.3 & 99.2 & 95.8 & 97.6 & 100.0 &
    \underline{92.7} & \underline{88.3} & 99.3 & 99.6 &
    \underline{89.5} & 99.4 & 96.3 \\
    gpt-oss-120b &
    97.2 & 99.5 & 96.2 & 97.6 & 100.0 &
    \underline{92.6} & \underline{89.4} & 98.5 & 100.0 &
    \underline{91.4} & 99.3 & 96.5 \\
    gpt-oss-20b &
    97.7 & 99.5 & 96.2 & 97.2 & 100.0 &
    \underline{90.4} & \underline{89.6} & 97.8 & 100.0 &
    \underline{90.6} & 99.1 & 96.2 \\
    Llama 3.3 &
    93.7 & 99.5 & 93.6 & 98.4 & 100.0 &
    \underline{90.9} & \underline{87.4} & 99.7 & 99.5 &
    \underline{86.9} & 99.2 & 95.3 \\
    Llama 4 Maverick &
    96.7 & 99.3 & 94.3 & 99.2 & 100.0 &
    \underline{90.9} & \underline{88.9} & 99.4 & 99.9 &
    \underline{91.7} & 99.4 & 96.3 \\
    \underline{Llama 4 Scout} &
    \underline{93.4} & \underline{97.9} & \underline{93.9} & \underline{97.2} & \underline{100.0} &
    \underline{86.9} & \underline{87.3} & \underline{99.0} & \underline{99.5} &
    \underline{88.7} & \underline{99.1} & \underline{94.8} \\
    \underline{Mistral NeMo} &
    \underline{72.0} & \underline{96.2} & \underline{78.1} & \underline{96.8} & \underline{75.3} &
    \underline{62.9} & \underline{75.2} & \underline{83.8} & \underline{60.3} &
    \underline{72.7} & \underline{98.4} & \underline{79.2} \\
    Qwen3-Instruct &
    99.5 & 99.4 & 97.3 & 99.2 & 100.0 &
    \underline{91.3} & \underline{88.9} & 99.5 & 99.7 &
    \underline{92.8} & 99.3 & 97.0 \\
    \hline
    Gemini 2.5 Flash &
    96.2 & 99.6 & 96.5 & 98.0 & 100.0 &
    \underline{92.8} & \underline{88.9} & 99.0 & 99.4 &
    \underline{92.0} & 99.4 & 96.5 \\
    GPT-5 Nano &
    98.0 & 99.5 & 96.5 & 97.2 & 100.0 &
    \underline{93.0} & \underline{89.0} & 99.2 & 100.0 &
    \underline{91.3} & 99.2 & 96.6 \\
    Grok 4 Fast &
    97.9 & 99.5 & 97.2 & 98.8 & 100.0 &
    \underline{95.7} & \underline{88.6} & 99.5 & 100.0 &
    \underline{90.2} & 99.4 & 97.0 \\
    \hline
  \end{tabular}
  }
  \caption{SIP success rate (\%) across tasks. Open-weight models are listed first, followed by closed-source models. Underlined rows and columns indicate the tasks and LLMs that are excluded because they do not satisfy the filtering criteria described in Section~\ref{sec:filtering}.}
  \label{tab:taskwise-accuracy-rotated}
\end{sidewaystable}

\subsection{Task Filtering Results}
\label{appendix:single_final}
\update{Table~\ref{tab:single-filtering} shows the retained instance percentage for each task, along with the corresponding maximum and minimum SIP success rates across models.}
\begin{table}
  \centering
  \scalebox{0.9}{
  \begin{tabular}{lccc}
    \toprule
    \textbf{Task} &
    \begin{tabular}{c}\textbf{Agreement}\\\textbf{(\%)}\end{tabular} &
    \begin{tabular}{c}\textbf{Max}\\\textbf{(\%)}\end{tabular} &
    \begin{tabular}{c}\textbf{Min}\\\textbf{(\%)}\end{tabular} \\
    \midrule
    Arithmetic & 89.0   & 99.5   & 93.7   \\
    Category   & 94.8   & 99.2   & 97.2   \\
    Language   & 98.8   & 99.6   & 99.2   \\
    NER        & 87.6   & 97.3   & 93.6   \\
    Parity     & 100.0~~  & 100.0~~  & 100.0~~  \\
    Sentiment  & 96.4   & 99.7   & 97.8   \\
    Word       & 98.3   & 100.0~~  & 99.4   \\
    WSD        & 98.6   & 99.4   & 99.1   \\
    \bottomrule
  \end{tabular}
  }
  \caption{Task filtering results showing agreement (i.e., the percentage of instances for which all LLMs produce correct SIP predictions), as well as the maximum and minimum SIP success rates across all LLMs (the actual agreement and minimum rate for \emph{Parity} is 99.96\%).}
  \label{tab:single-filtering}
\end{table}

\section{Experimental Results}
\label{appendix:exp}

Here we present the success rate and failure breakdown for each model and task.

\subsection{Success Rate}
\label{appendix:success_rate_details}
\subsubsection{Task Success Rate for Models}
\label{appendix:success_rate}
\begin{figure}[t]
  \centering
  \setlength{\abovecaptionskip}{2pt}
  \setlength{\belowcaptionskip}{0pt}

  \includegraphics[width=0.9\linewidth]{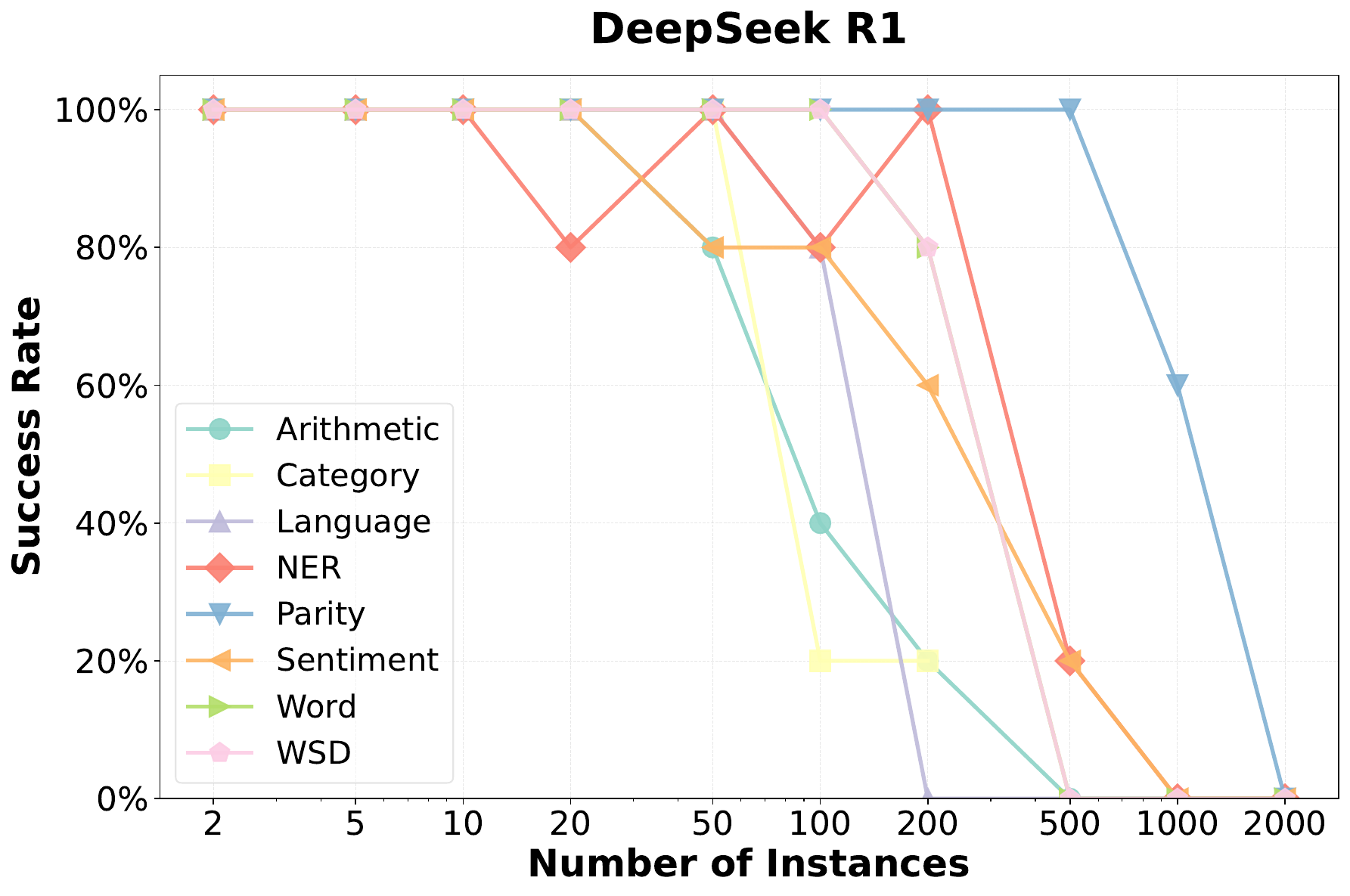}
  \includegraphics[width=0.9\linewidth]{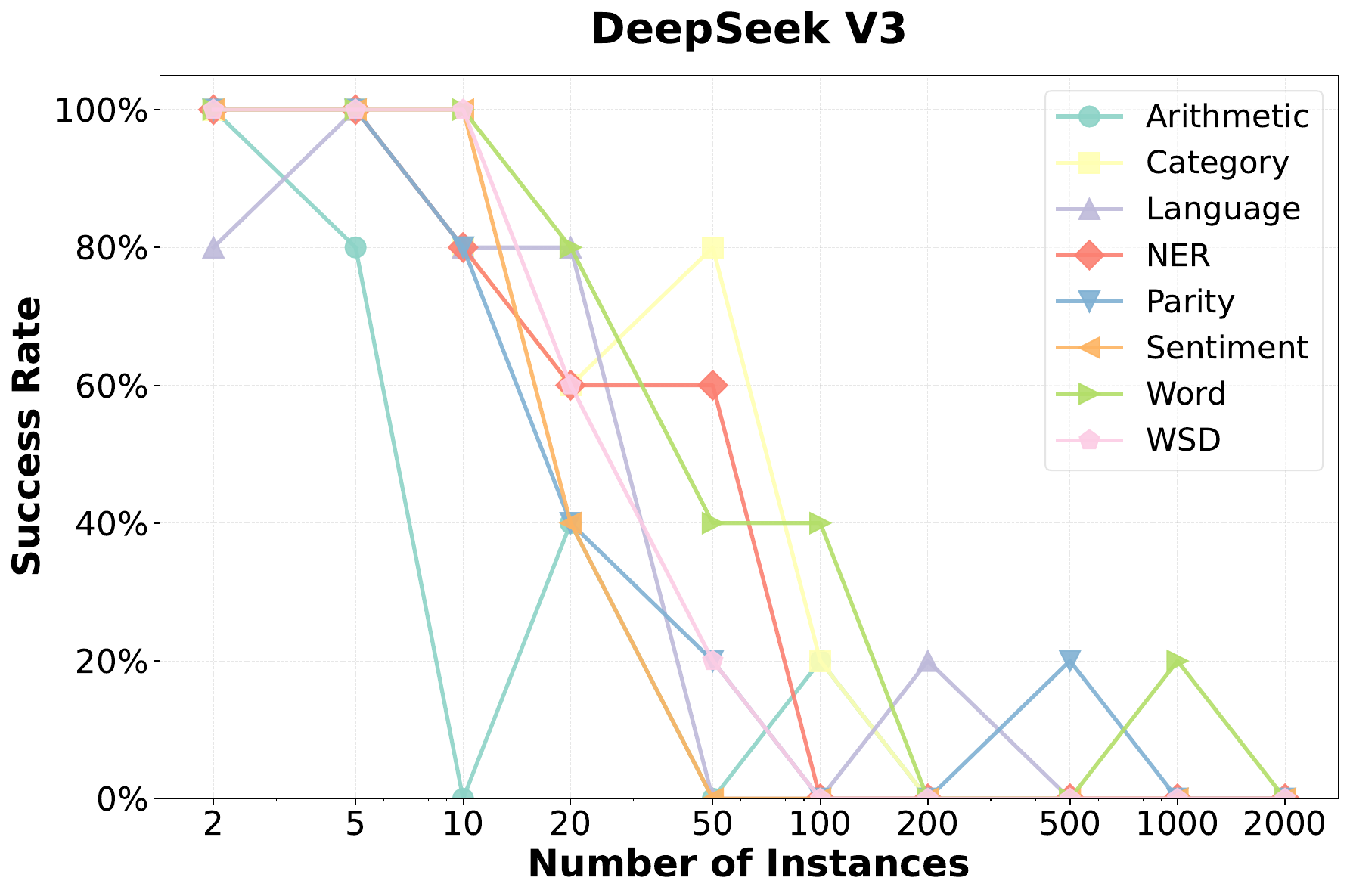}
  \includegraphics[width=0.9\linewidth]{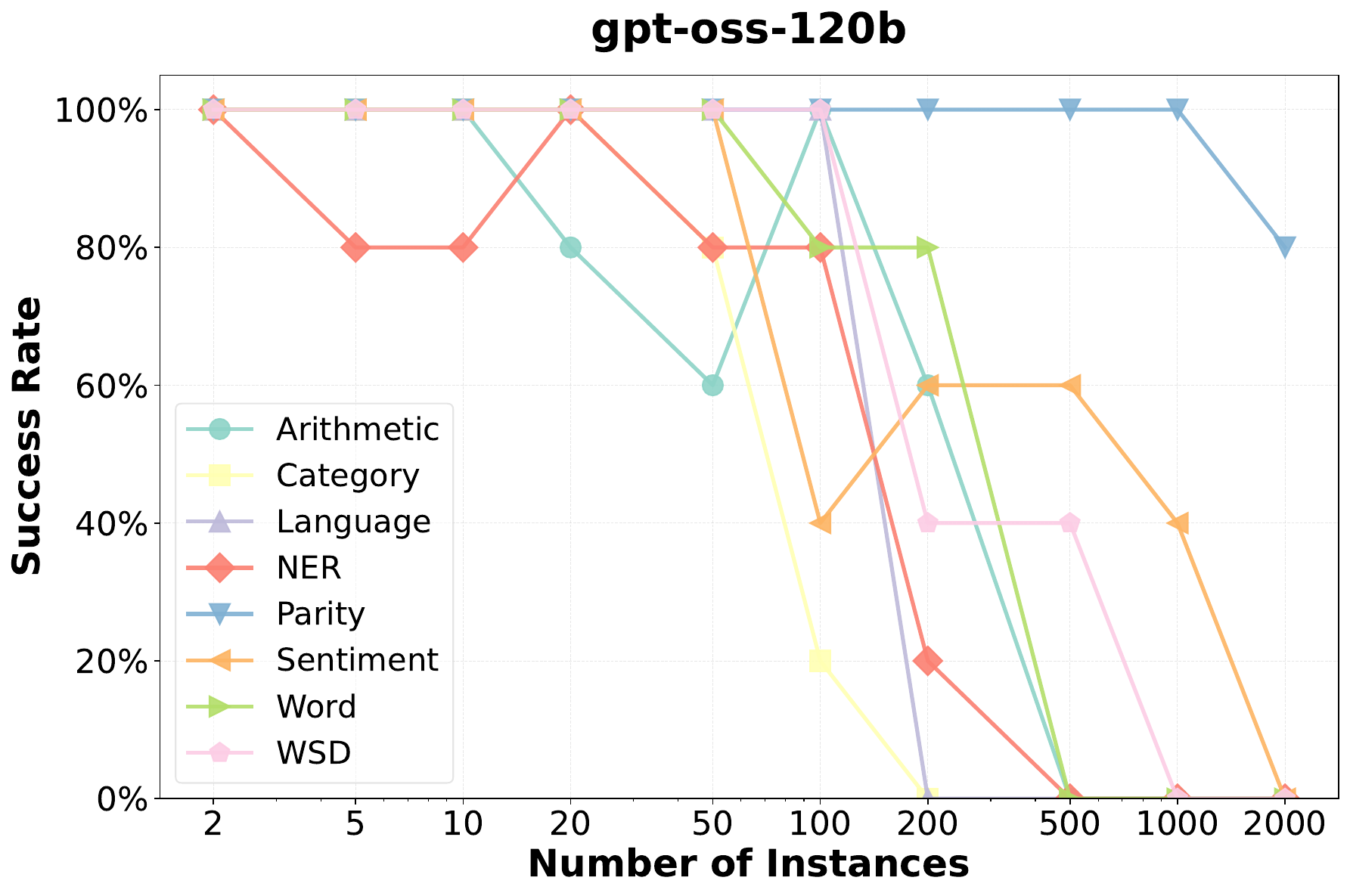}
  \includegraphics[width=0.9\linewidth]{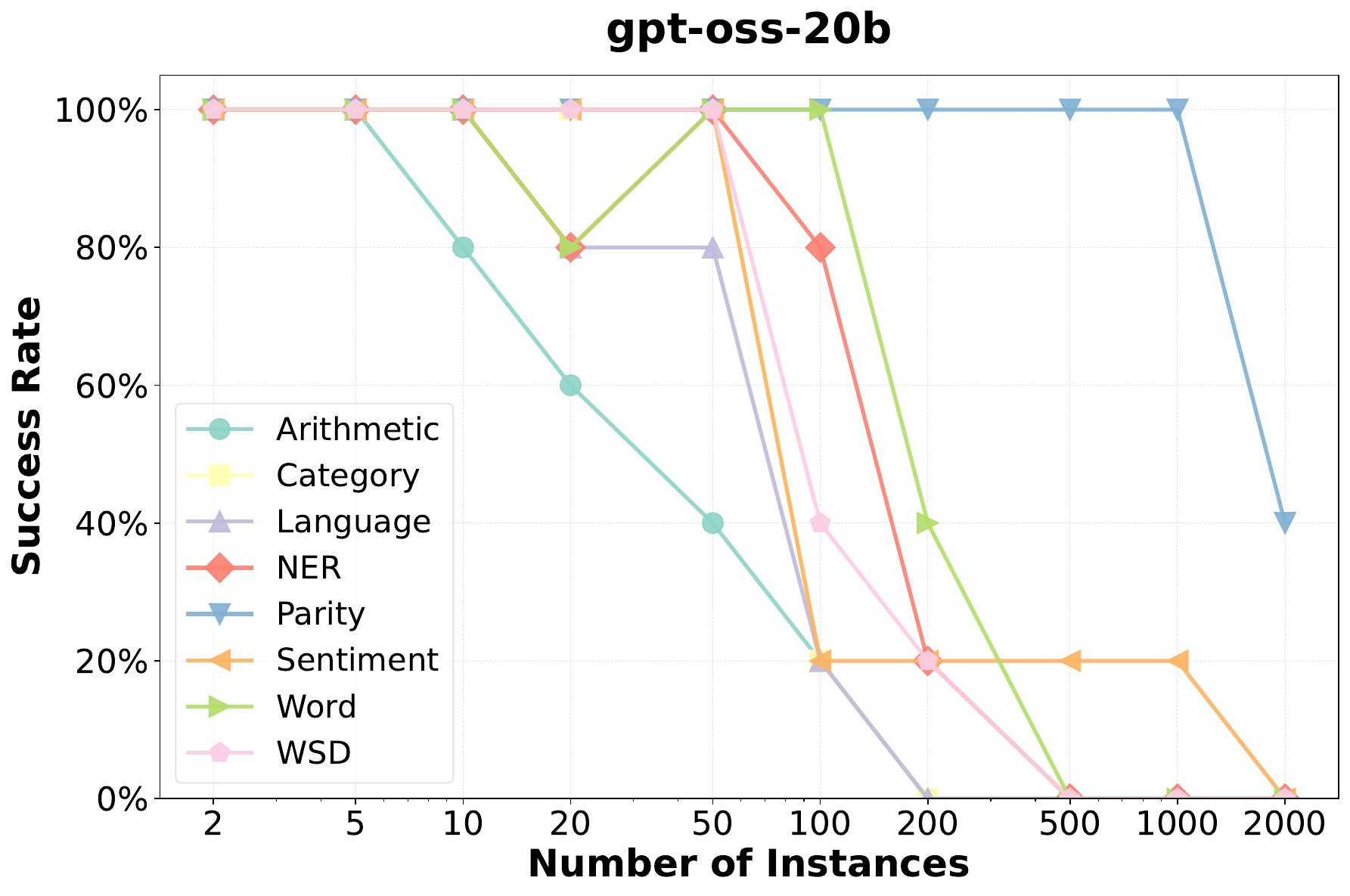}

  \caption{Success rate of models.}
  \label{fig:success_rate_models_1}
\end{figure}

\begin{figure}[t]
  \centering
  \setlength{\abovecaptionskip}{2pt}
  \setlength{\belowcaptionskip}{0pt}

  \includegraphics[width=0.9\linewidth]{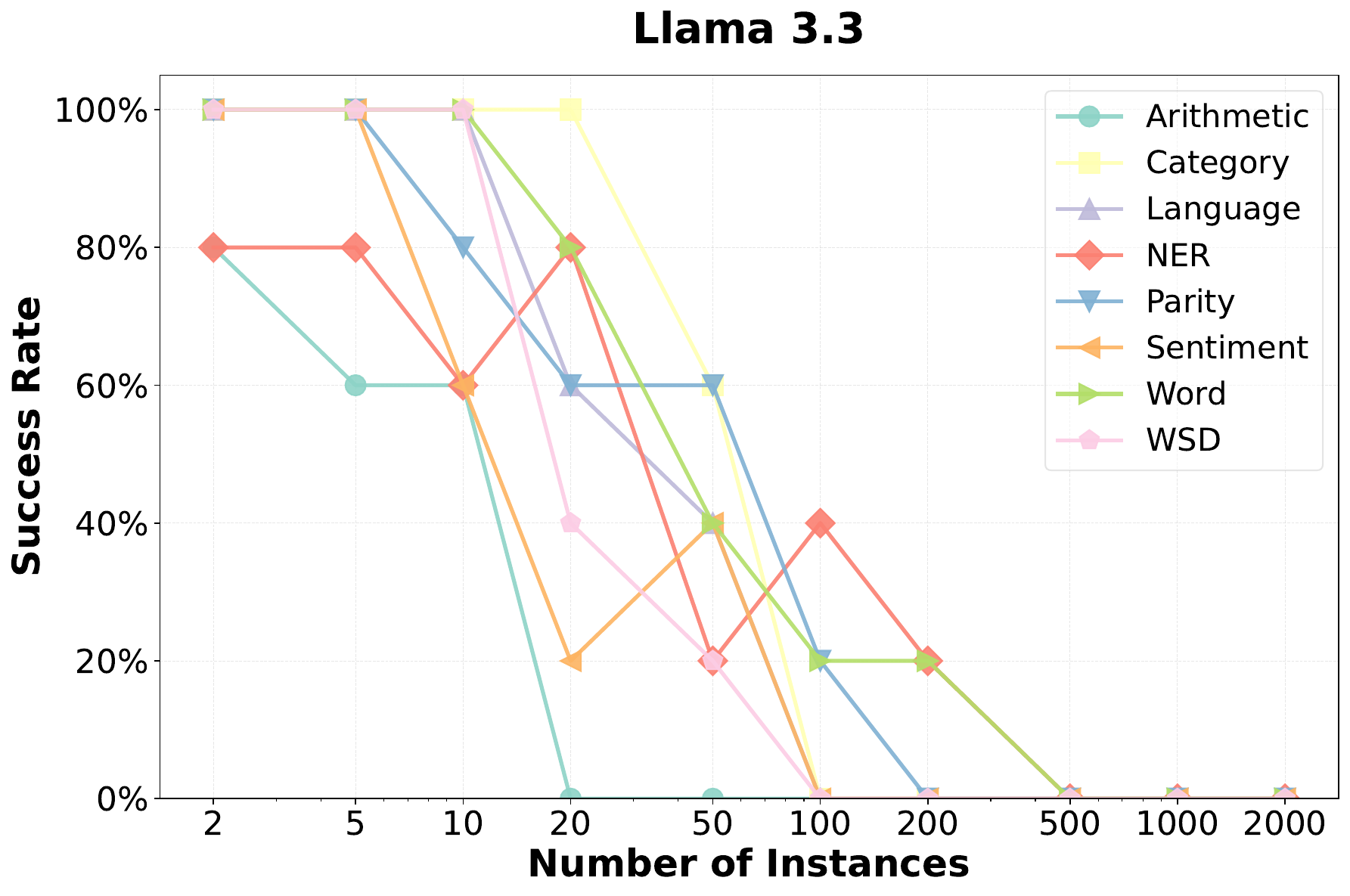}
  \includegraphics[width=0.9\linewidth]{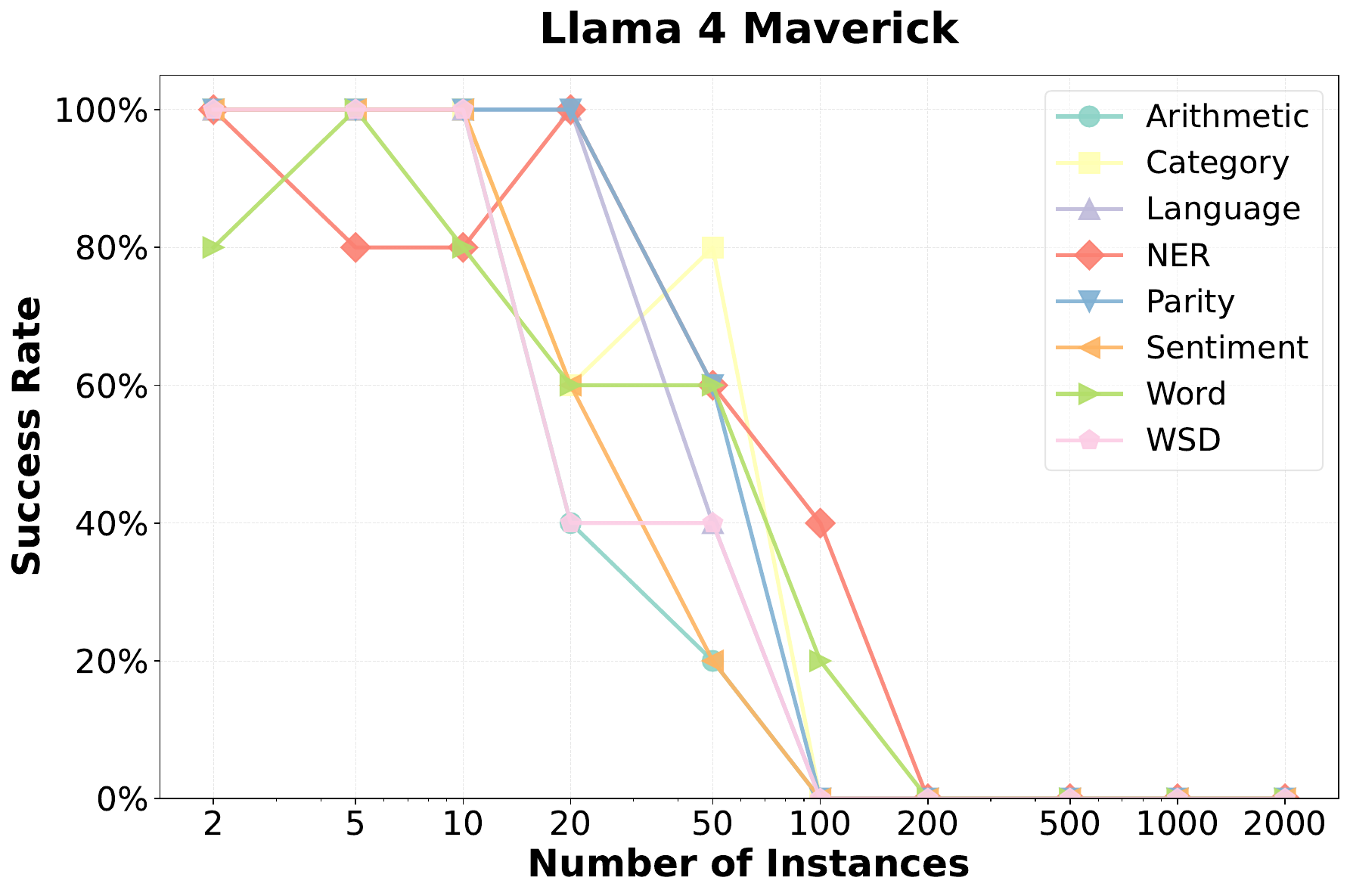}
  \includegraphics[width=0.9\linewidth]{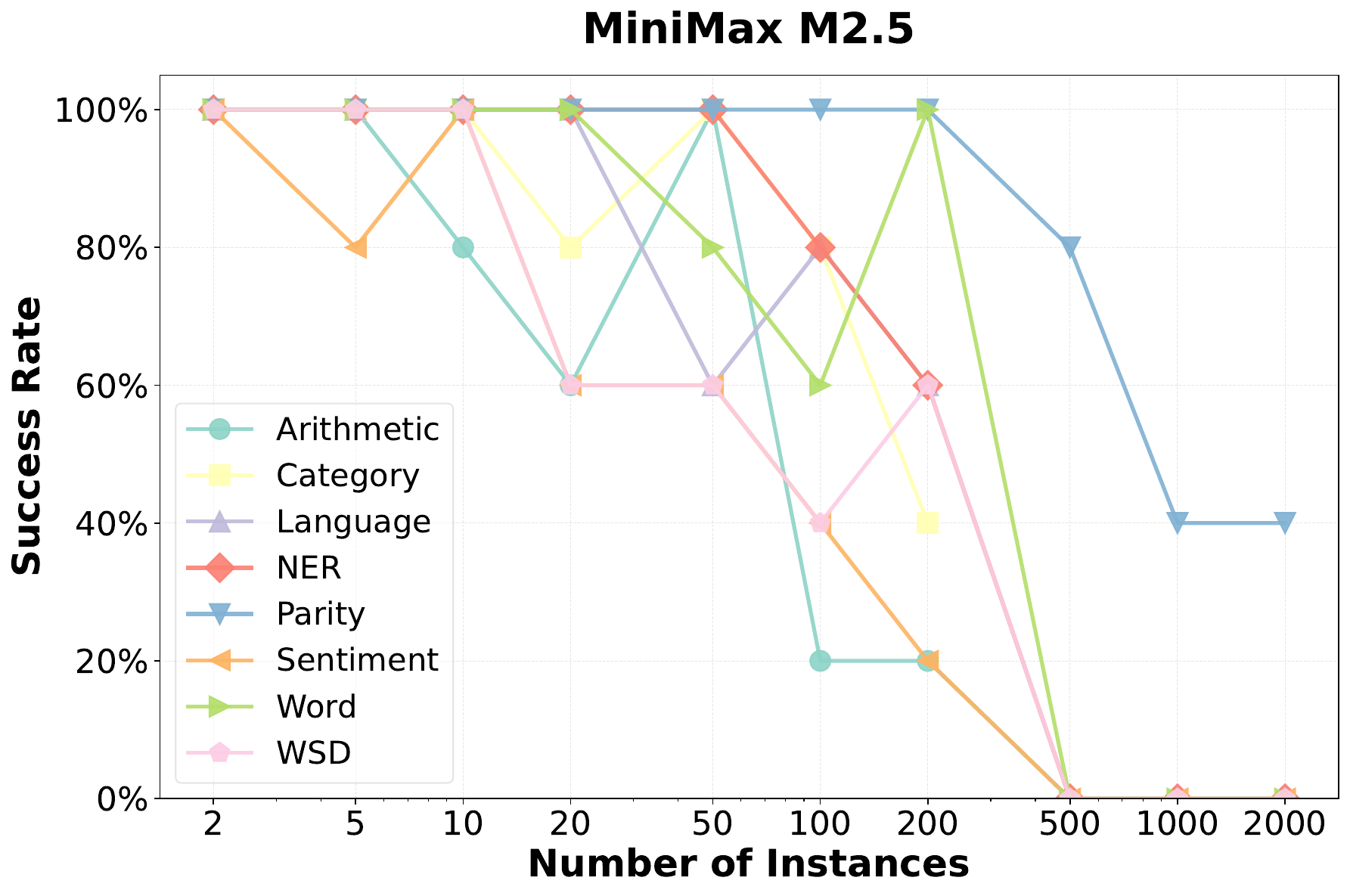}
  \includegraphics[width=0.9\linewidth]{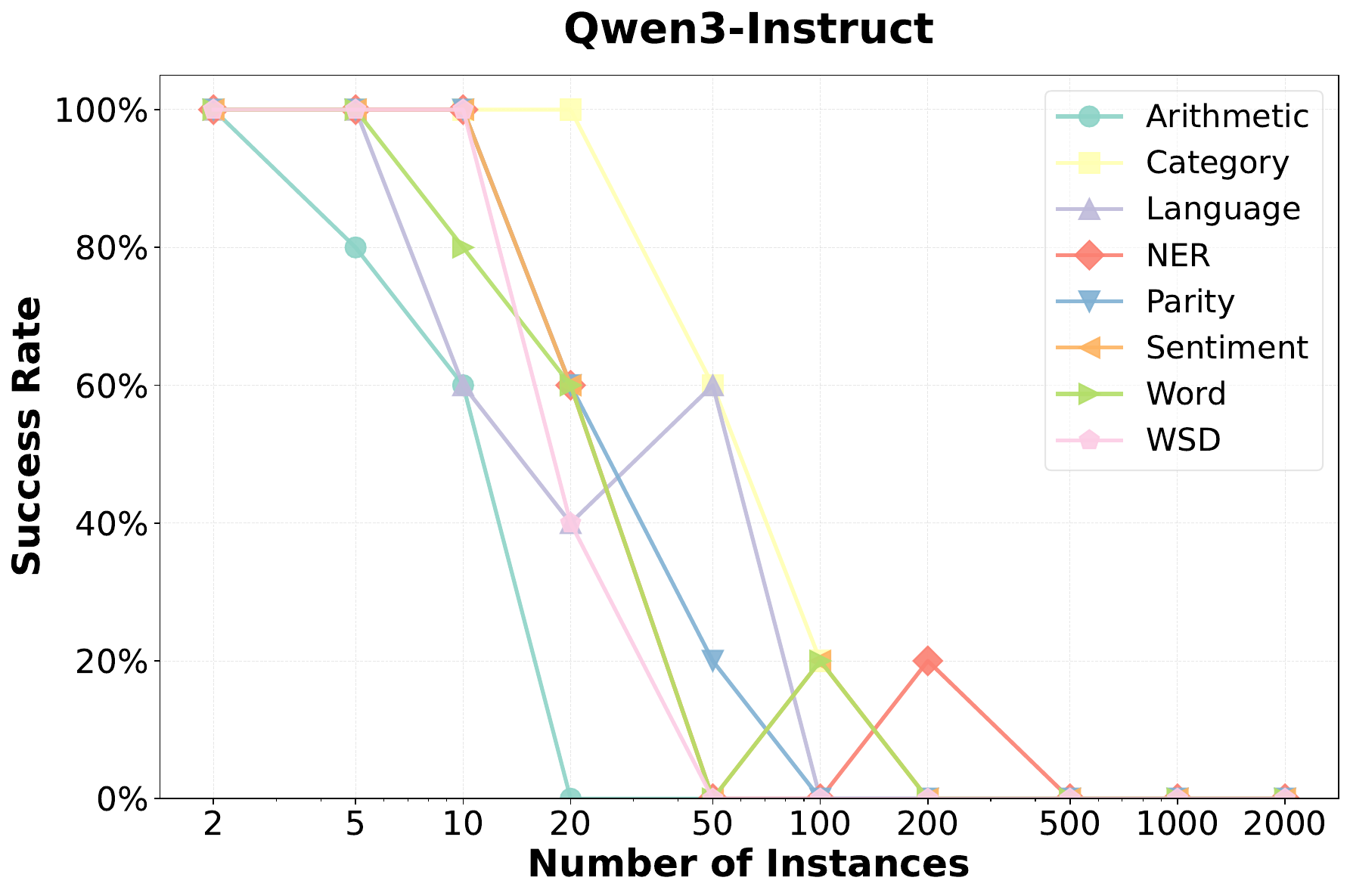}
  \caption{Success rate of models.}
  \label{fig:success_rate_models_2}
\end{figure}

\begin{figure}[t]
  \centering
  \setlength{\abovecaptionskip}{2pt}
  \setlength{\belowcaptionskip}{0pt}

  \includegraphics[width=0.9\linewidth]{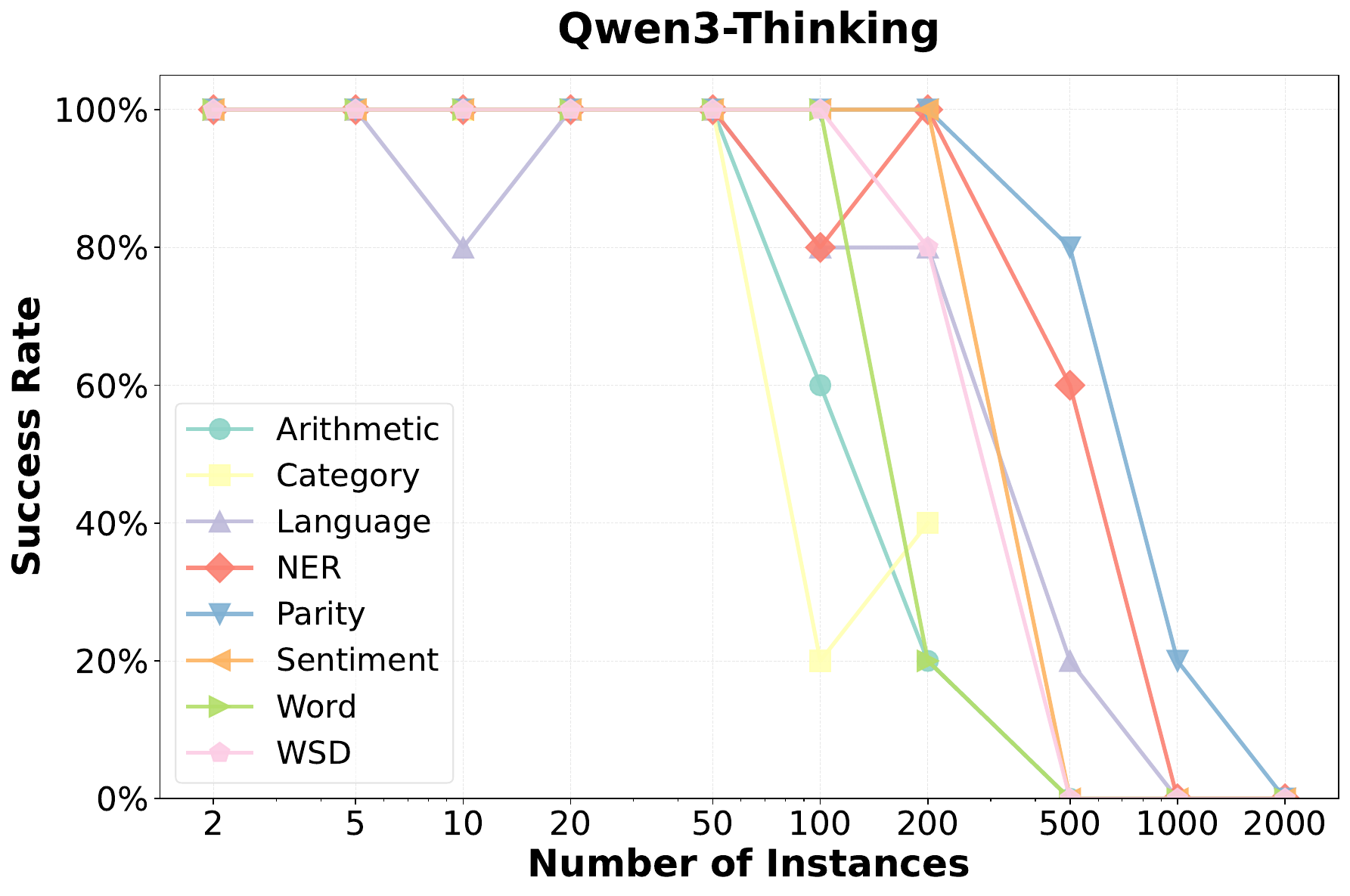}
  \includegraphics[width=0.9\linewidth]{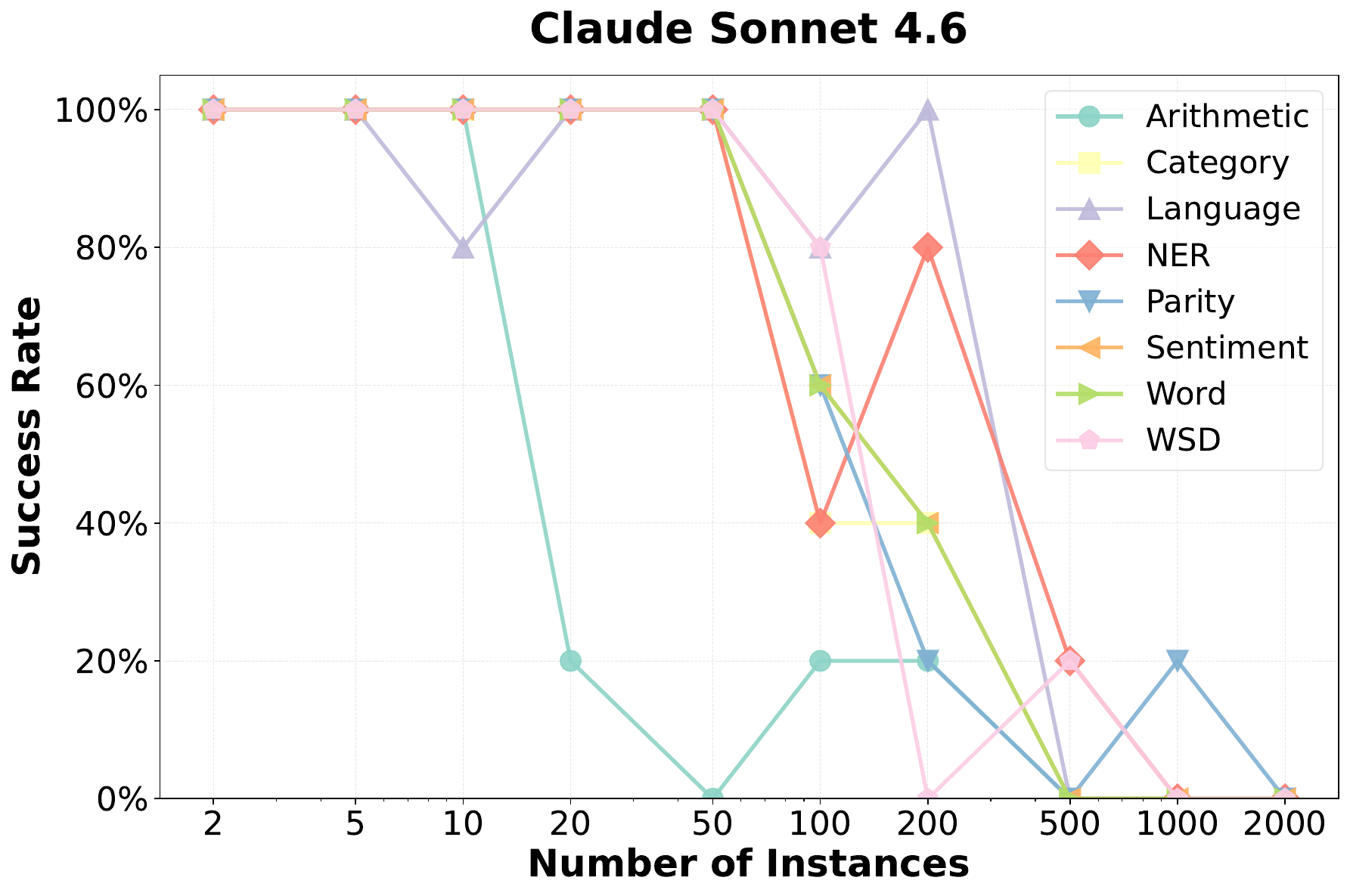}
  \includegraphics[width=0.9\linewidth]{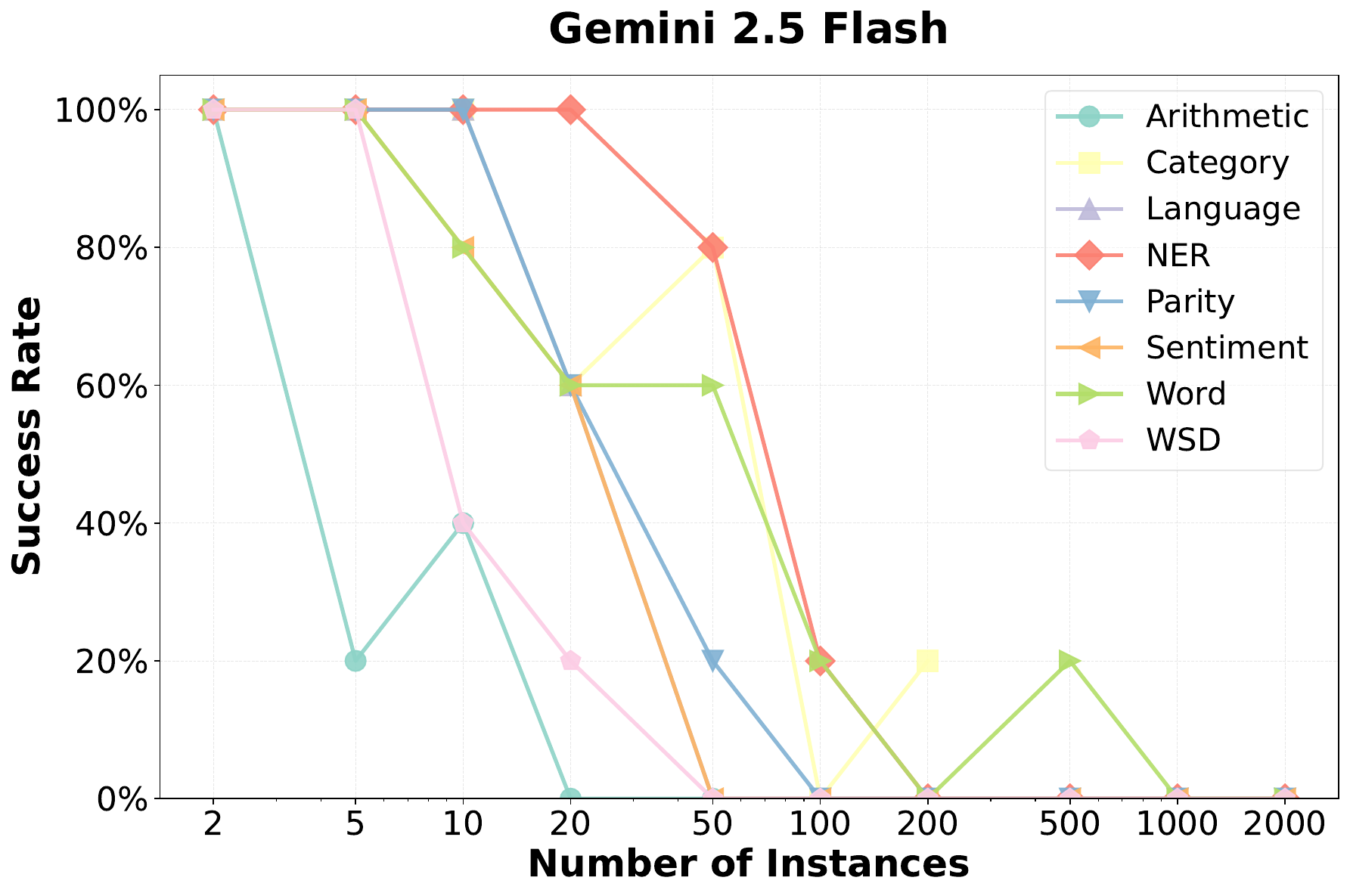}
  \includegraphics[width=0.9\linewidth]{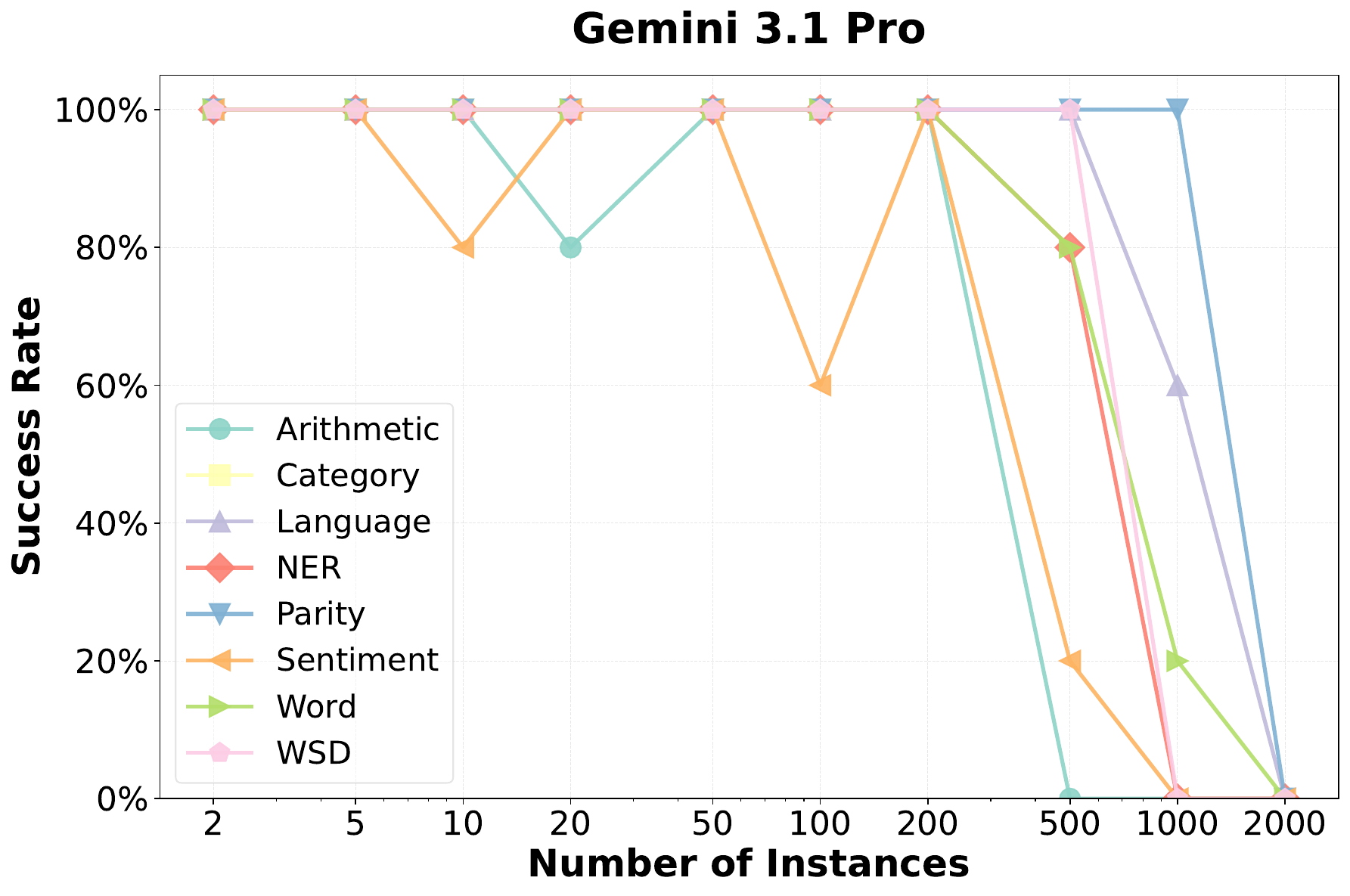}

  \caption{Success rate of models.}
  \label{fig:success_rate_models_3}
\end{figure}

\begin{figure}[t]
  \centering
  \setlength{\abovecaptionskip}{2pt}
  \setlength{\belowcaptionskip}{0pt}

  \includegraphics[width=0.9\linewidth]{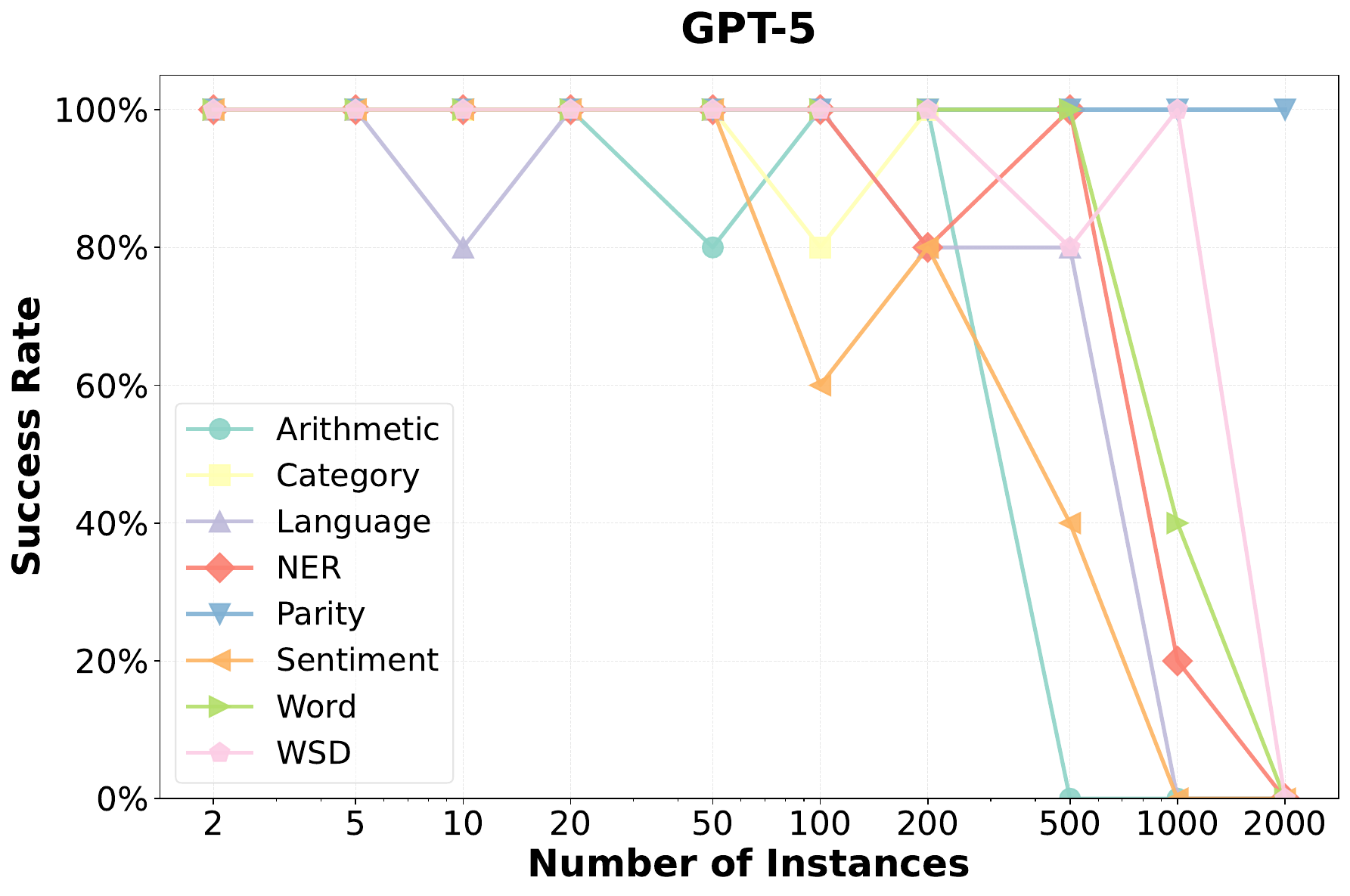}
  \includegraphics[width=0.9\linewidth]{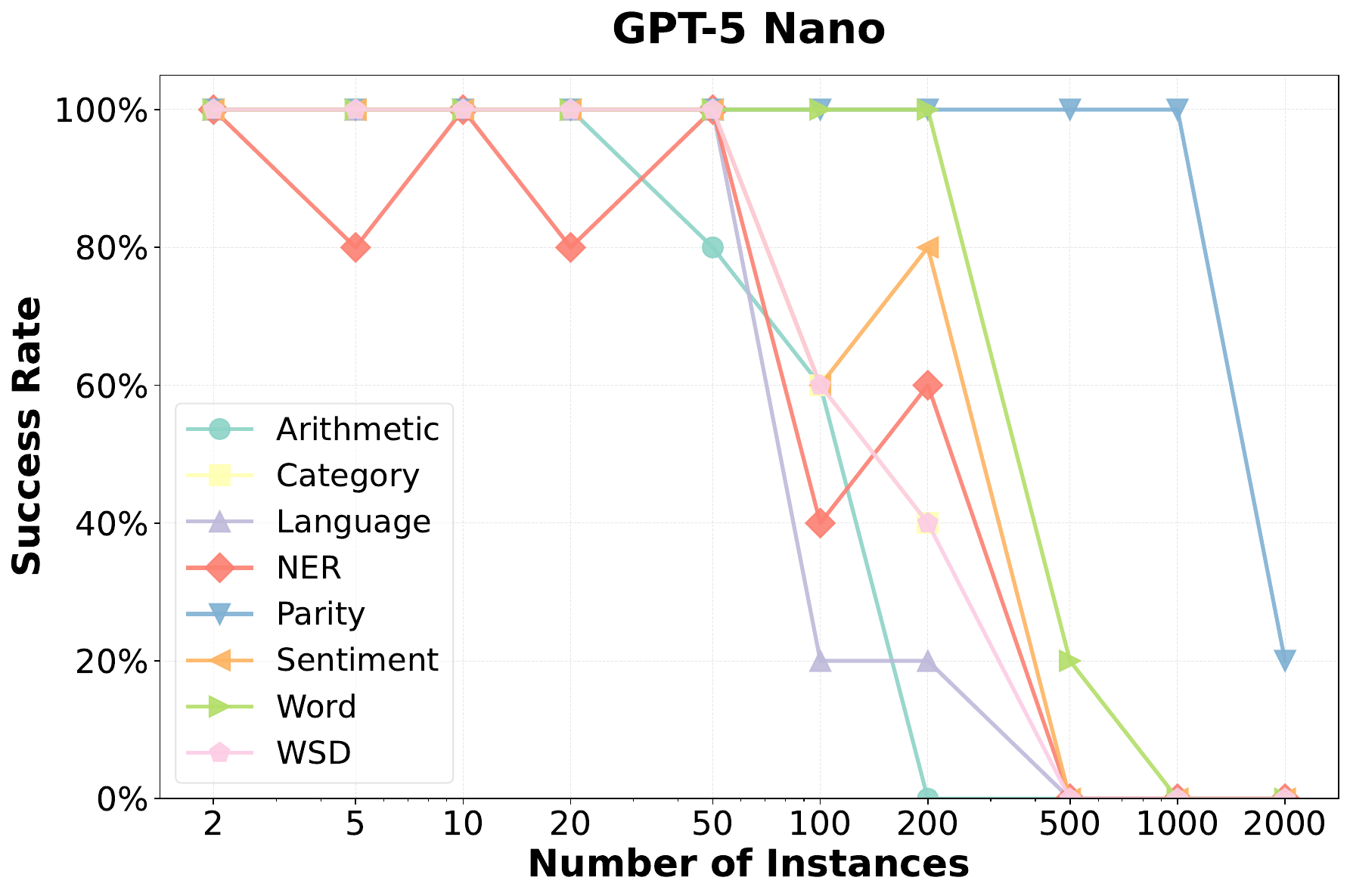}
    \includegraphics[width=0.9\linewidth]{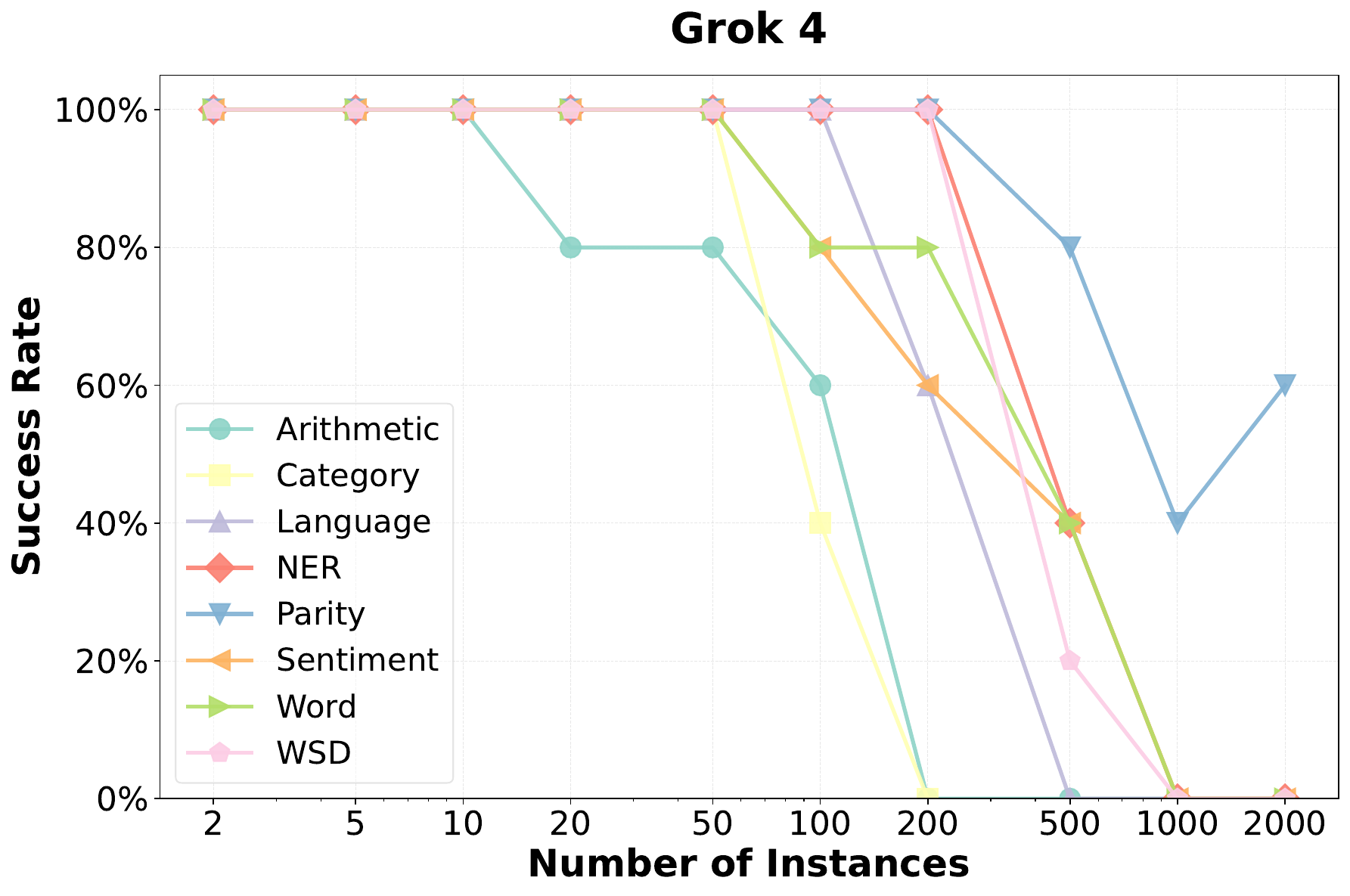}
  \includegraphics[width=0.9\linewidth]{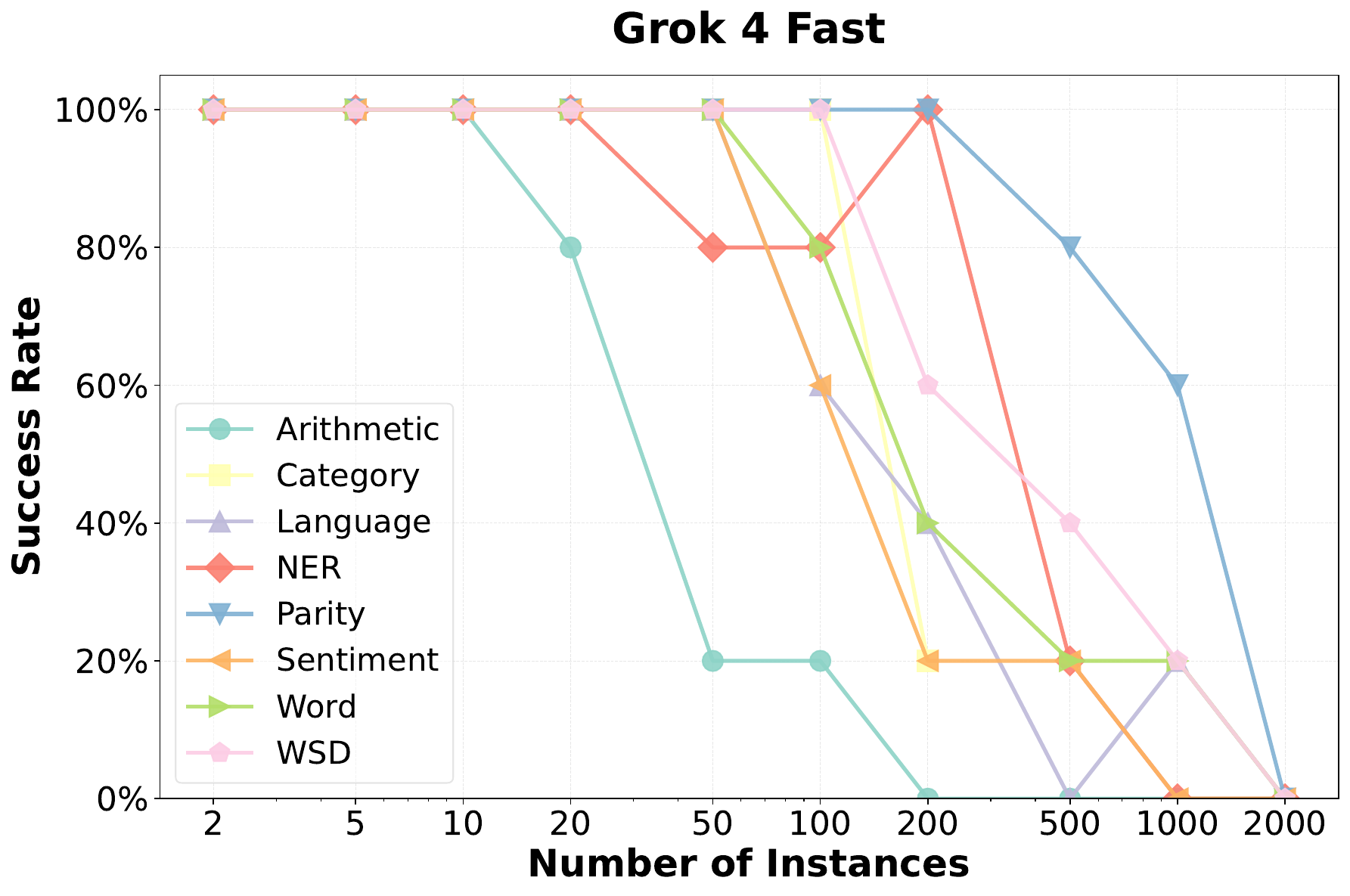}

  \caption{Success rate of models.}
  \label{fig:success_rate_models_4}
\end{figure}

Figure~\ref{fig:success_rate_models_1}, Figure~\ref{fig:success_rate_models_2}, Figure~\ref{fig:success_rate_models_3} and Figure~\ref{fig:success_rate_models_4} show task success rate for each model.

\subsubsection{Model Success Rate for Tasks}
\label{appendix:model_success_rate_tasks}
\begin{figure}[t]
  \centering
  \setlength{\abovecaptionskip}{2pt}
  \setlength{\belowcaptionskip}{0pt}

  \includegraphics[width=0.9\linewidth]{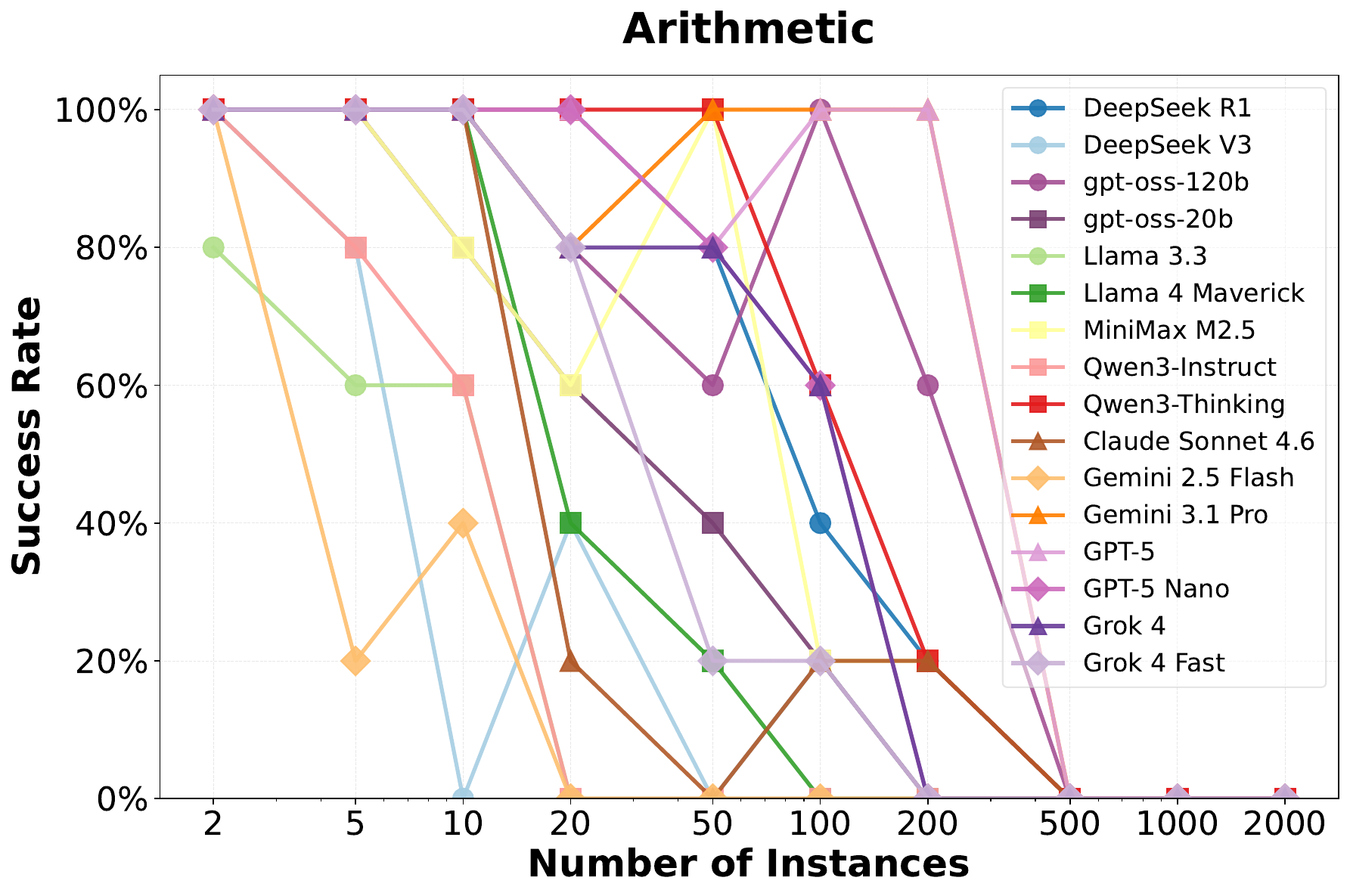}
  \includegraphics[width=0.9\linewidth]{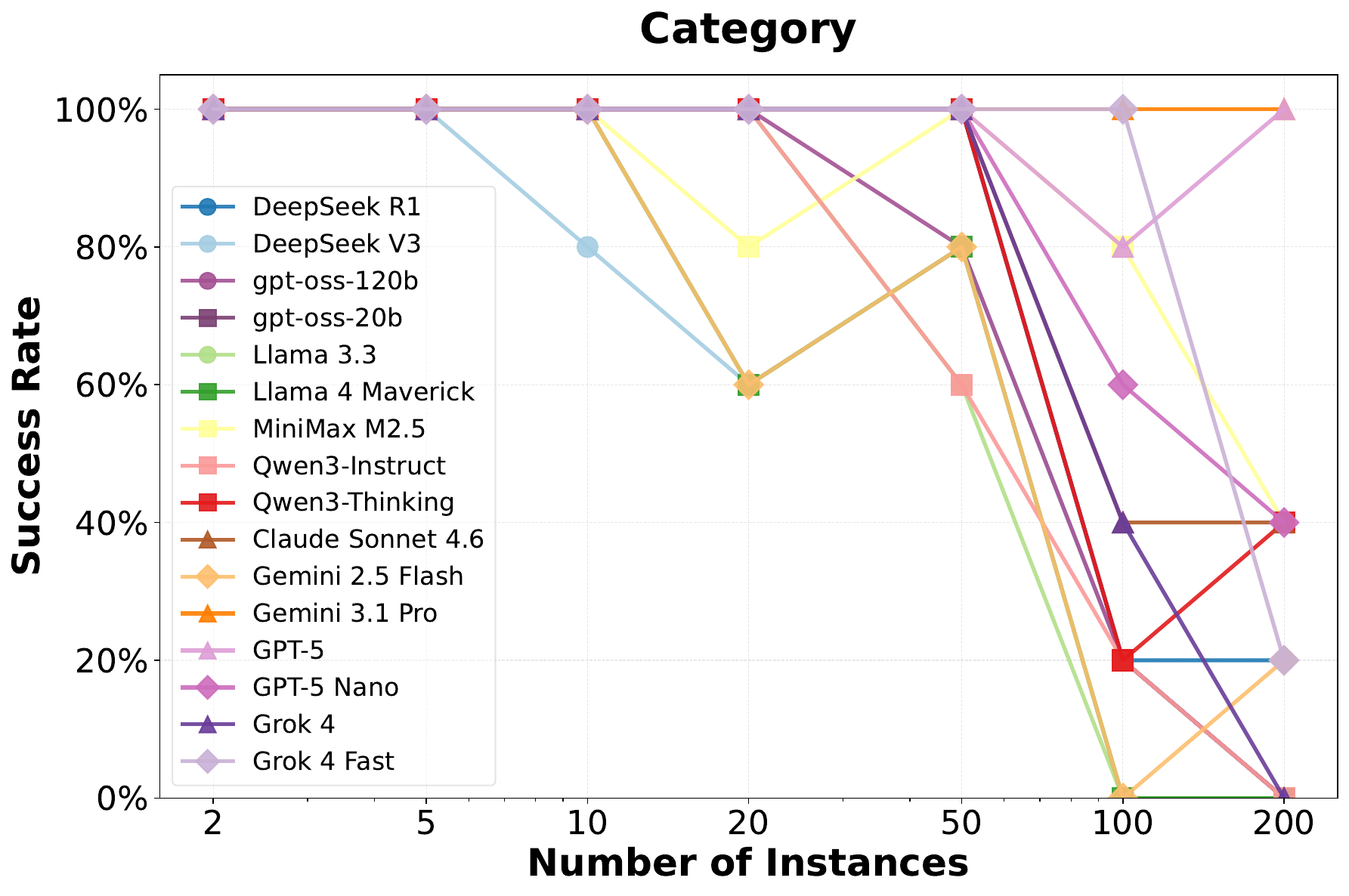}
  \includegraphics[width=0.9\linewidth]{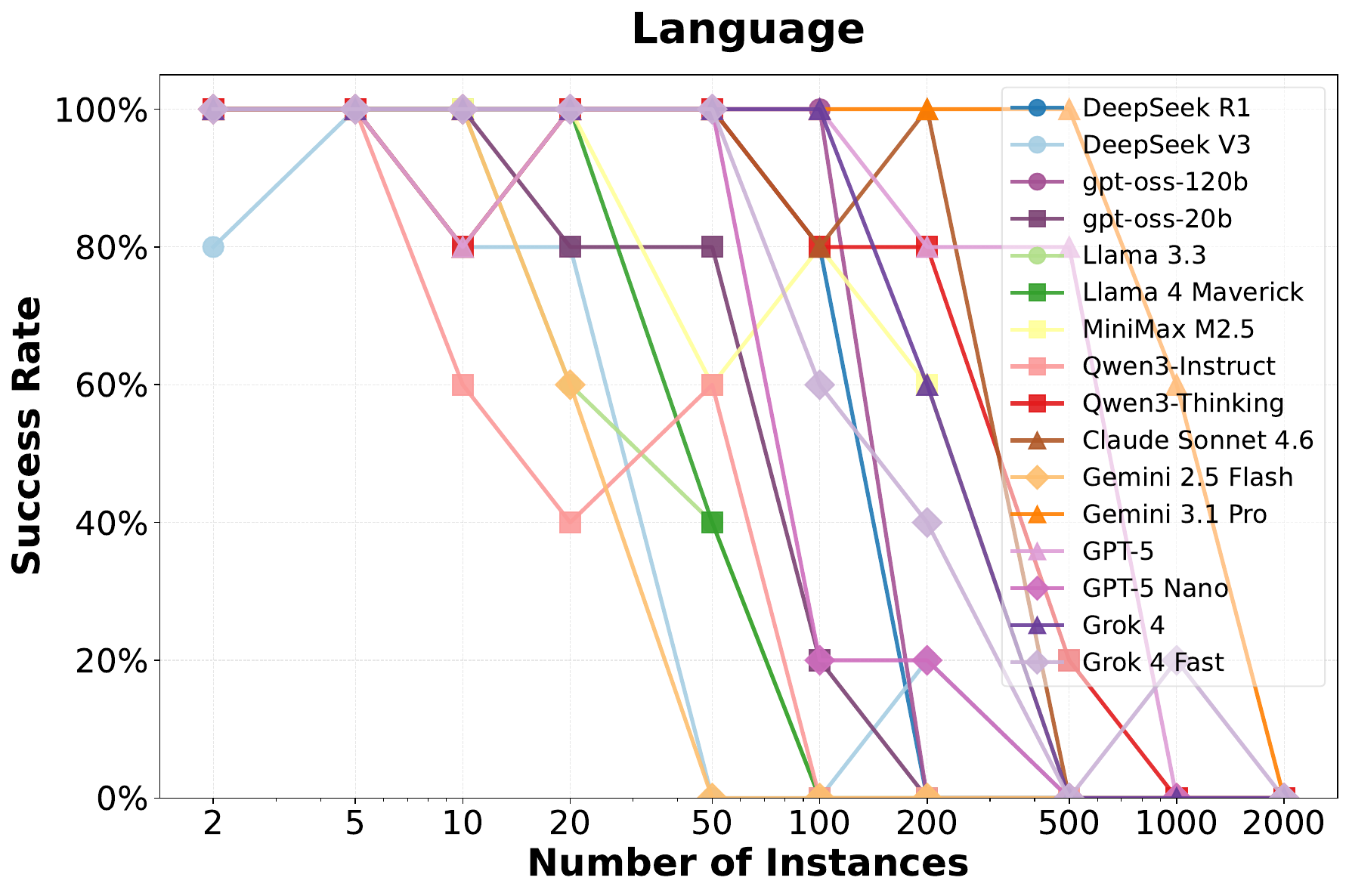}
  \includegraphics[width=0.9\linewidth]{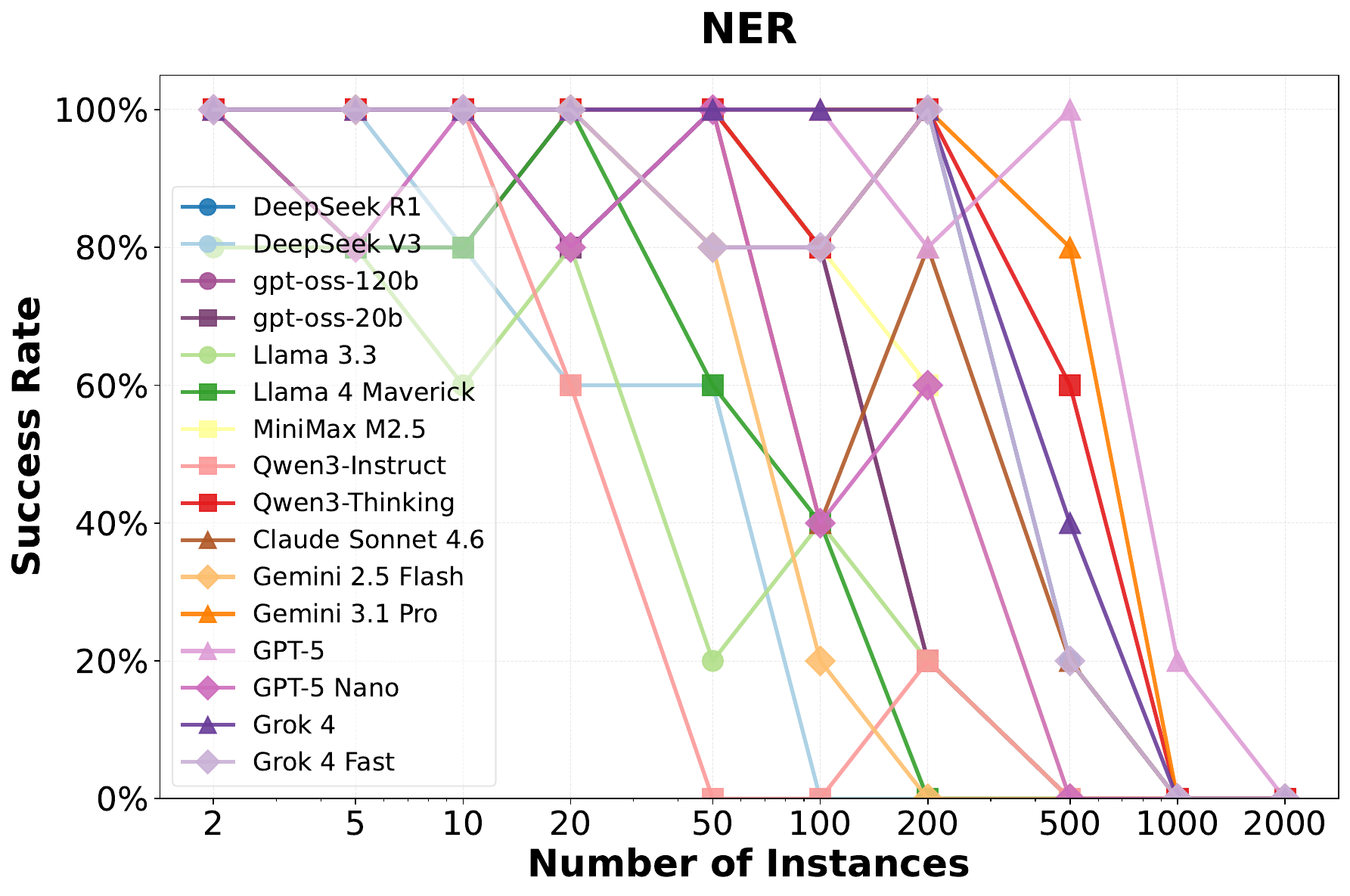}

  \caption{Model success rate for tasks.}
  \label{fig:success_rate_tasks_1}
\end{figure}

\begin{figure}[t]
  \centering
  \setlength{\abovecaptionskip}{2pt}
  \setlength{\belowcaptionskip}{0pt}

  \includegraphics[width=0.9\linewidth]{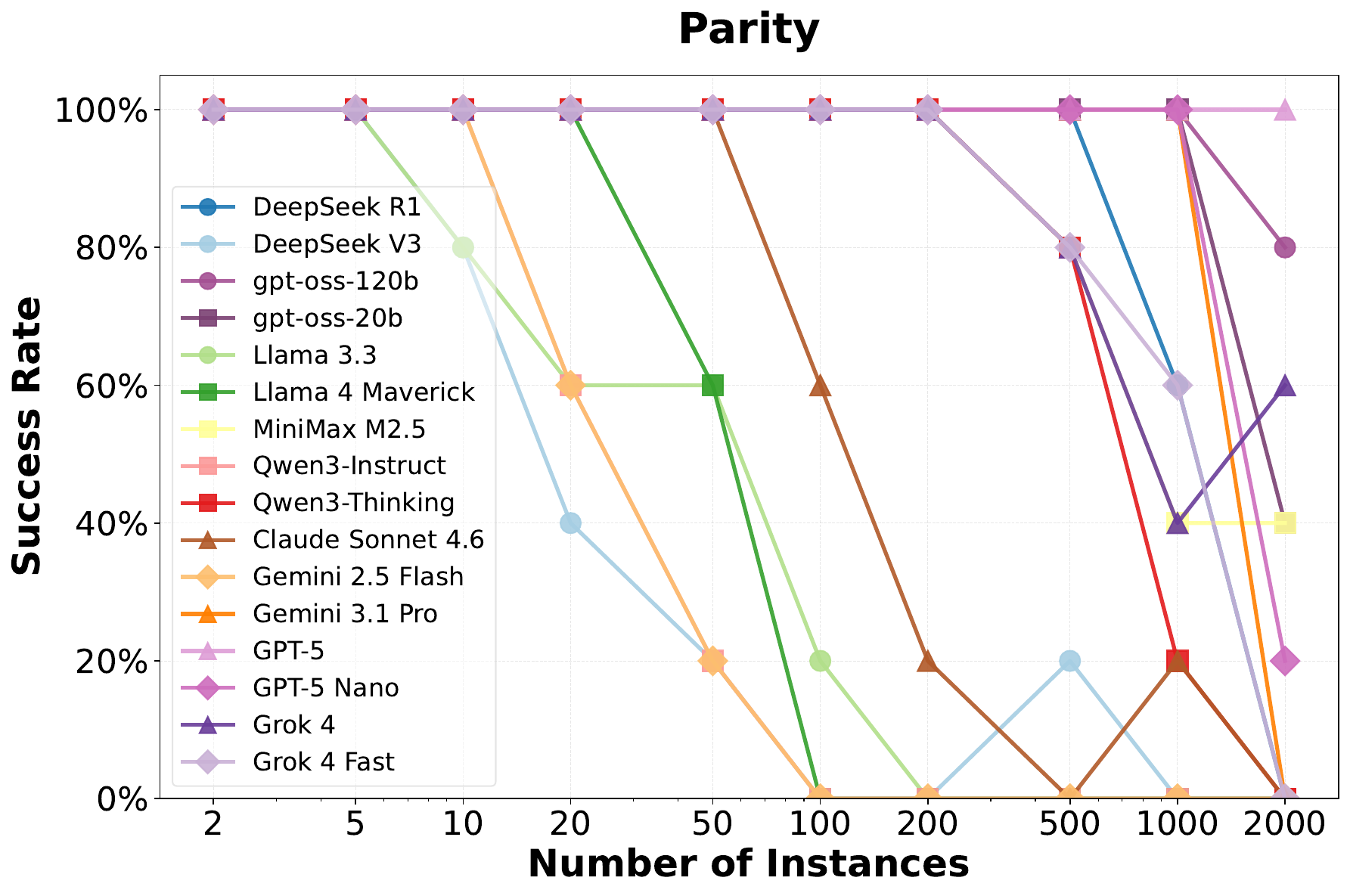}
  \includegraphics[width=0.9\linewidth]{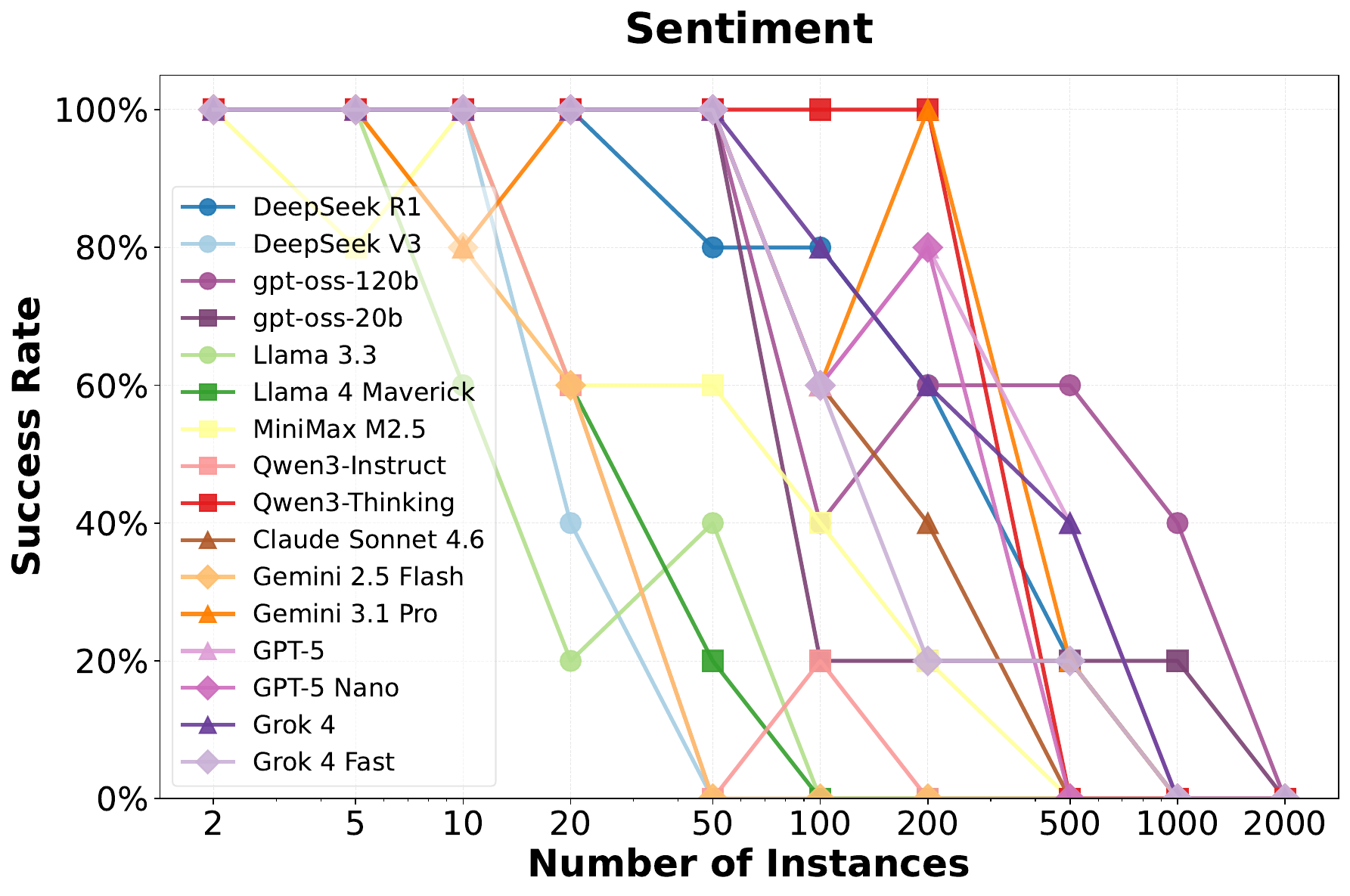}
  \includegraphics[width=0.9\linewidth]{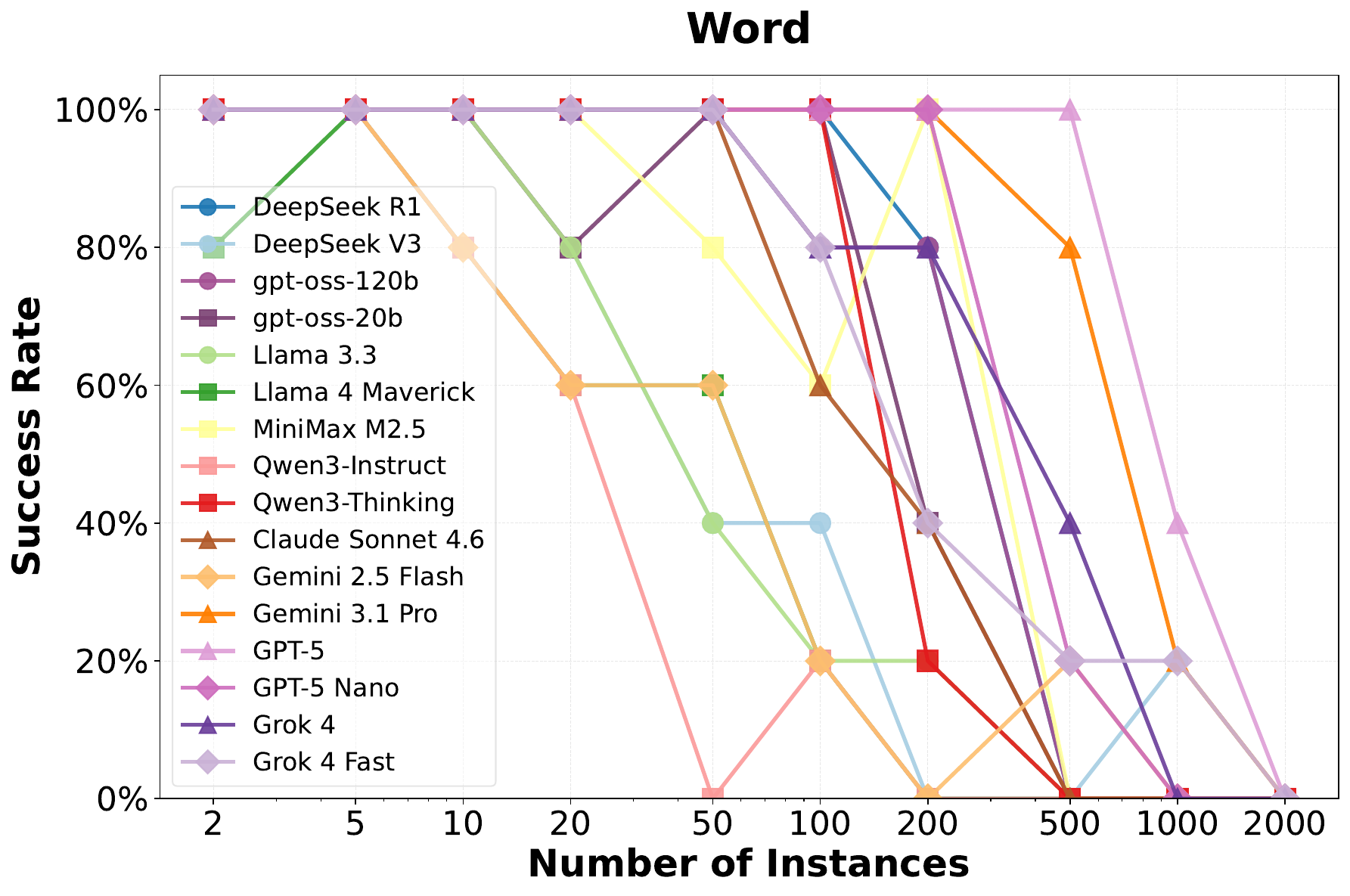}
  \includegraphics[width=0.9\linewidth]{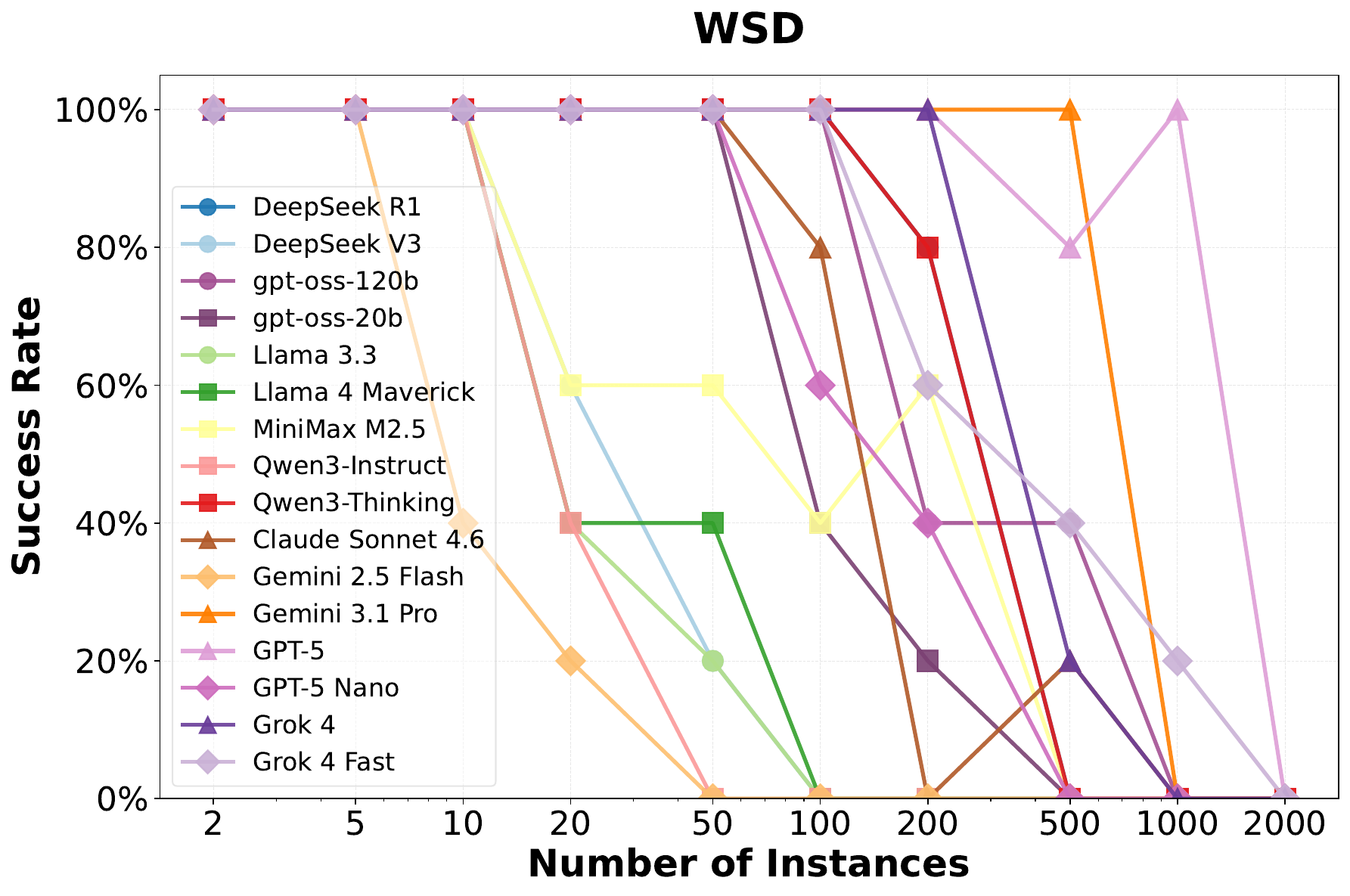}

  \caption{Model success rate for tasks.}
  \label{fig:success_rate_tasks_2}
\end{figure}

Figure~\ref{fig:success_rate_tasks_1} and Figure~\ref{fig:success_rate_tasks_2} show model success rate for each task.

\subsection{Failure Breakdown}
\label{appendix:failure_breakdown_details}
\subsubsection{Failure Breakdown for Models}
\label{appendix:error_breakdown_model}
\begin{figure}[t]
  \centering
  \setlength{\abovecaptionskip}{2pt}
  \setlength{\belowcaptionskip}{0pt}

  \includegraphics[width=0.9\linewidth]{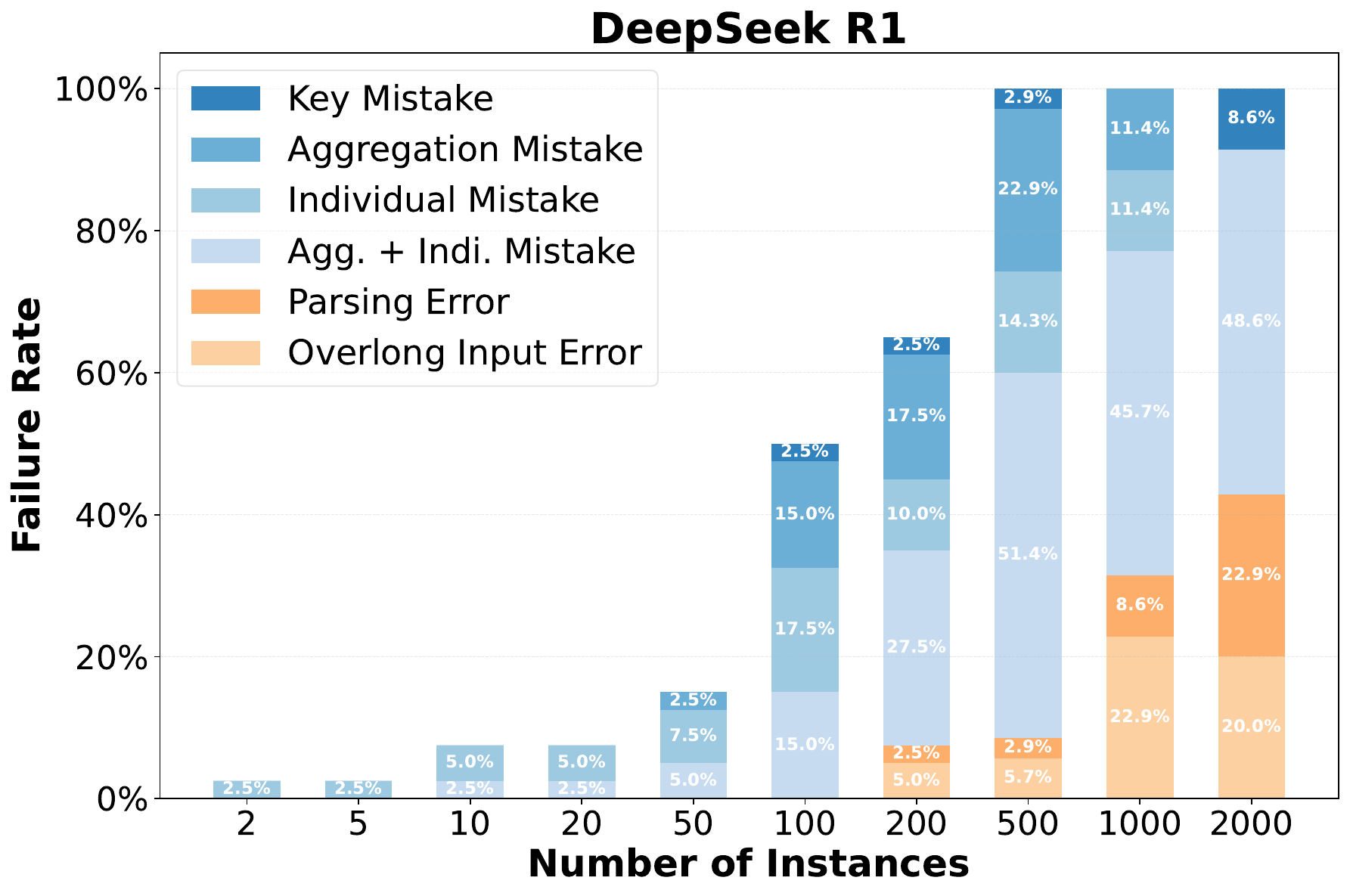}
  \includegraphics[width=0.9\linewidth]{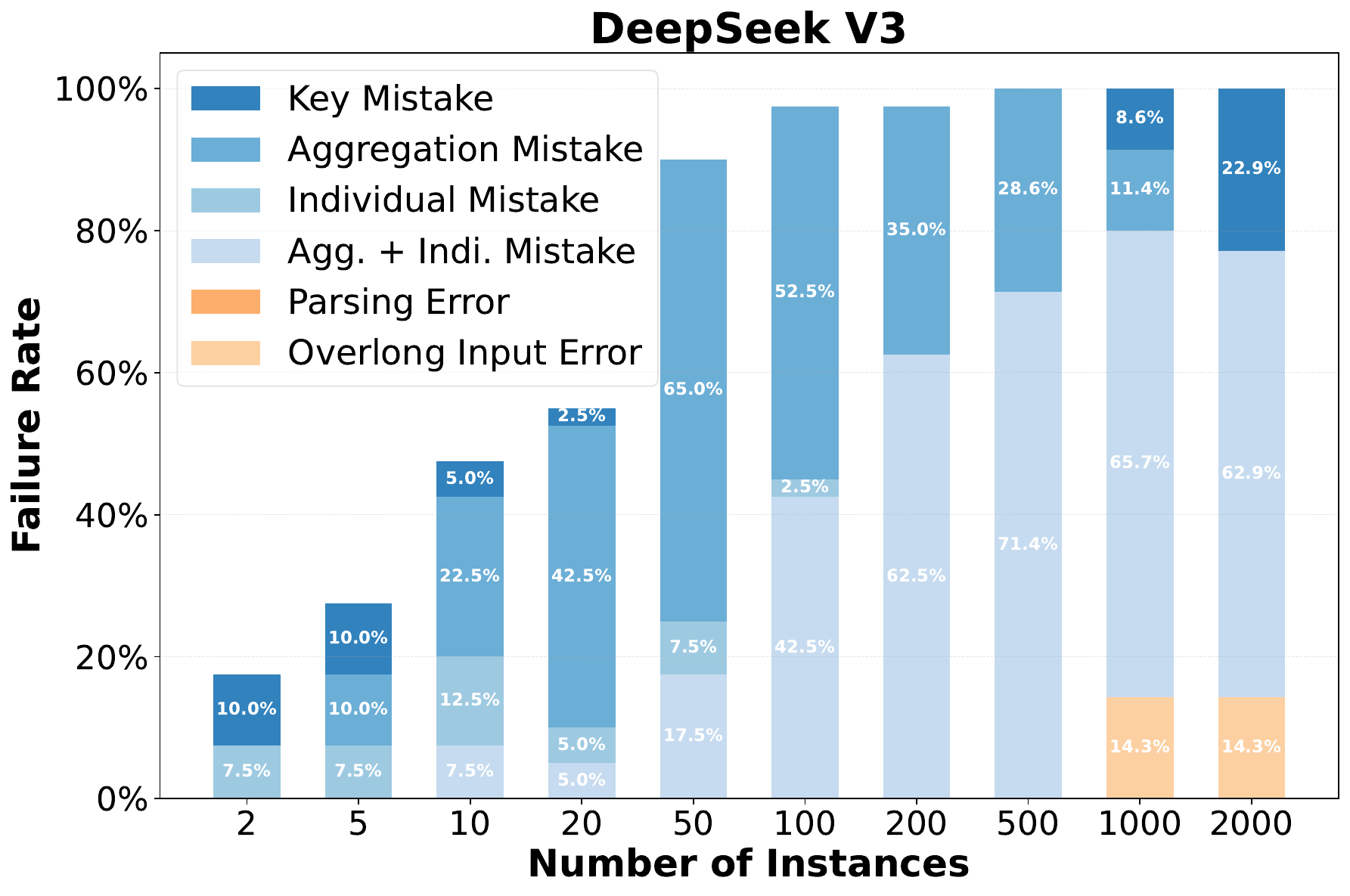}
  \includegraphics[width=0.9\linewidth]{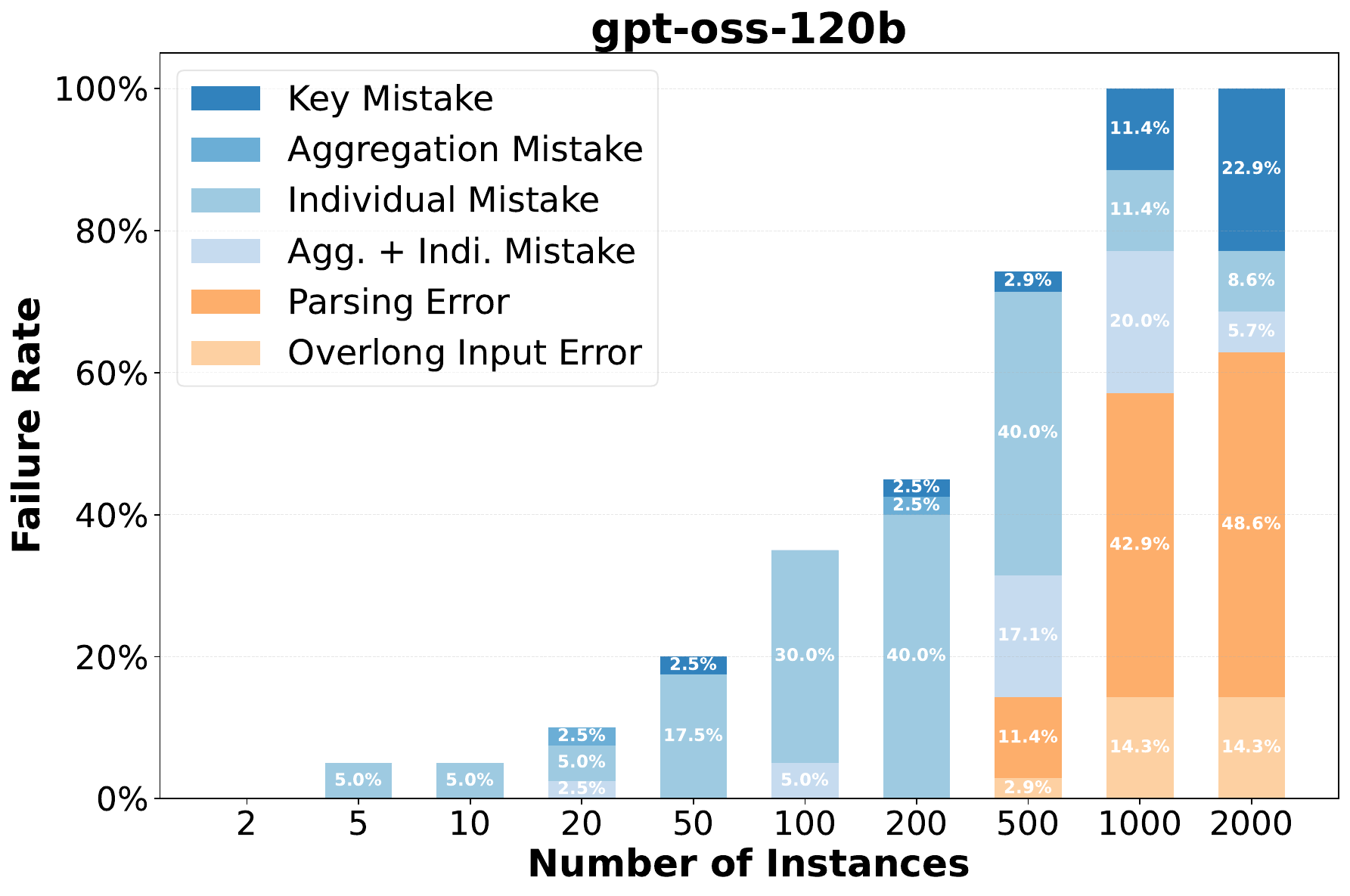}
  \includegraphics[width=0.9\linewidth]{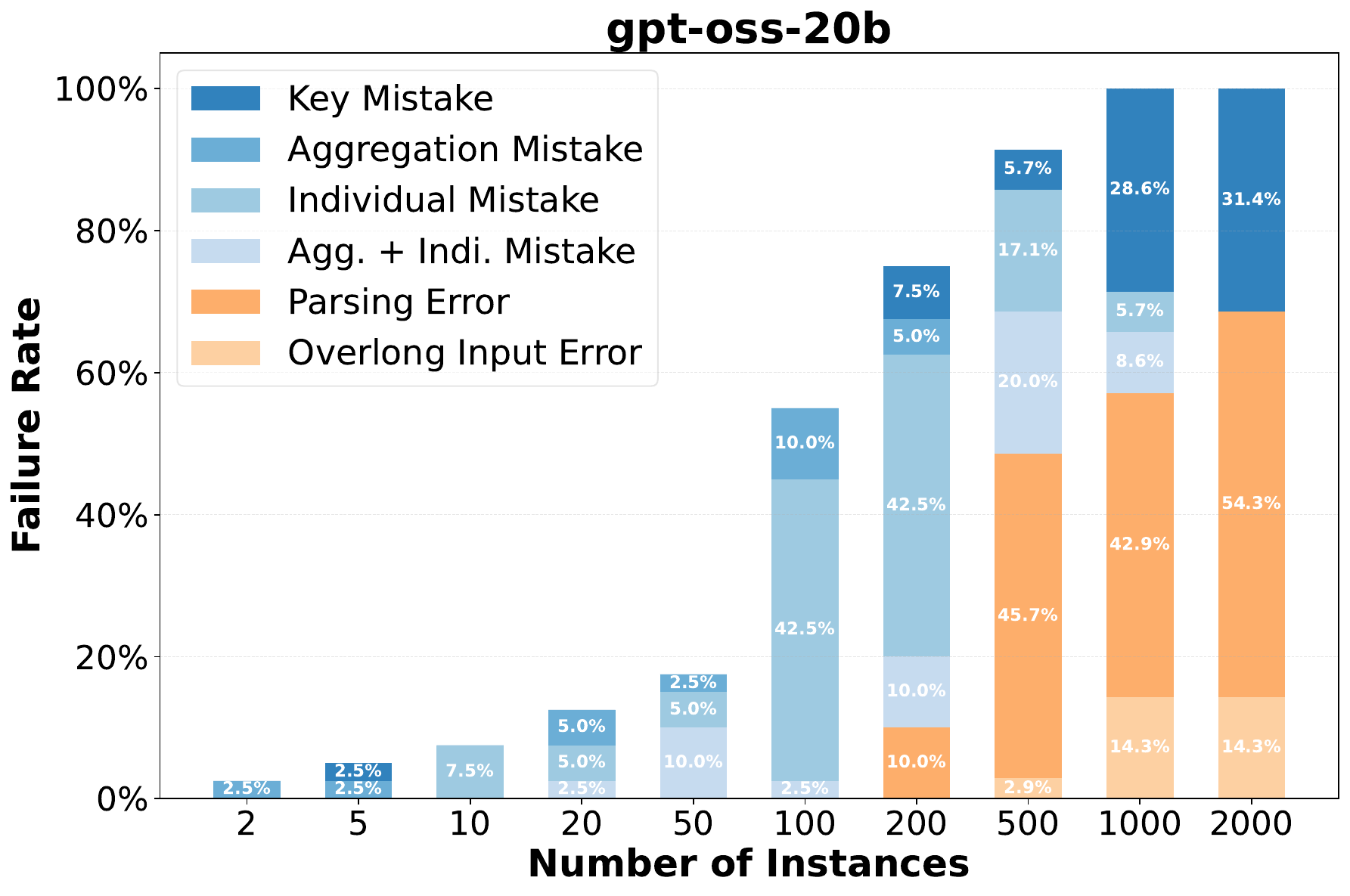}

  \caption{Failure breakdown for models.}
  \label{fig:error_breakdown_models_1}
\end{figure}

\begin{figure}[t]
  \centering
  \setlength{\abovecaptionskip}{2pt}
  \setlength{\belowcaptionskip}{0pt}

  \includegraphics[width=0.9\linewidth]{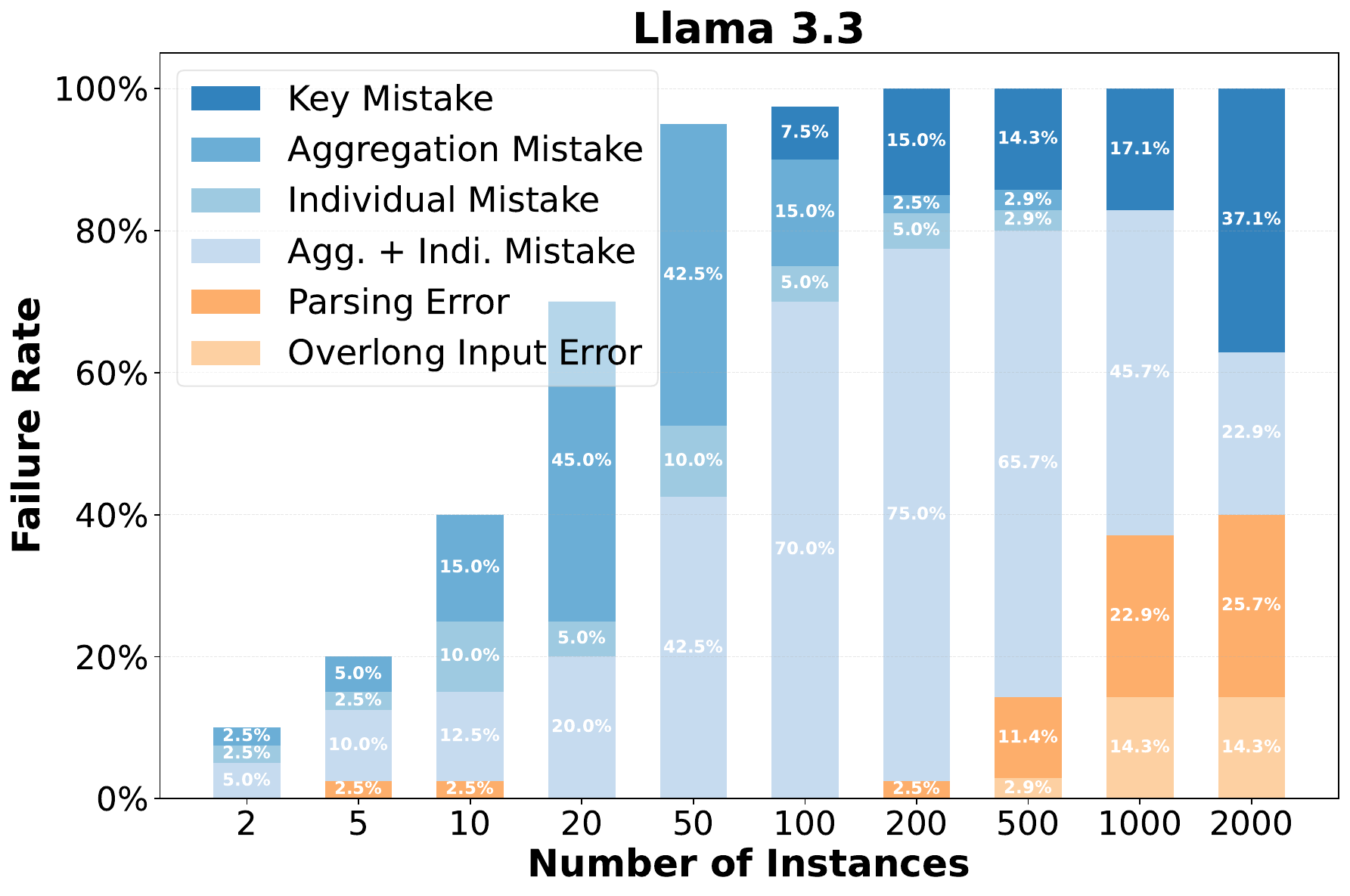}
  \includegraphics[width=0.9\linewidth]{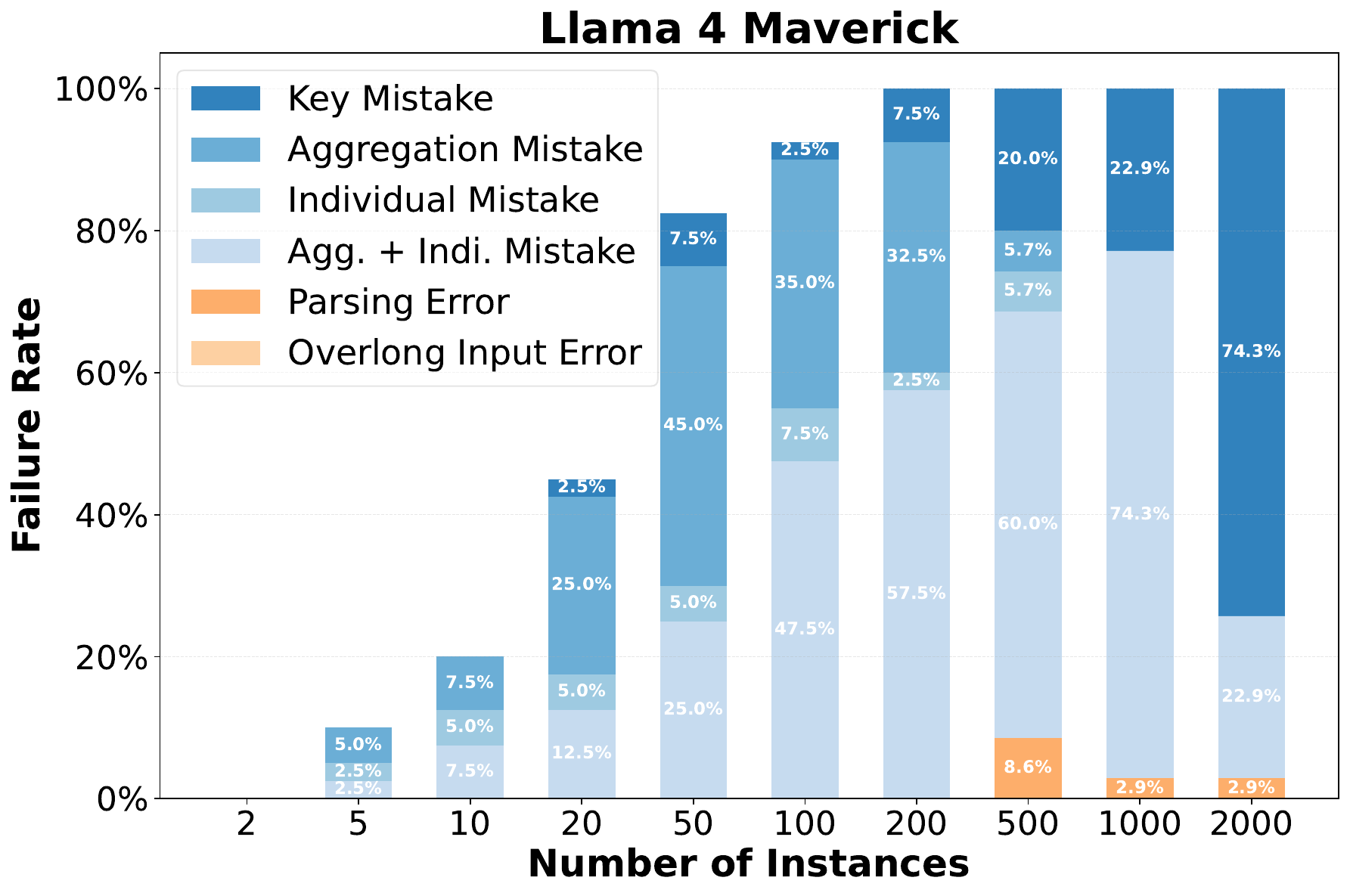}
  \includegraphics[width=0.9\linewidth]{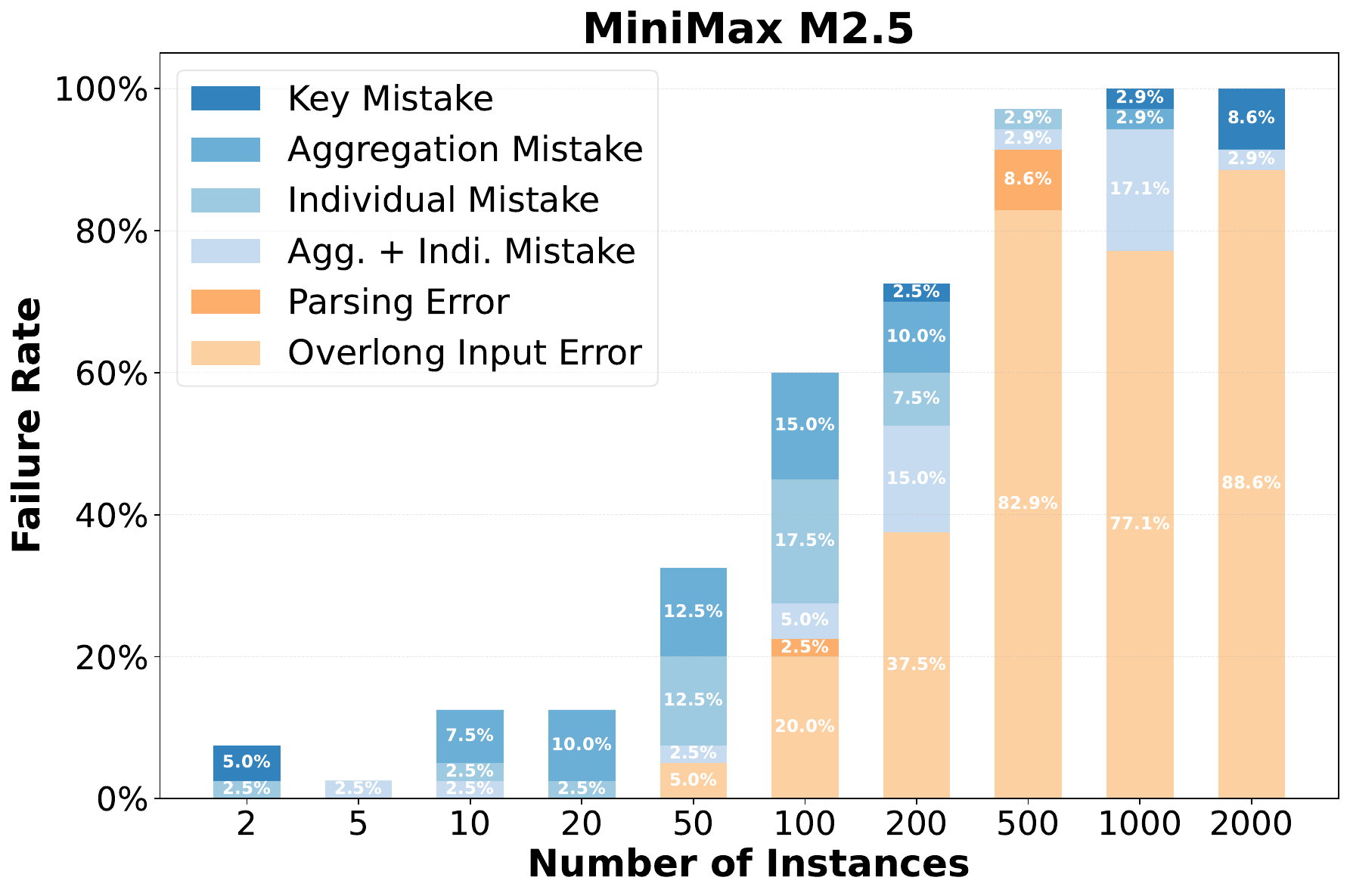}
  \includegraphics[width=0.9\linewidth]{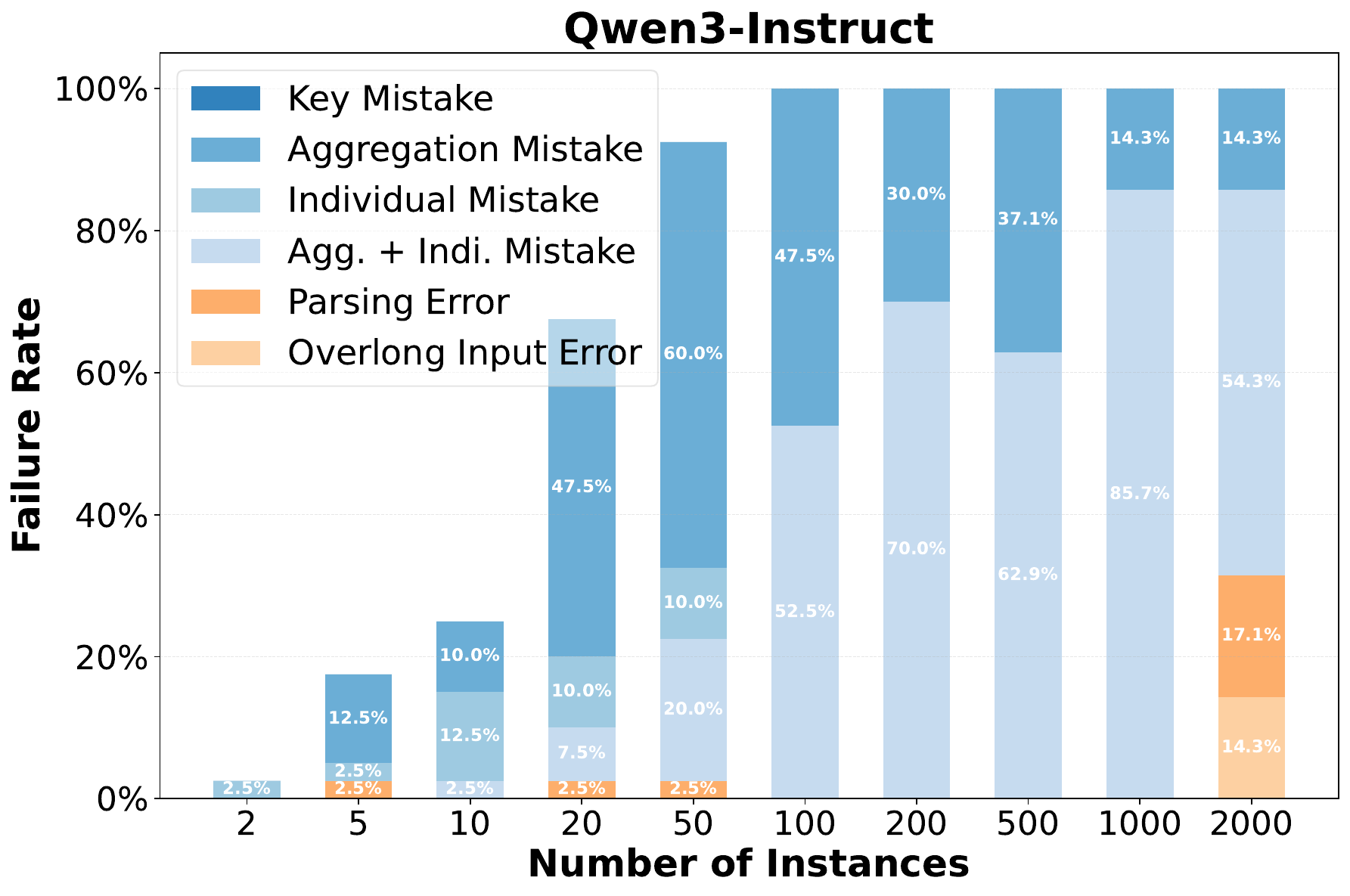}

  \caption{Failure breakdown for models.}
  \label{fig:error_breakdown_models_2}
\end{figure}

\begin{figure}[t]
  \centering
  \setlength{\abovecaptionskip}{2pt}
  \setlength{\belowcaptionskip}{0pt}

  \includegraphics[width=0.9\linewidth]{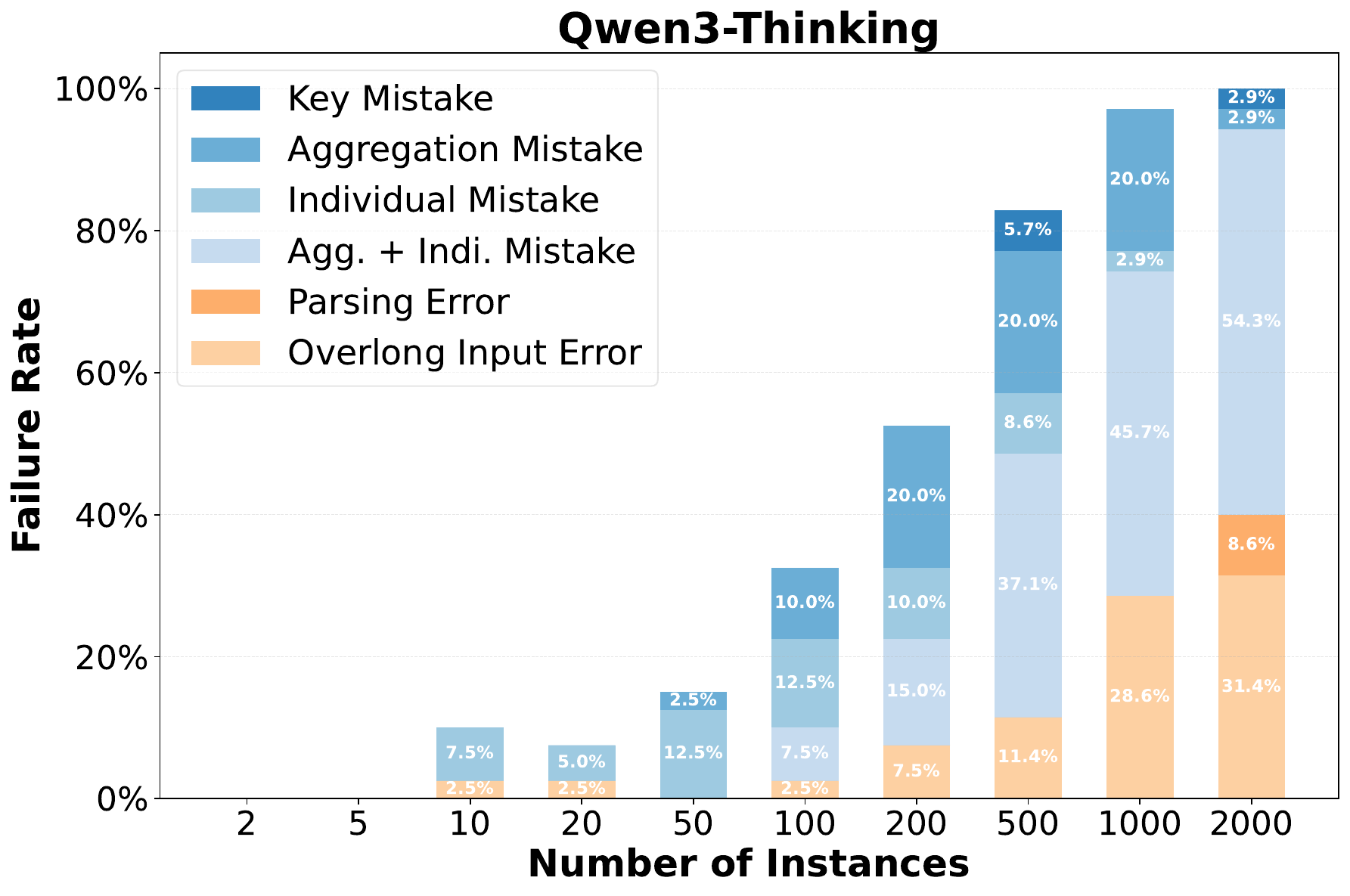}
  \includegraphics[width=0.9\linewidth]{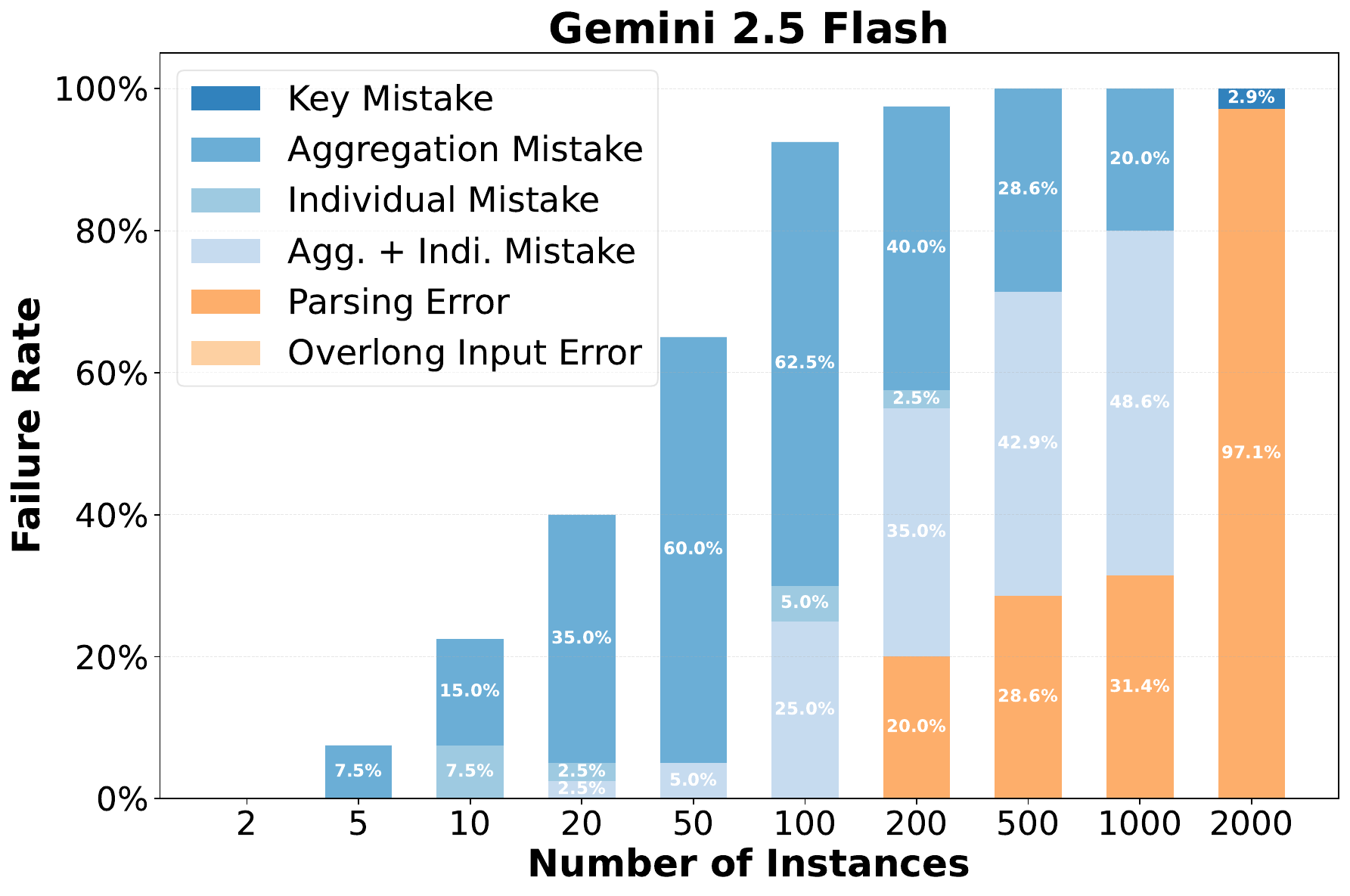}
  \includegraphics[width=0.9\linewidth]{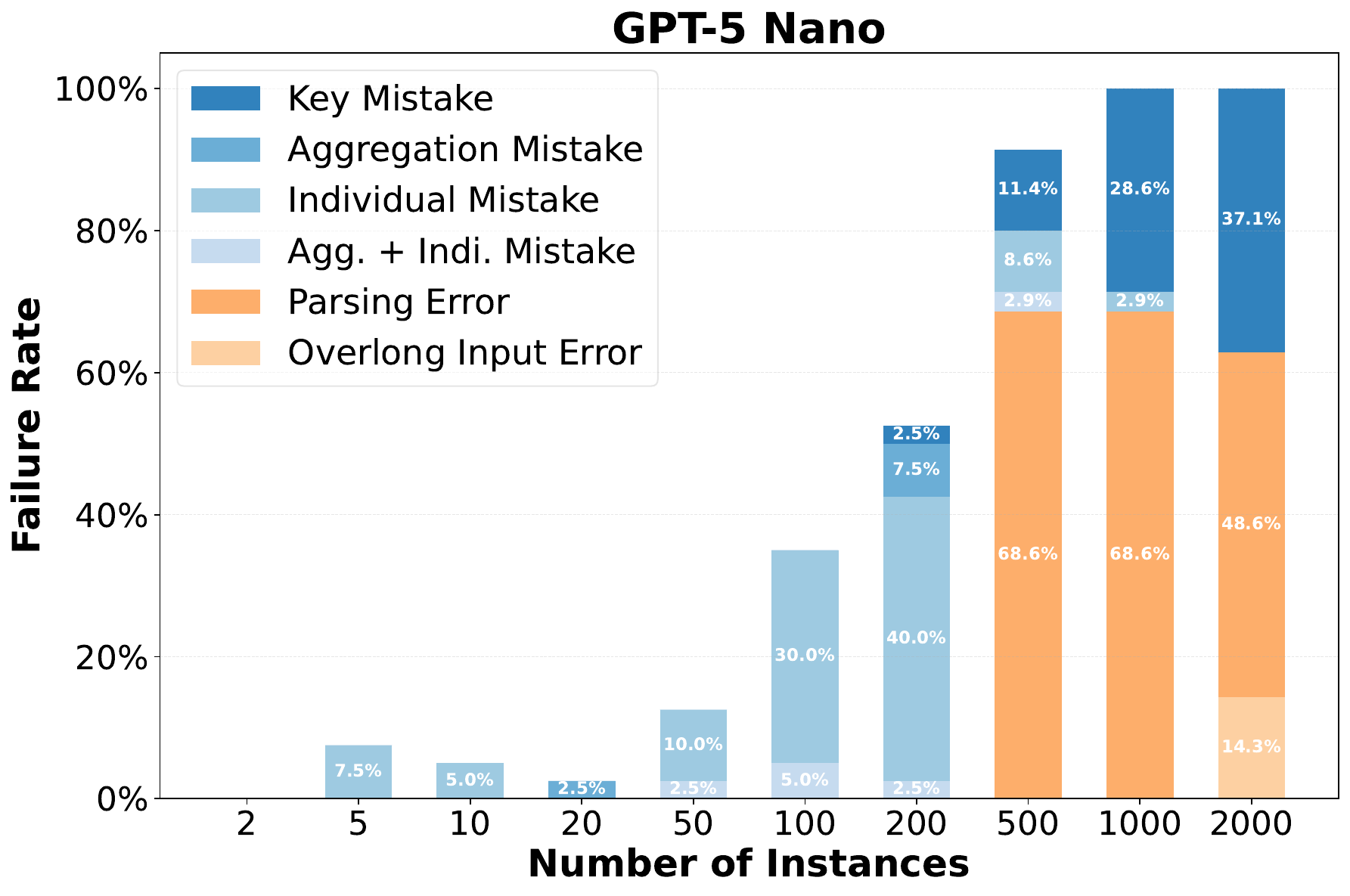}
  \includegraphics[width=0.9\linewidth]{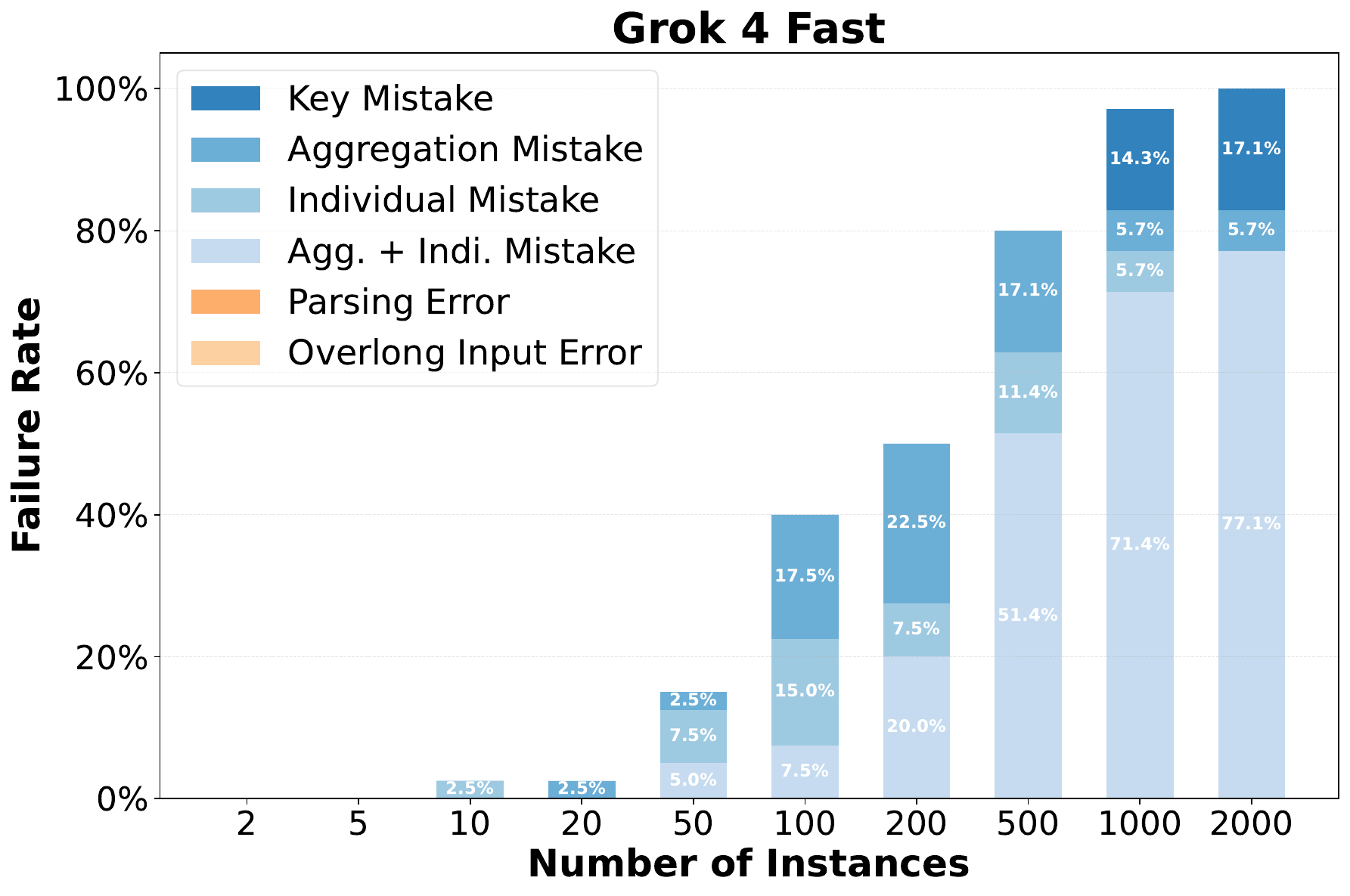}

  \caption{Failure breakdown for models.}
  \label{fig:error_breakdown_models_3}
\end{figure}

Figure~\ref{fig:error_breakdown_models_1}, Figure~\ref{fig:error_breakdown_models_2} and Figure~\ref{fig:error_breakdown_models_3} show failure breakdown for models, averaged across all tasks.

\subsubsection{Failure Breakdown for Tasks}
\label{appendix:error_breakdown}
\begin{figure}[t]
  \centering
  \setlength{\abovecaptionskip}{2pt}
  \setlength{\belowcaptionskip}{0pt}

  \includegraphics[width=0.9\linewidth]{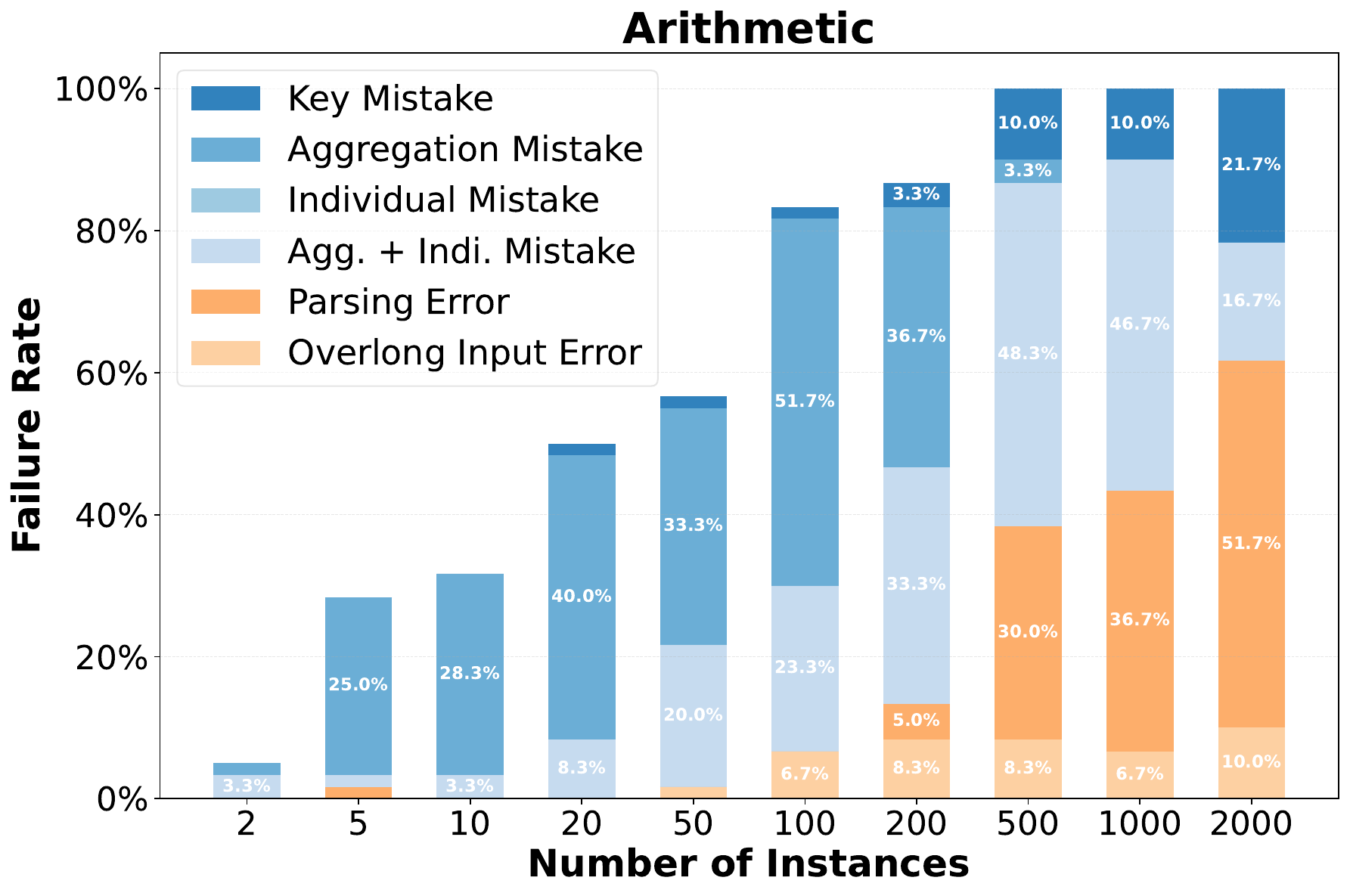}
  \includegraphics[width=0.9\linewidth]{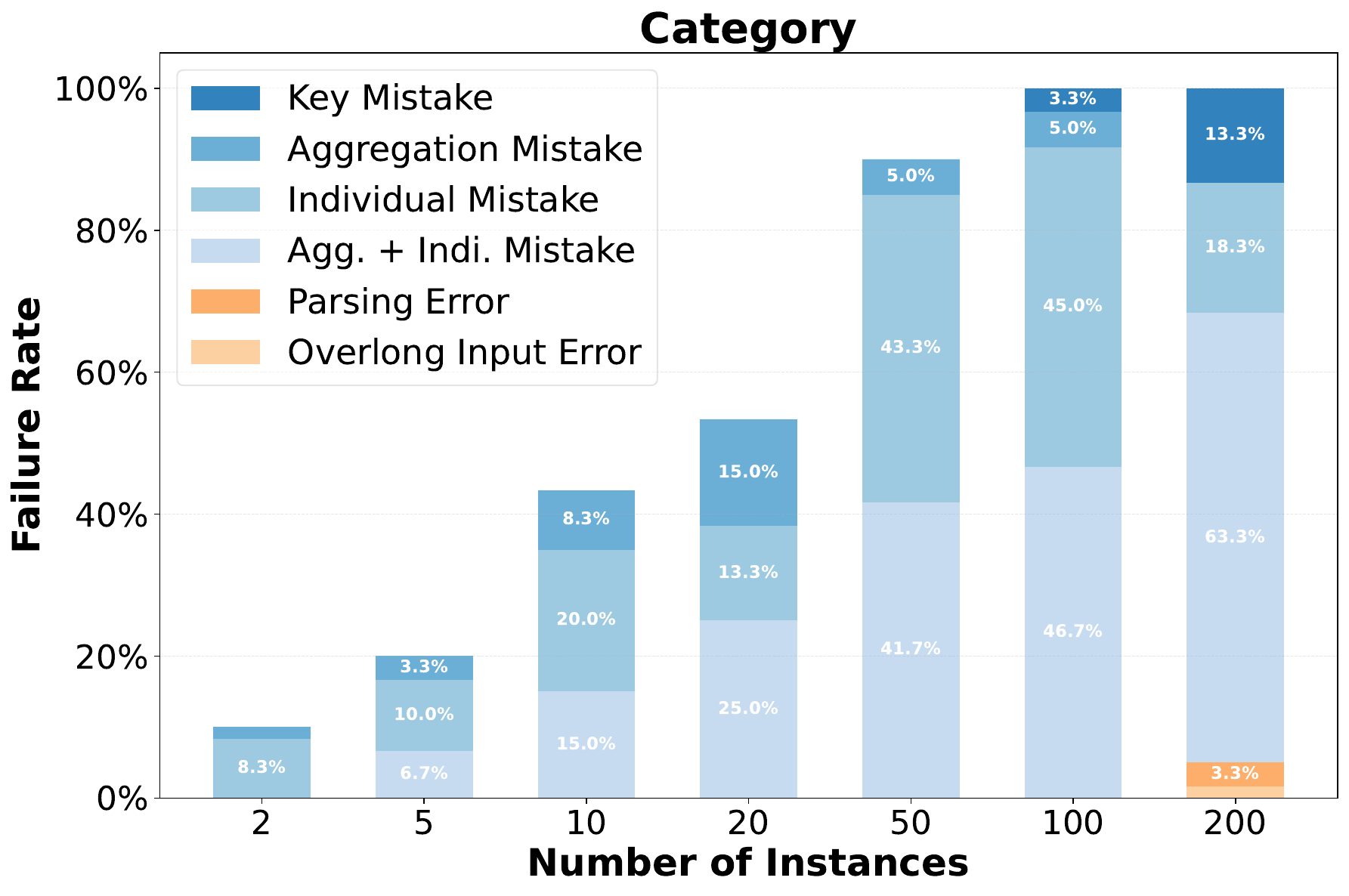}
  \includegraphics[width=0.9\linewidth]{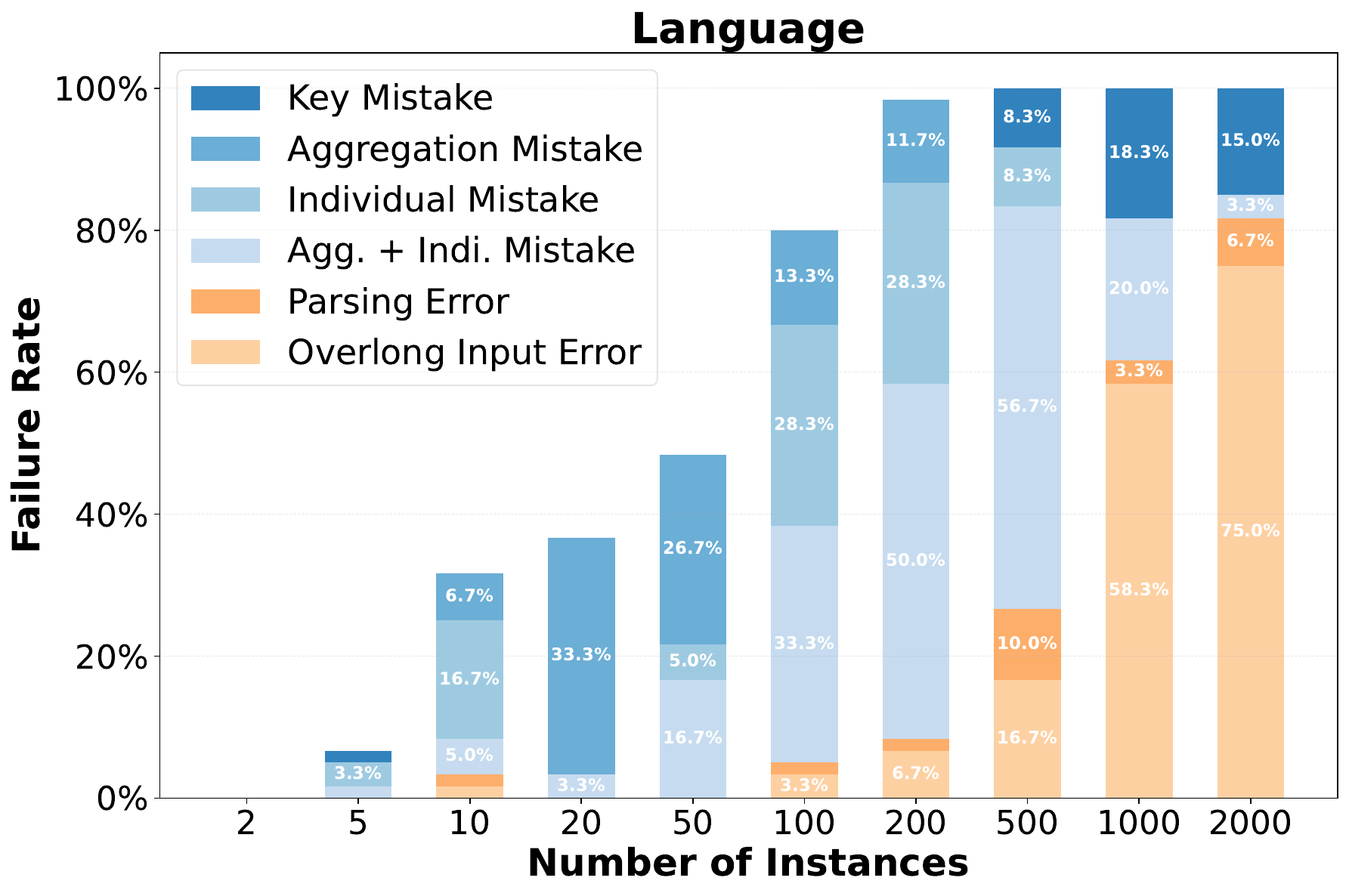}
  \includegraphics[width=0.9\linewidth]{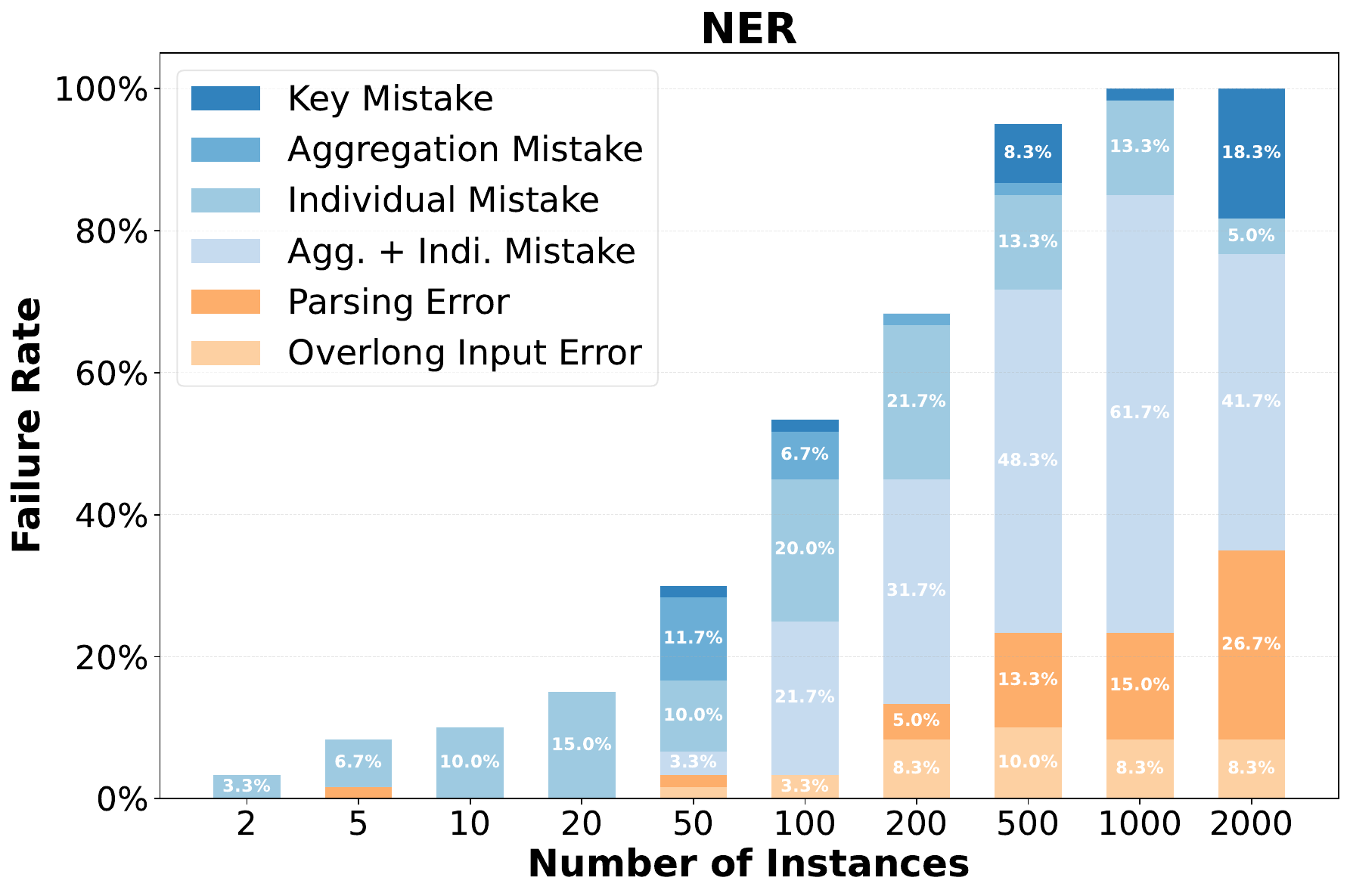}

  \caption{Failure breakdown for tasks.}
  \label{fig:error_breakdown_tasks_1}
\end{figure}

\begin{figure}[t]
  \centering
  \setlength{\abovecaptionskip}{2pt}
  \setlength{\belowcaptionskip}{0pt}

  \includegraphics[width=0.9\linewidth]{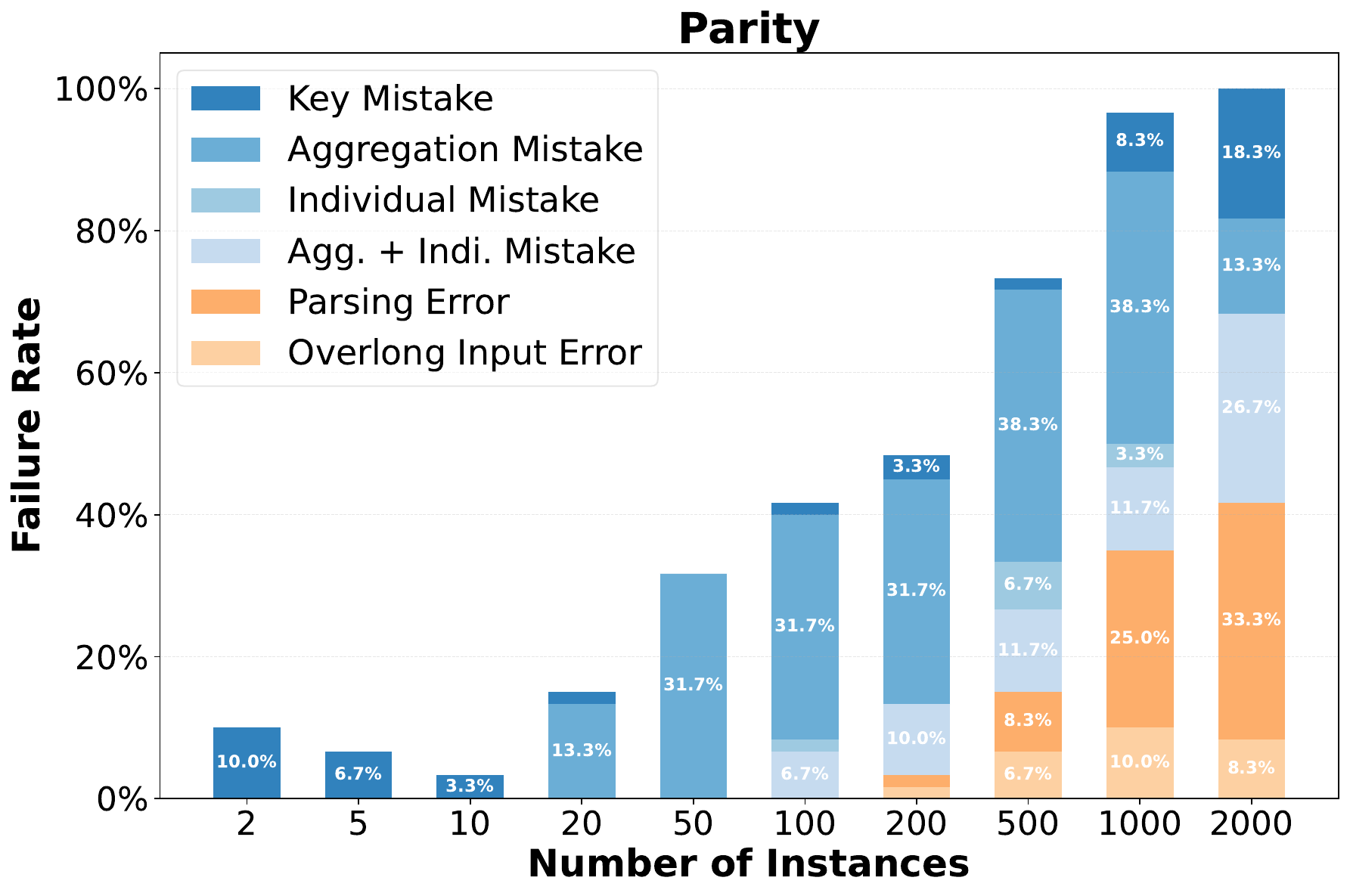}
  \includegraphics[width=0.9\linewidth]{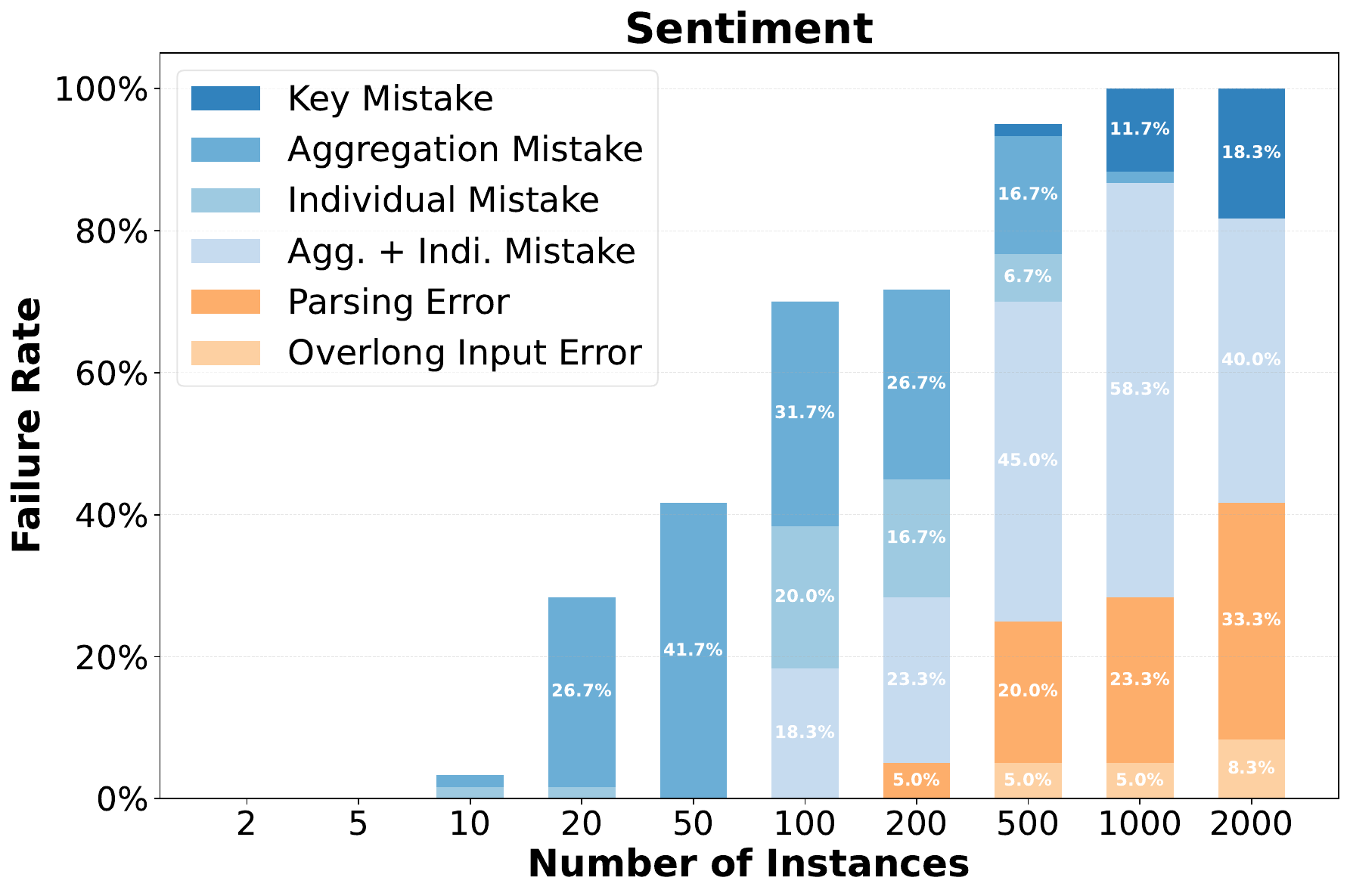}
  \includegraphics[width=0.9\linewidth]{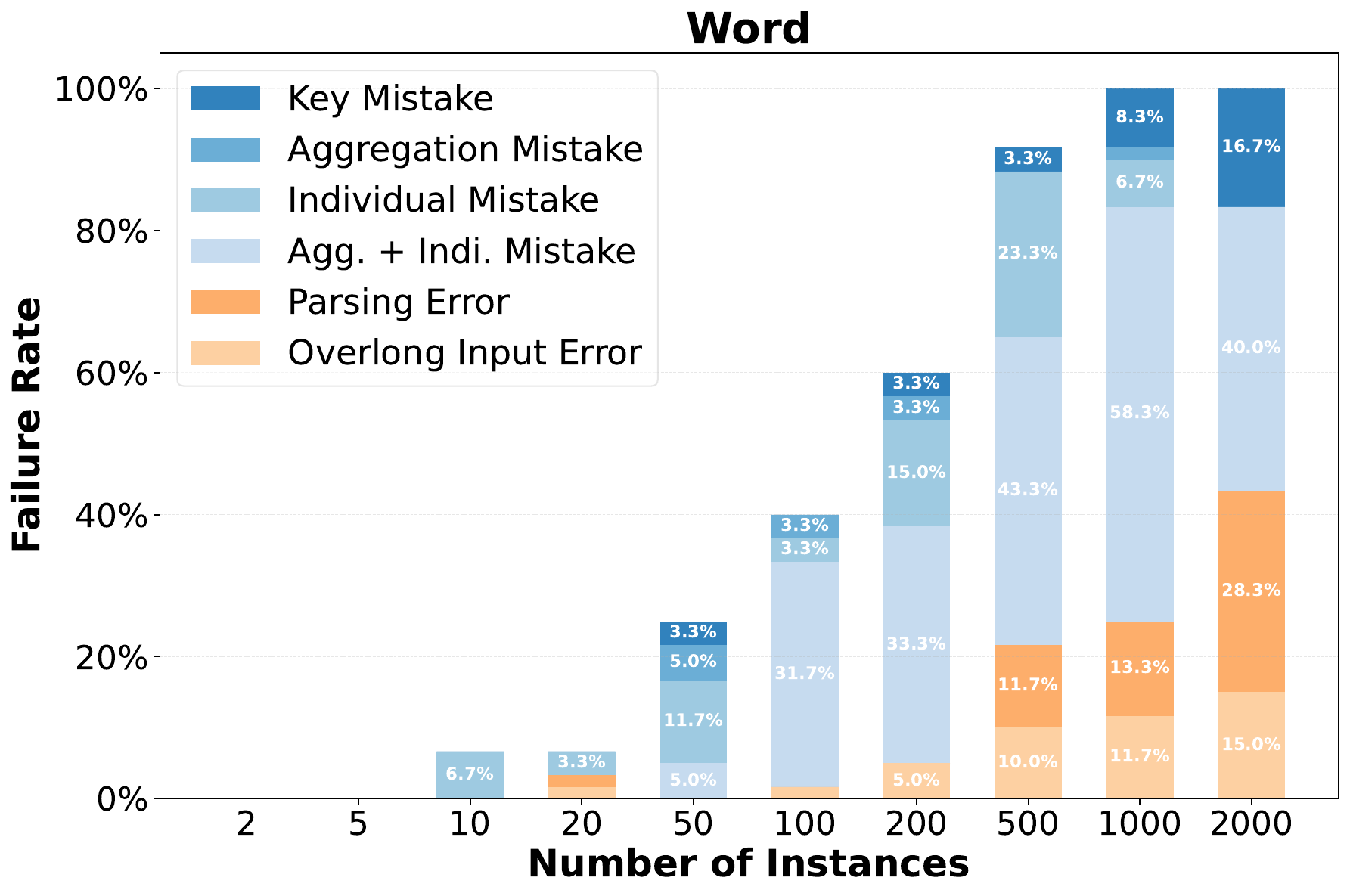}
  \includegraphics[width=0.9\linewidth]{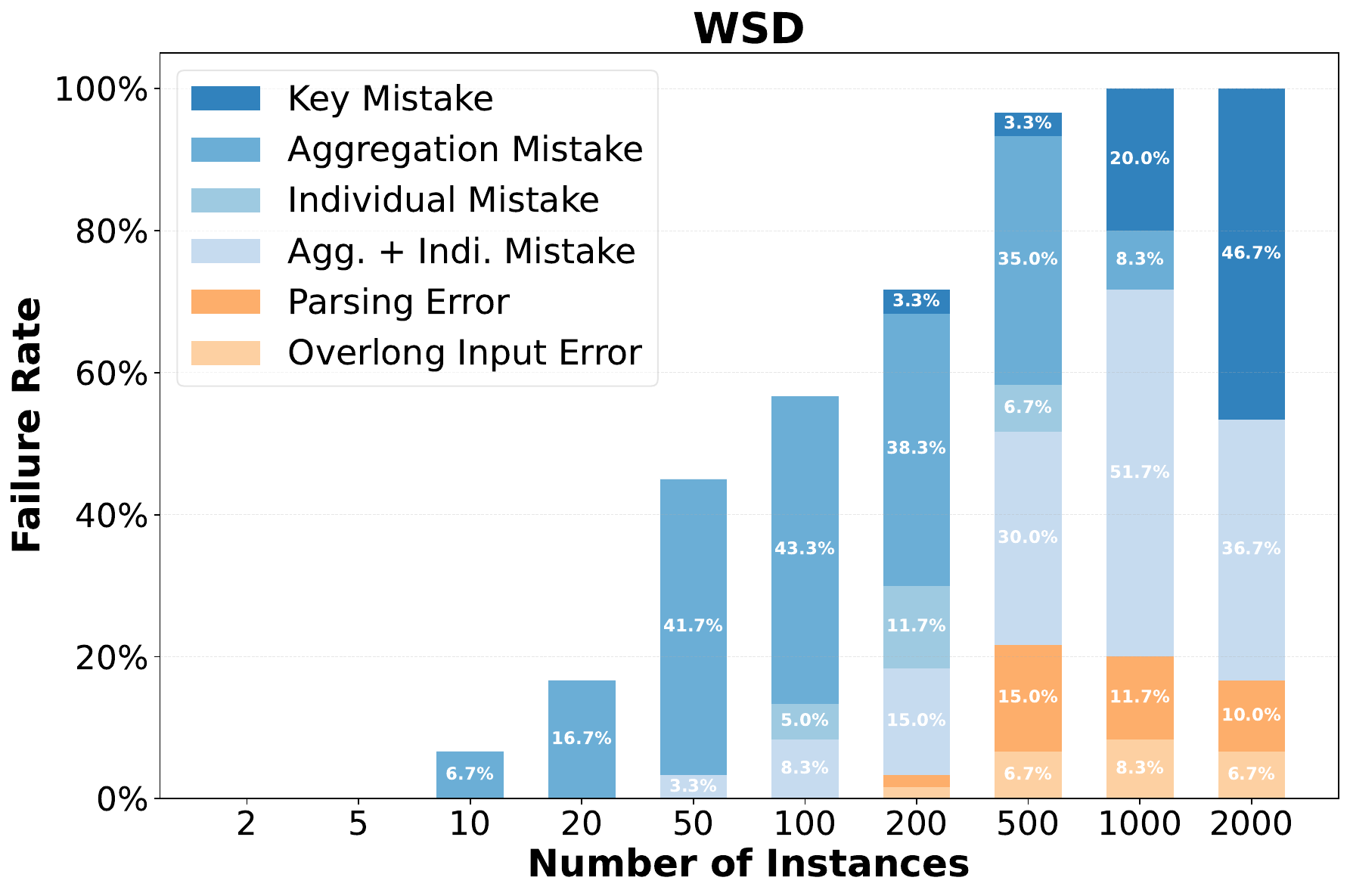}

  \caption{Failure breakdown for tasks.}
  \label{fig:error_breakdown_tasks_2}
\end{figure}

Figure~\ref{fig:error_breakdown_tasks_1} and Figure~\ref{fig:error_breakdown_tasks_2} show failure breakdown for tasks, averaged across all models.

\begin{figure}[t]
  \includegraphics[width=\columnwidth]{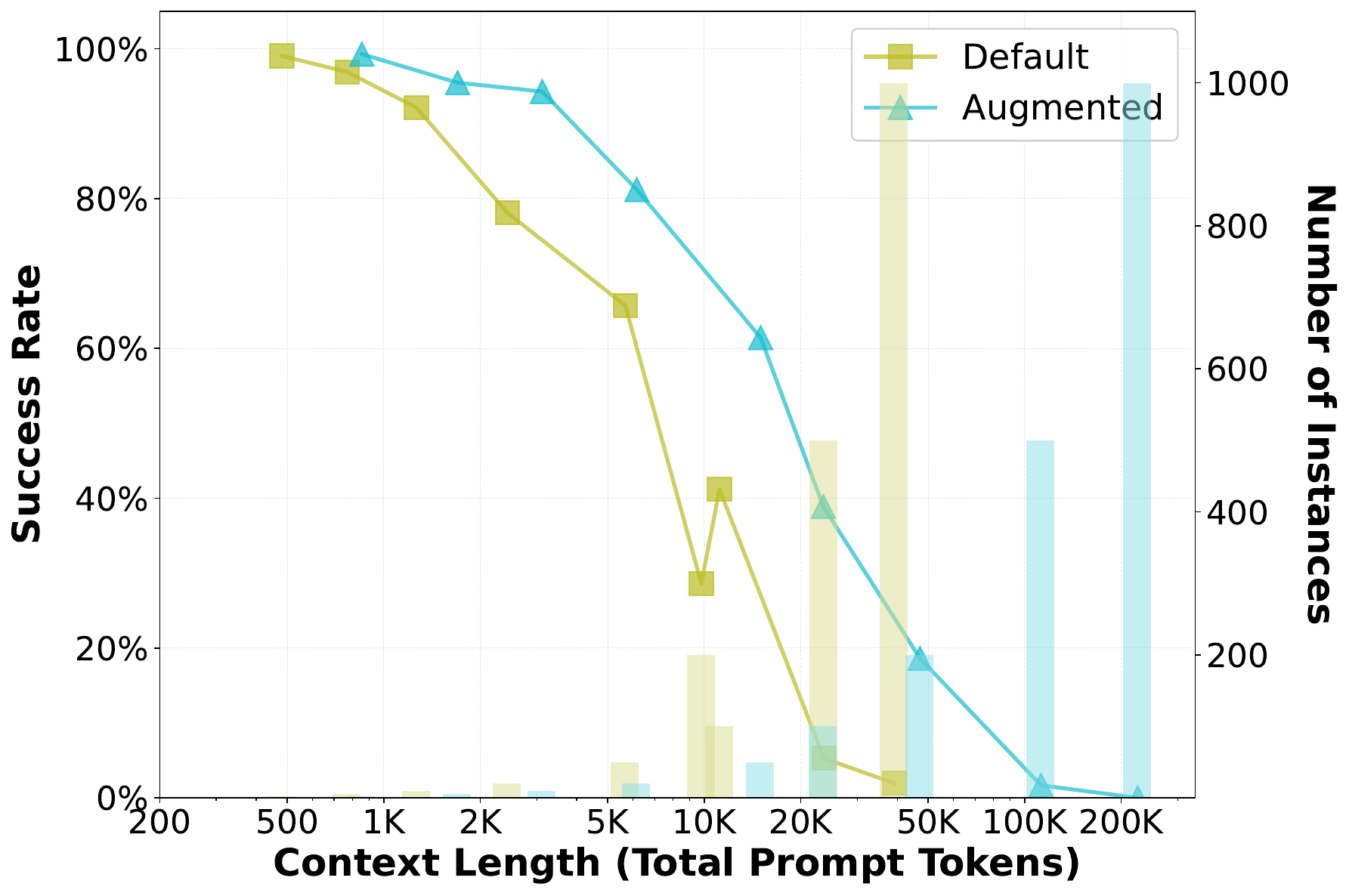}
  \caption{\camera{Success rate (lines) and the number of instances (bars) in the artificial length setting as total prompt token length increases. Error bars indicate standard deviation across five random seeds.}}
  \label{fig:artificial_overall_context}
\end{figure}

\camera{\subsection{Context Length}}
\label{appendix:context_length}
\camera{\subsubsection{Default and Augmented Success Rate}
Figure~\ref{fig:artificial_overall_context} compares average performance across all tasks and models between the default and artificially augmented settings, as the context length increases.}

\begin{figure}[t]
  \centering
  \setlength{\abovecaptionskip}{2pt}
  \setlength{\belowcaptionskip}{0pt}

  \includegraphics[width=0.9\linewidth]{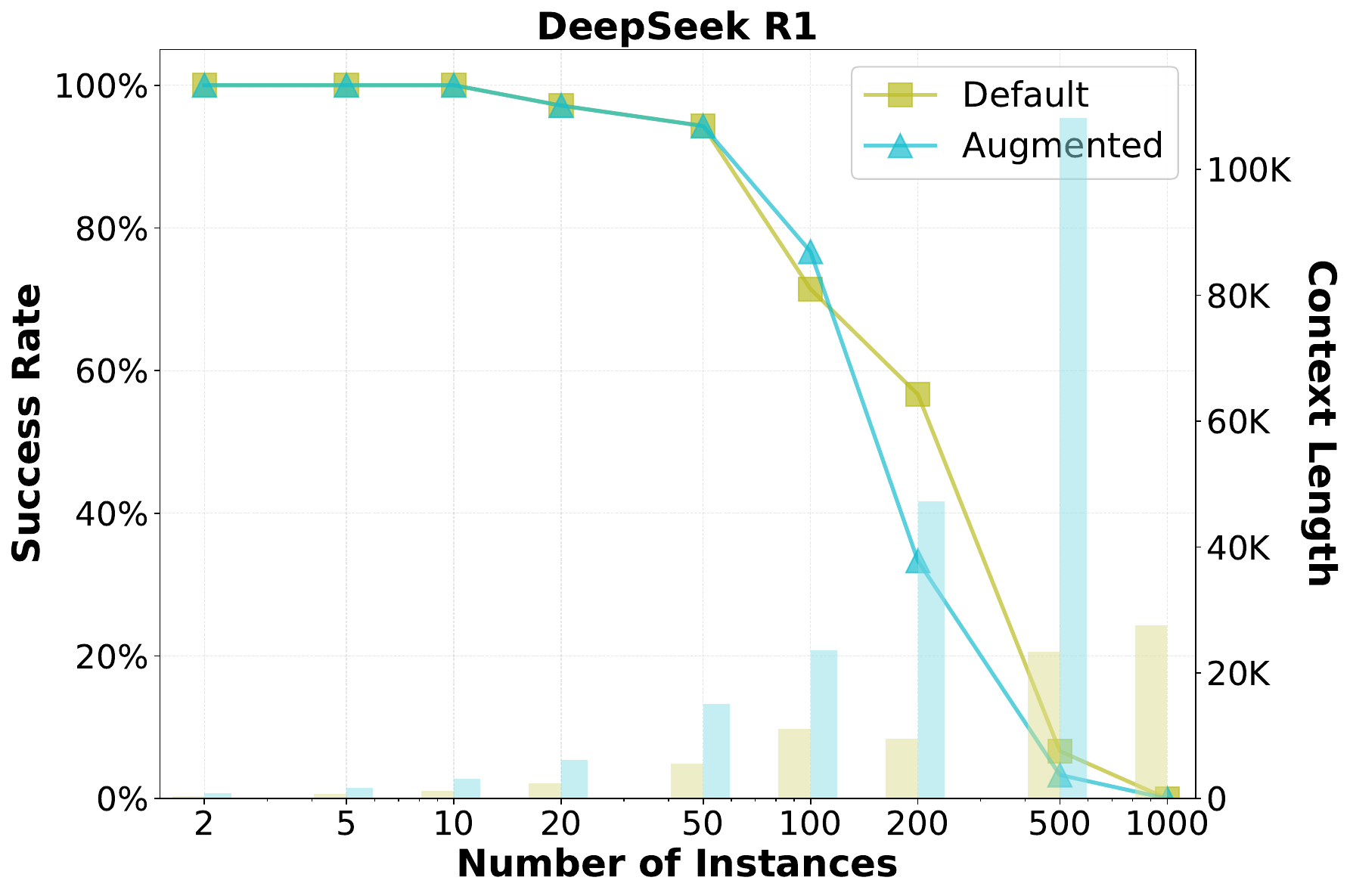}
  \includegraphics[width=0.9\linewidth]{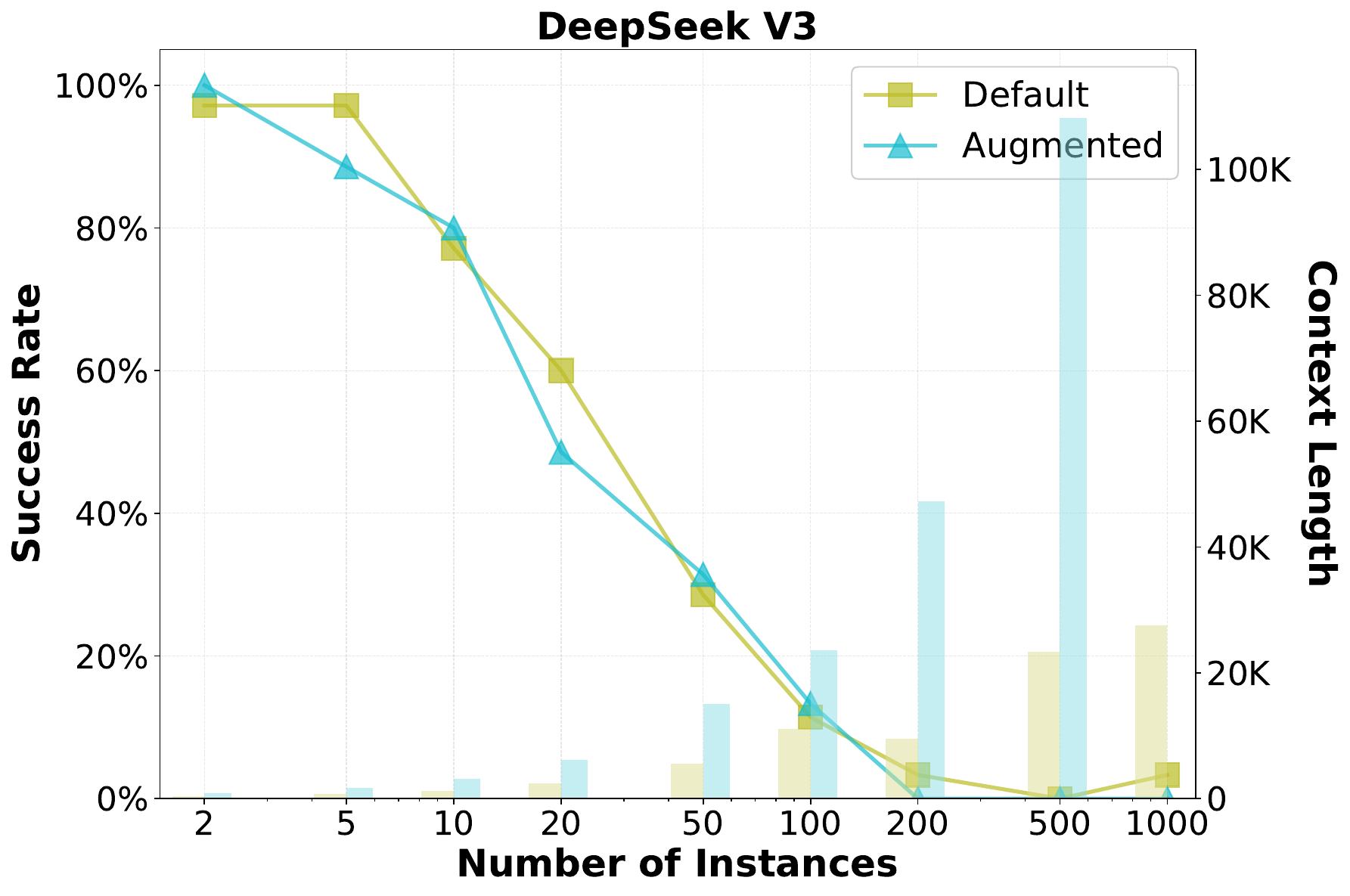}
  \includegraphics[width=0.9\linewidth]{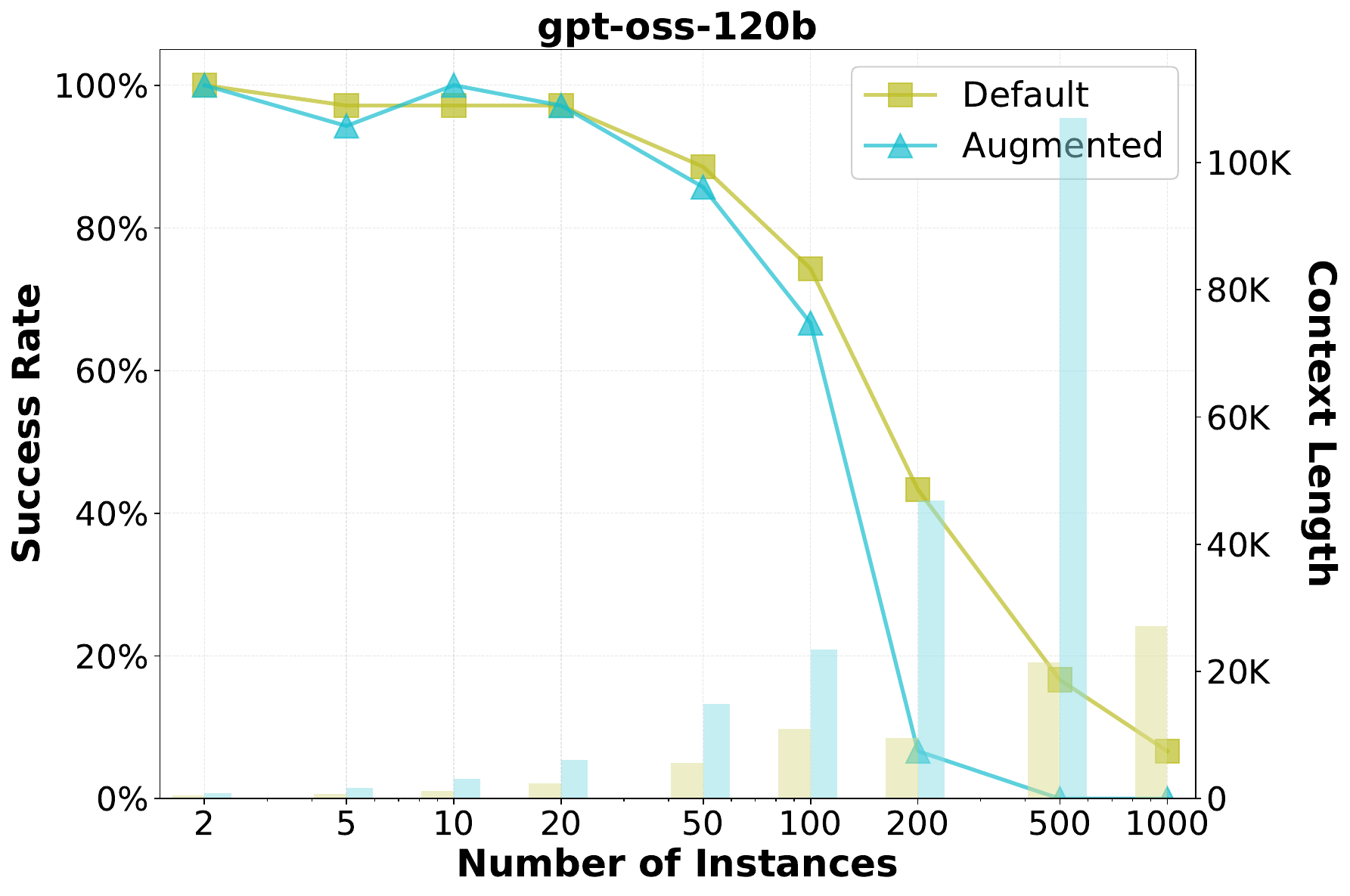}
  \includegraphics[width=0.9\linewidth]{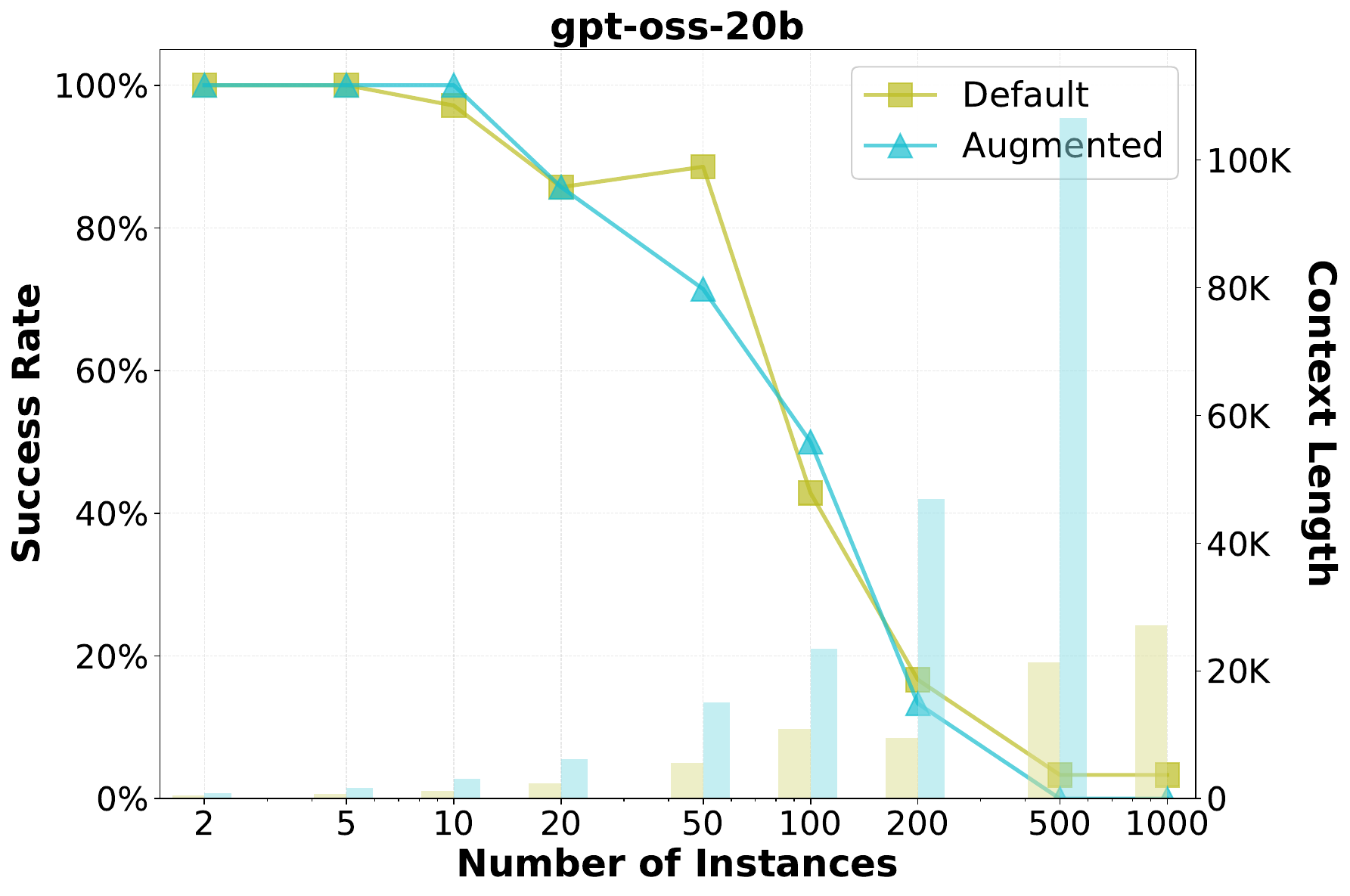}

  \caption{Success rate of models.}
  \label{fig:context_length_models_1}
\end{figure}

\begin{figure}[t]
  \centering
  \setlength{\abovecaptionskip}{2pt}
  \setlength{\belowcaptionskip}{0pt}

  \includegraphics[width=0.9\linewidth]{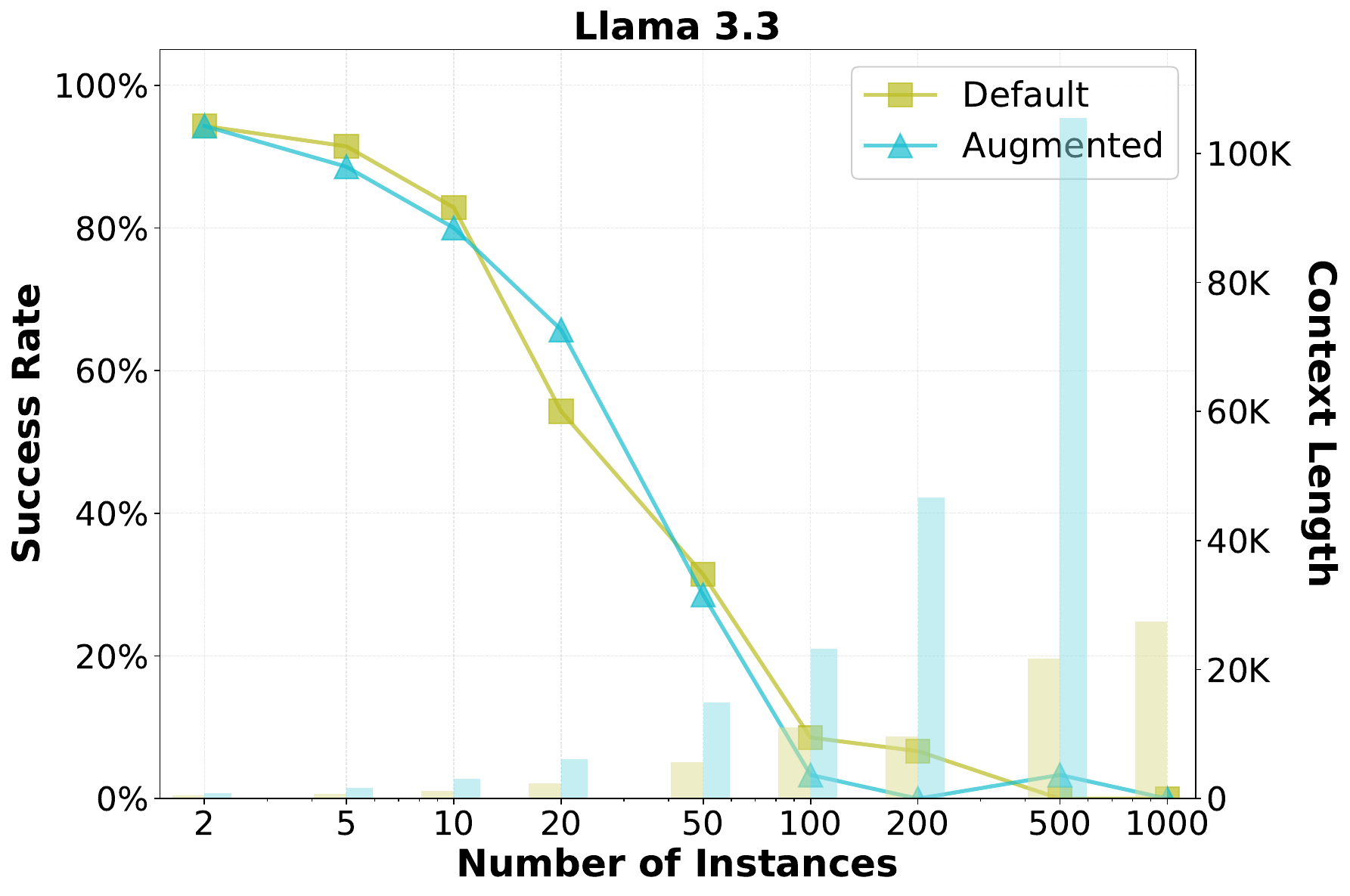}
  \includegraphics[width=0.9\linewidth]{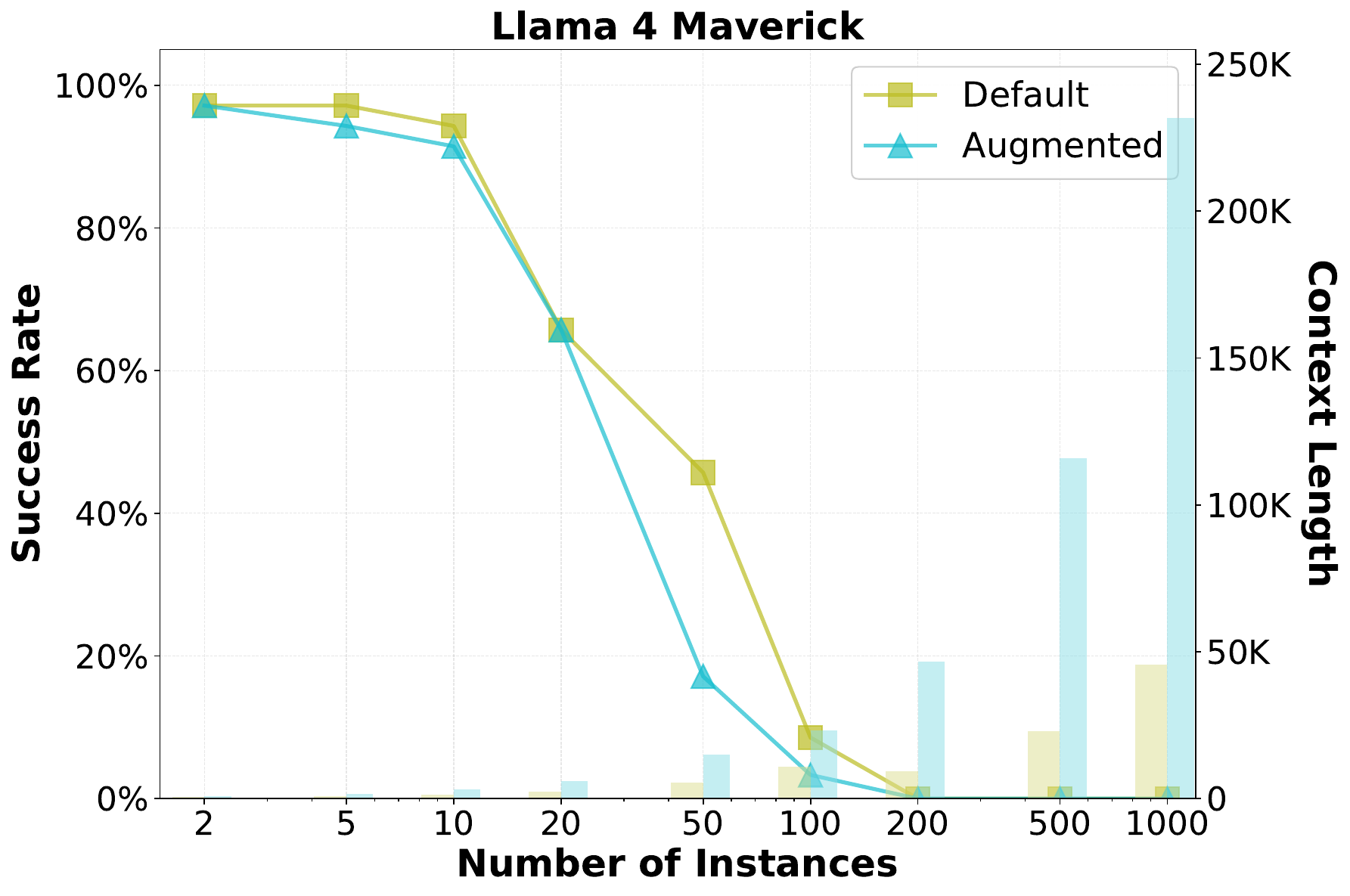}
  \includegraphics[width=0.9\linewidth]{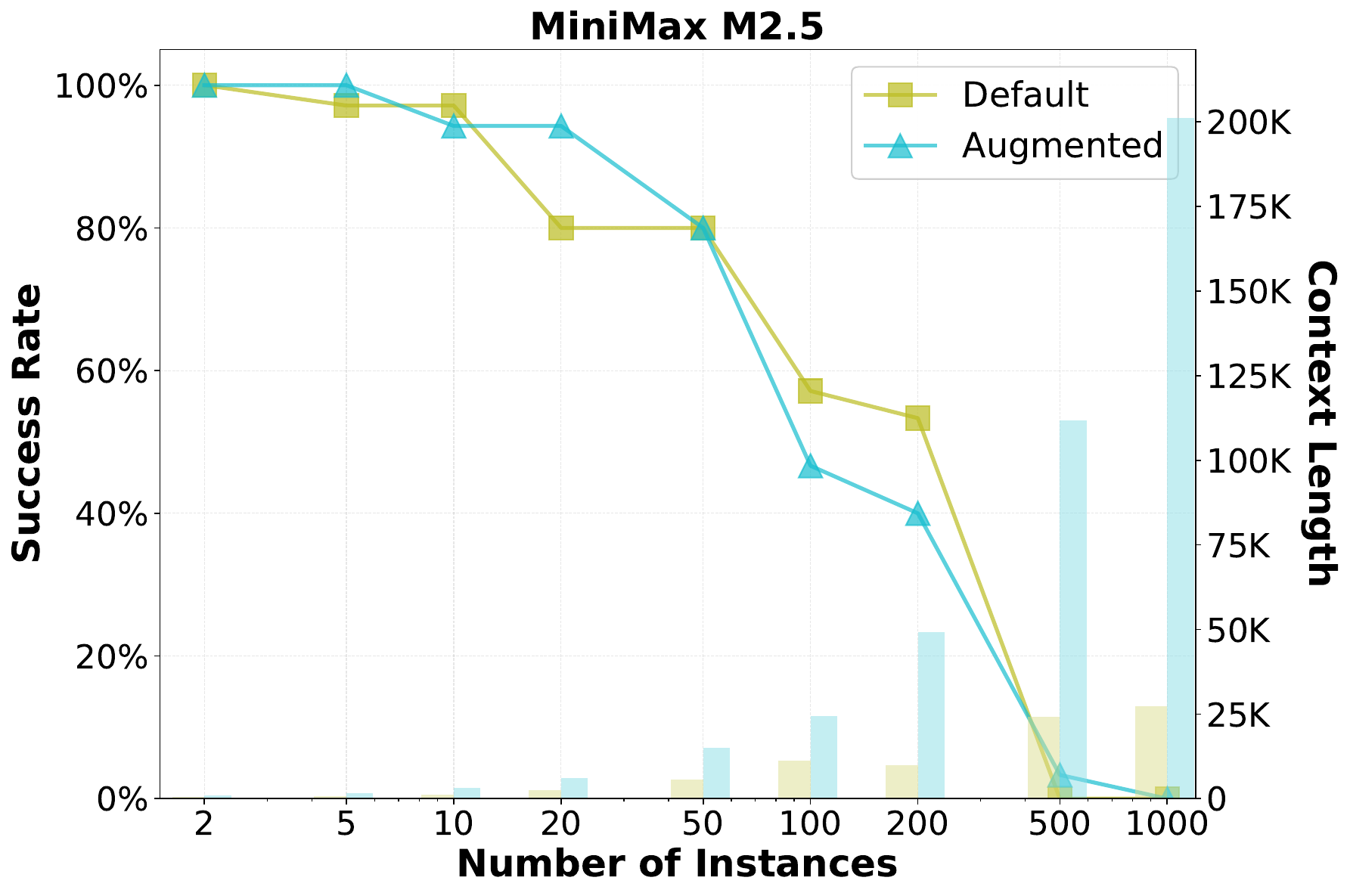}
  \includegraphics[width=0.9\linewidth]{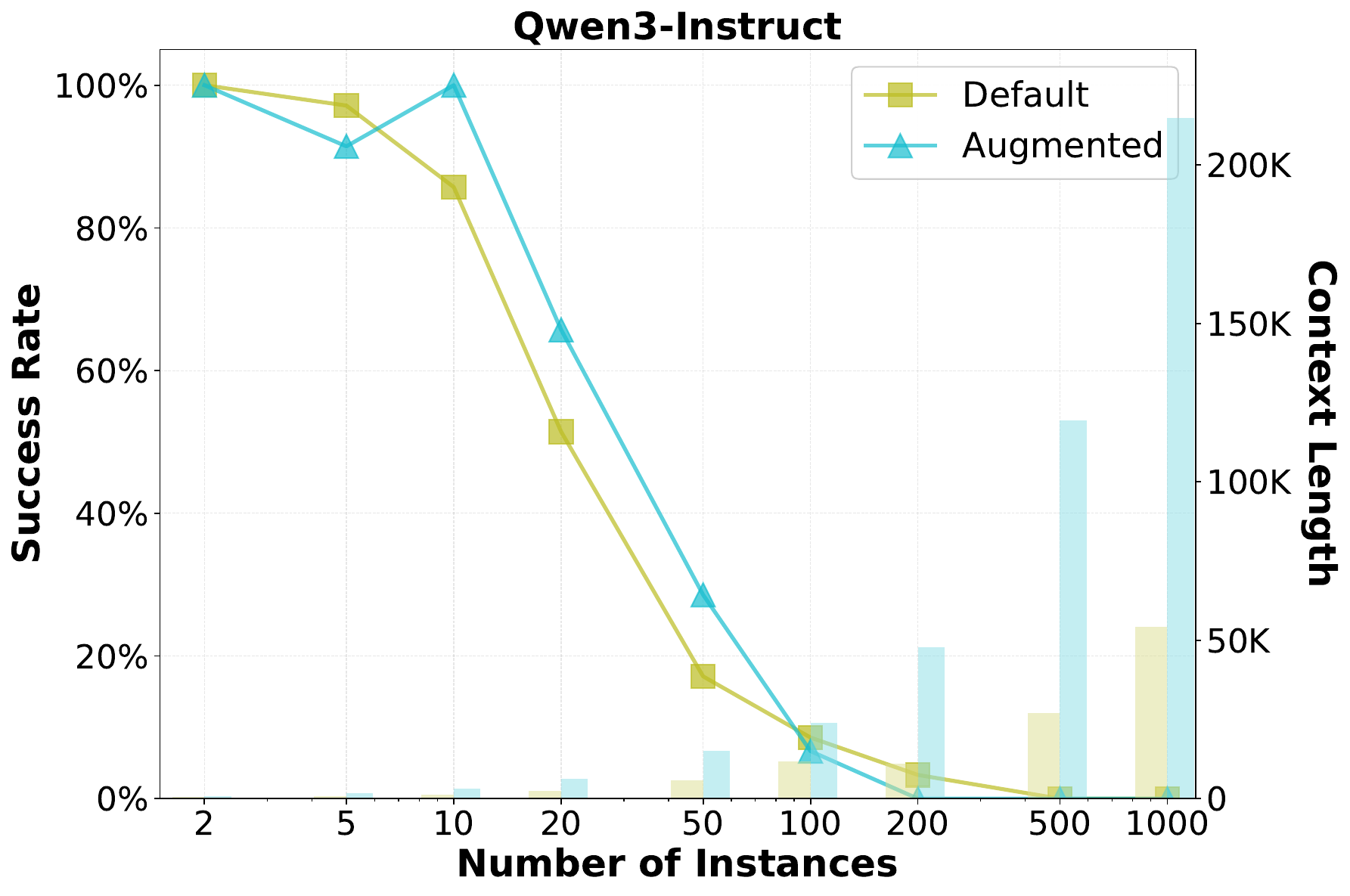}
  \caption{Success rate of models.}
  \label{fig:context_length_models_2}
\end{figure}

\begin{figure}[t]
  \centering
  \setlength{\abovecaptionskip}{2pt}
  \setlength{\belowcaptionskip}{0pt}

  \includegraphics[width=0.9\linewidth]{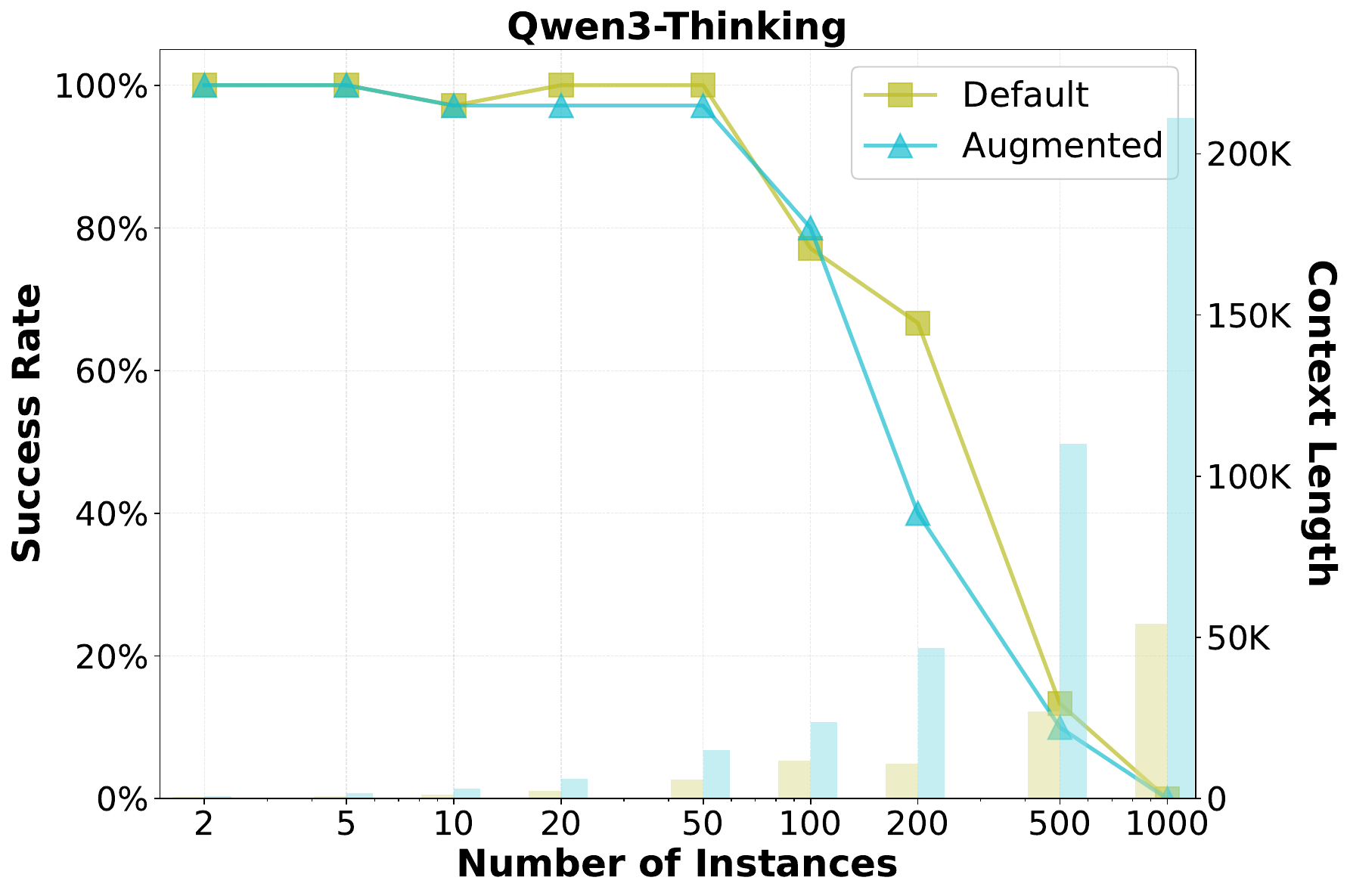}
  \includegraphics[width=0.9\linewidth]{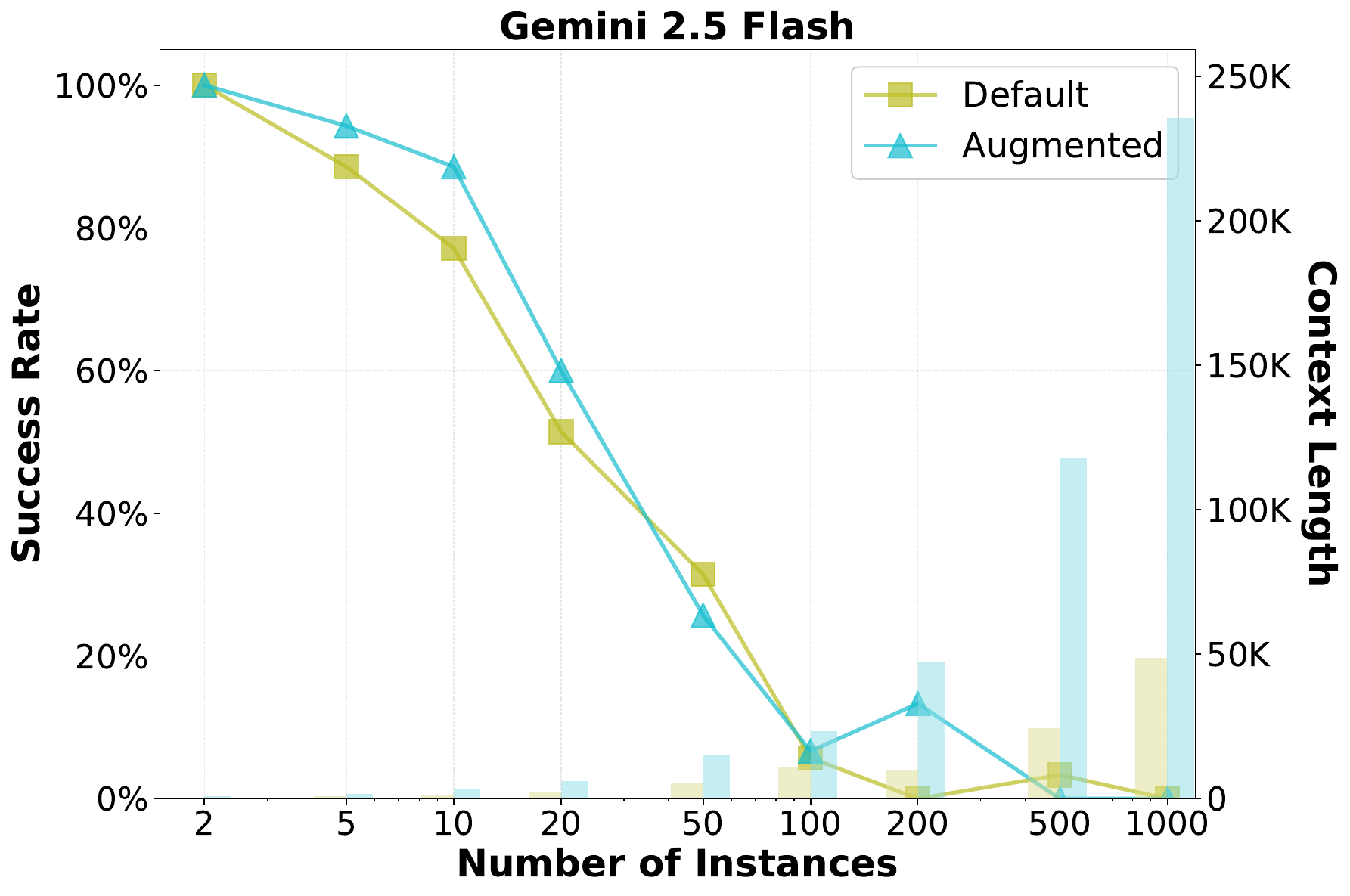}
  \includegraphics[width=0.9\linewidth]{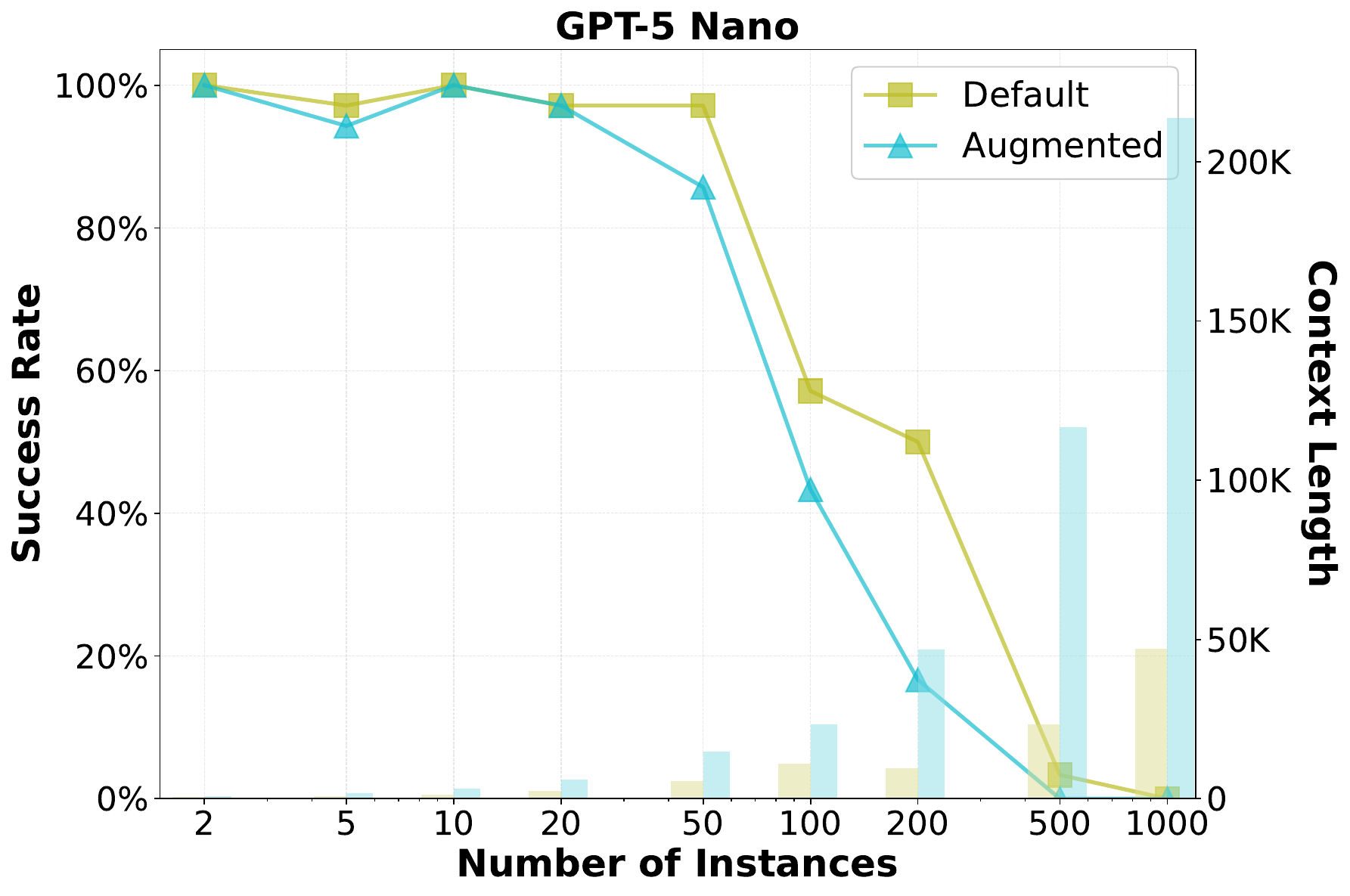}
  \includegraphics[width=0.9\linewidth]{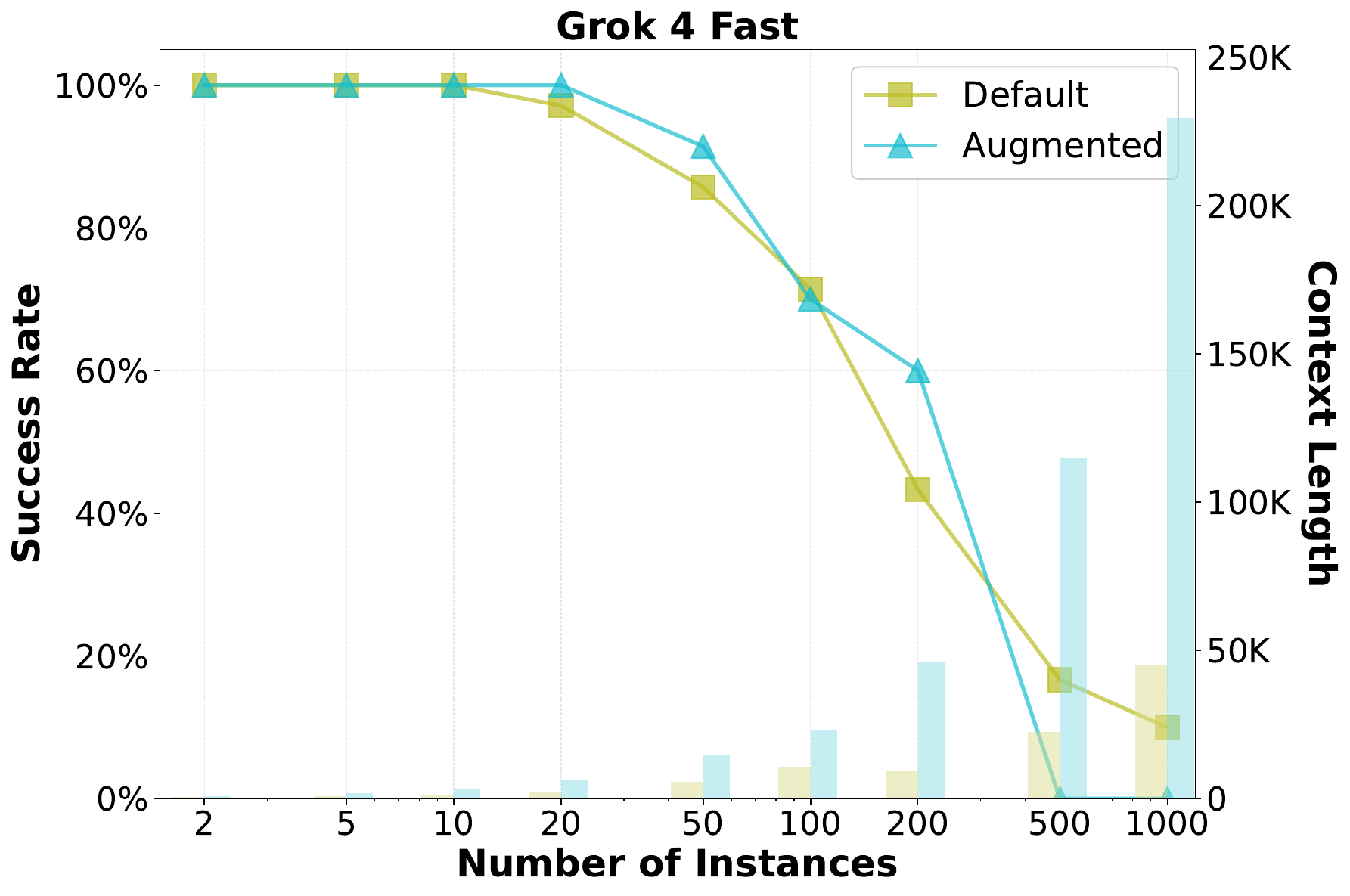}

  \caption{Success rate of models.}
  \label{fig:context_length_models_3}
\end{figure}

\camera{\subsubsection{Default and Augmented Success Rate for Models}
\label{appendix:context_length_models}
Figure~\ref{fig:context_length_models_1}, Figure~\ref{fig:context_length_models_2} and Figure~\ref{fig:context_length_models_3} show task success rate for each model.}

\begin{figure}[t]
  \centering
  \setlength{\abovecaptionskip}{2pt}
  \setlength{\belowcaptionskip}{0pt}

  \includegraphics[width=0.9\linewidth]{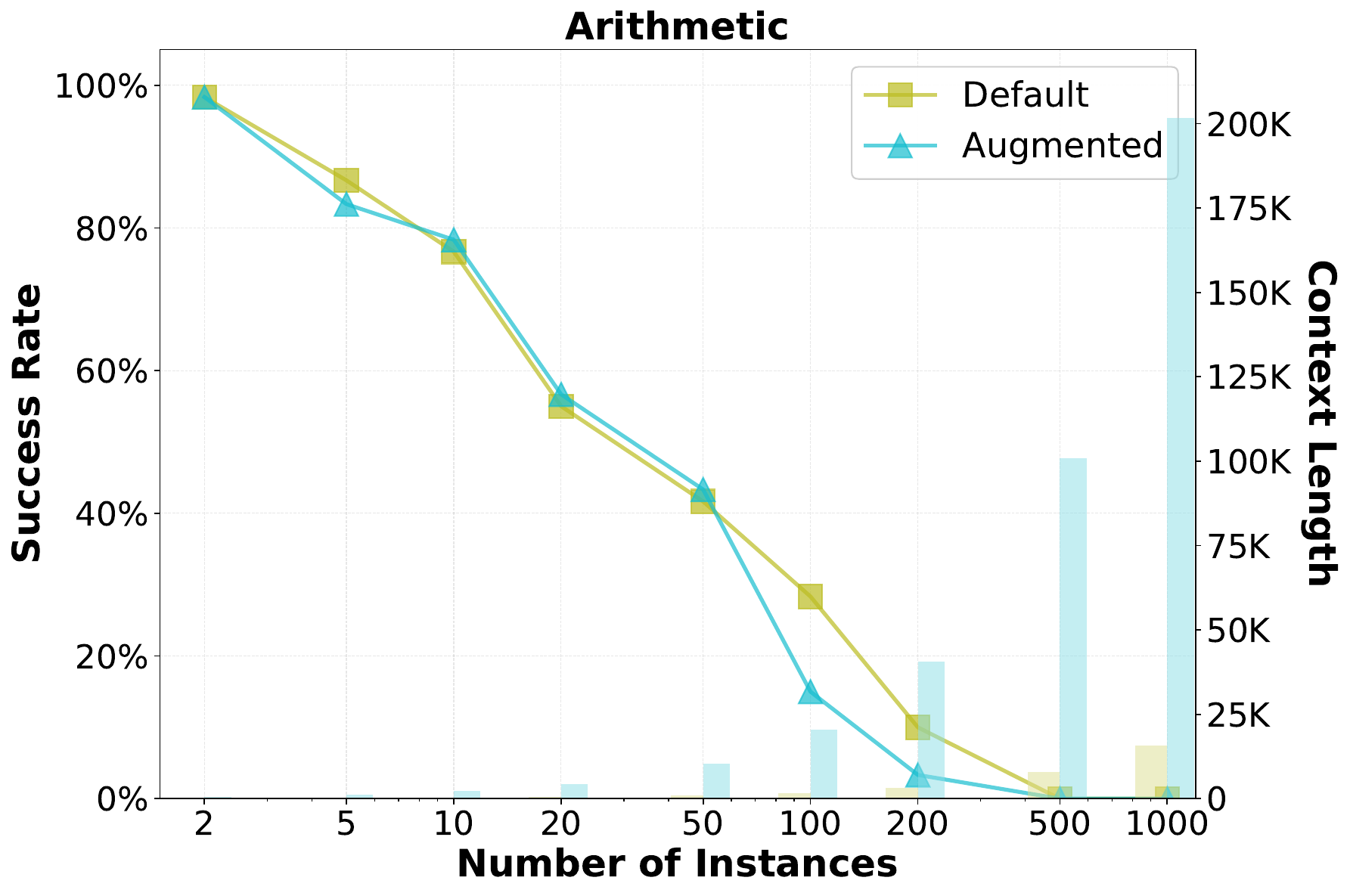}
  \includegraphics[width=0.9\linewidth]{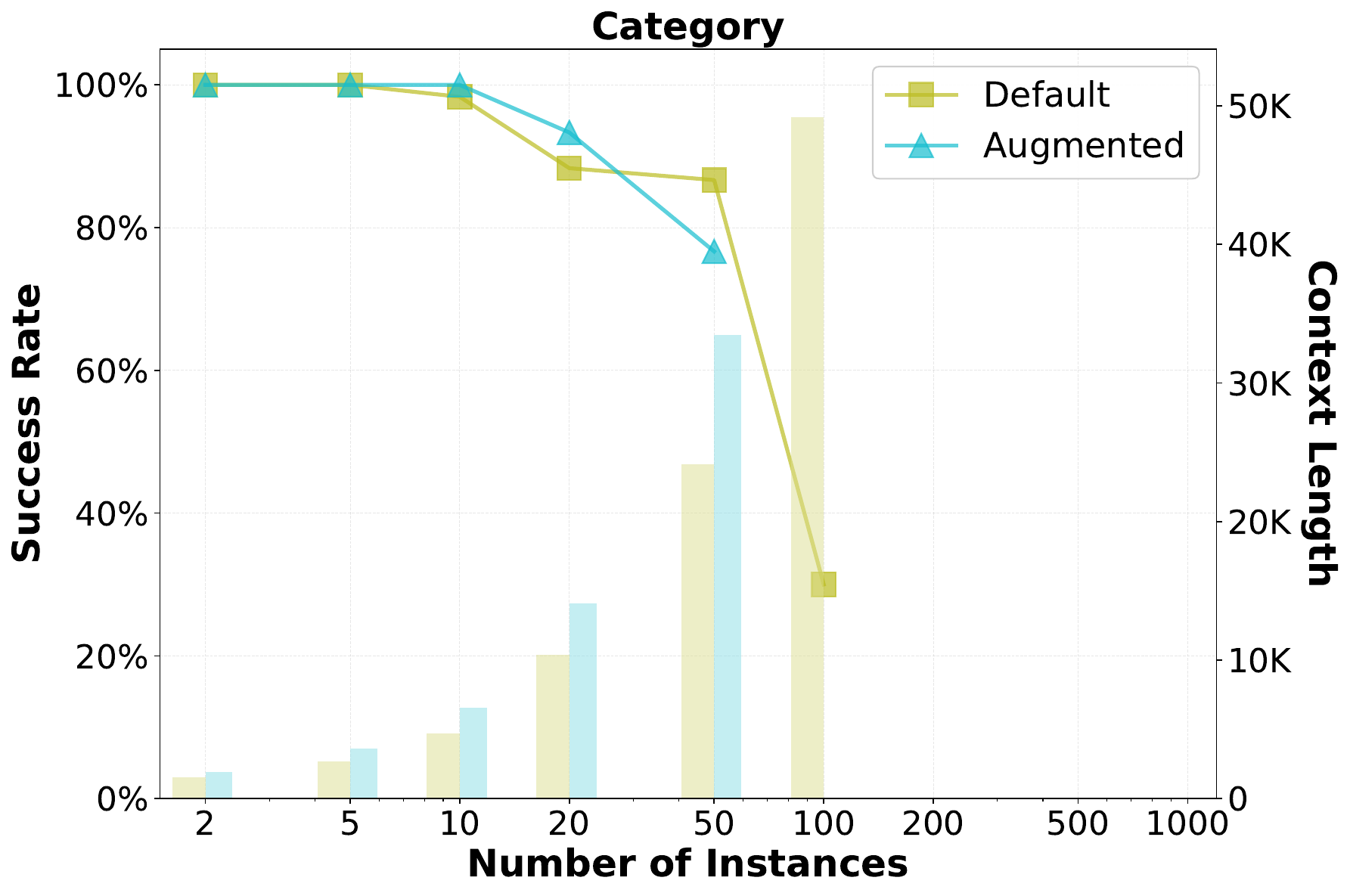}
  \includegraphics[width=0.9\linewidth]{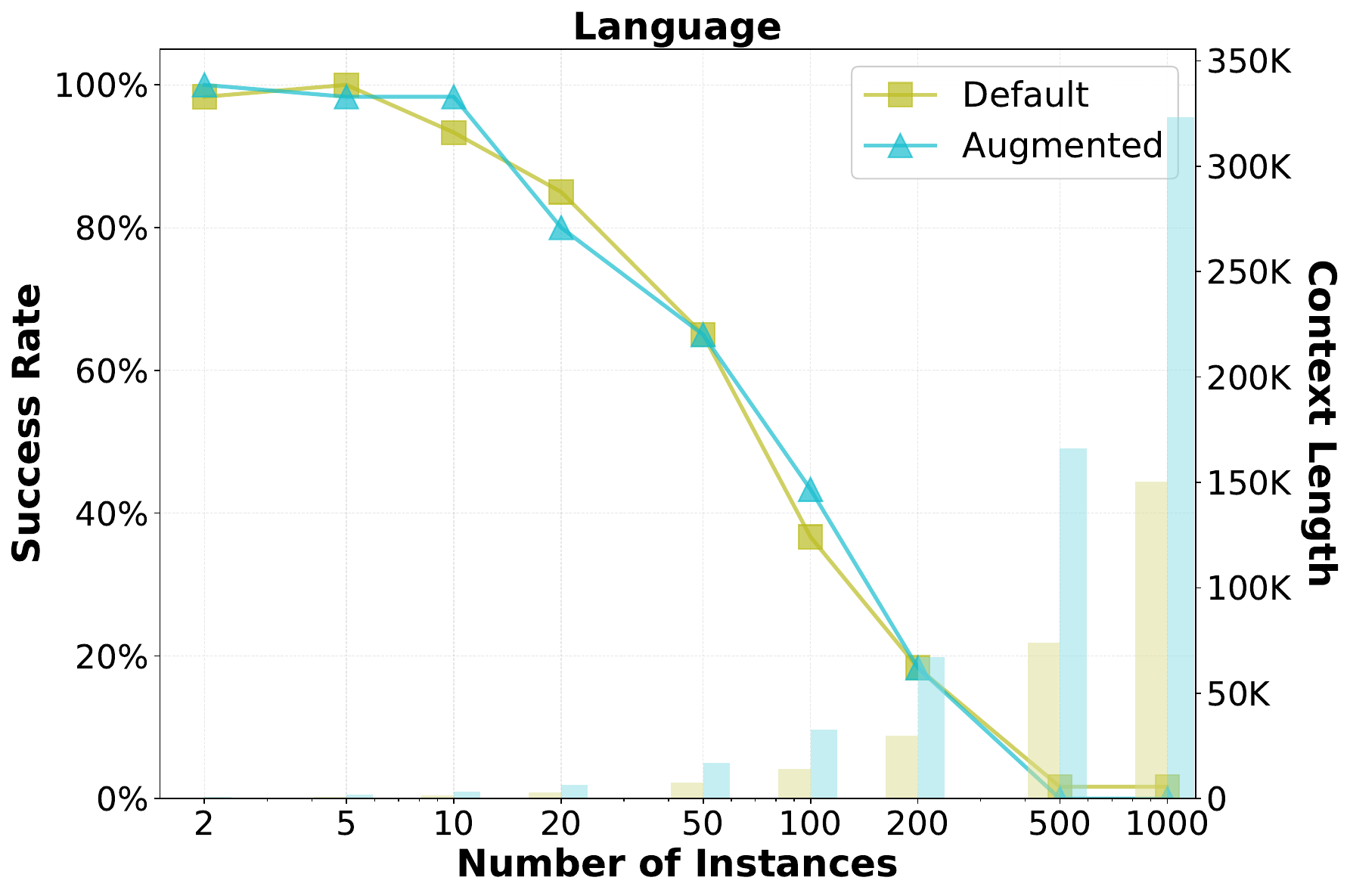}
  \includegraphics[width=0.9\linewidth]{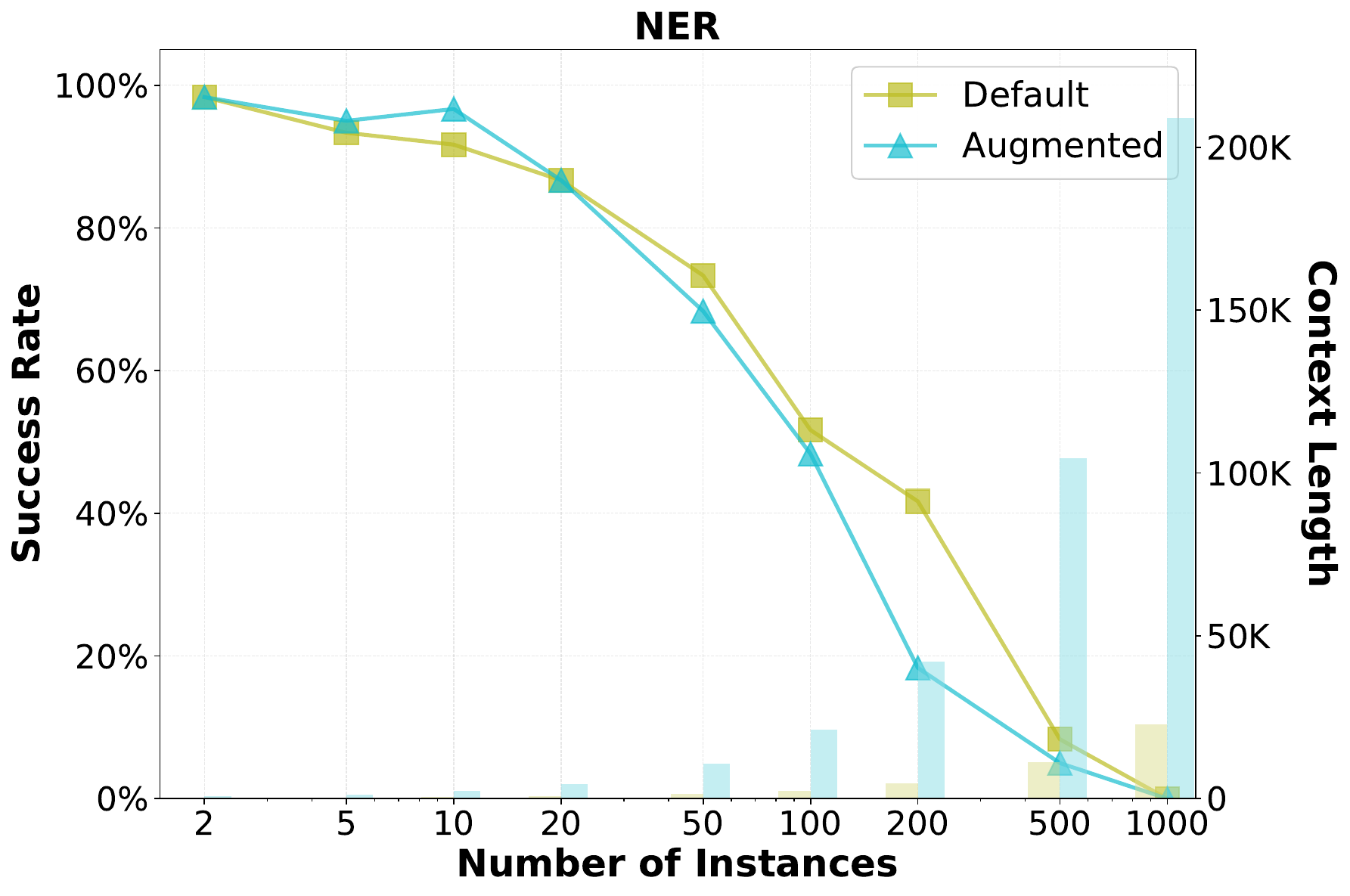}

  \caption{Success rate for tasks.}
  \label{fig:context_length_tasks_1}
\end{figure}

\begin{figure}[t]
  \centering
  \setlength{\abovecaptionskip}{2pt}
  \setlength{\belowcaptionskip}{0pt}

  \includegraphics[width=0.9\linewidth]{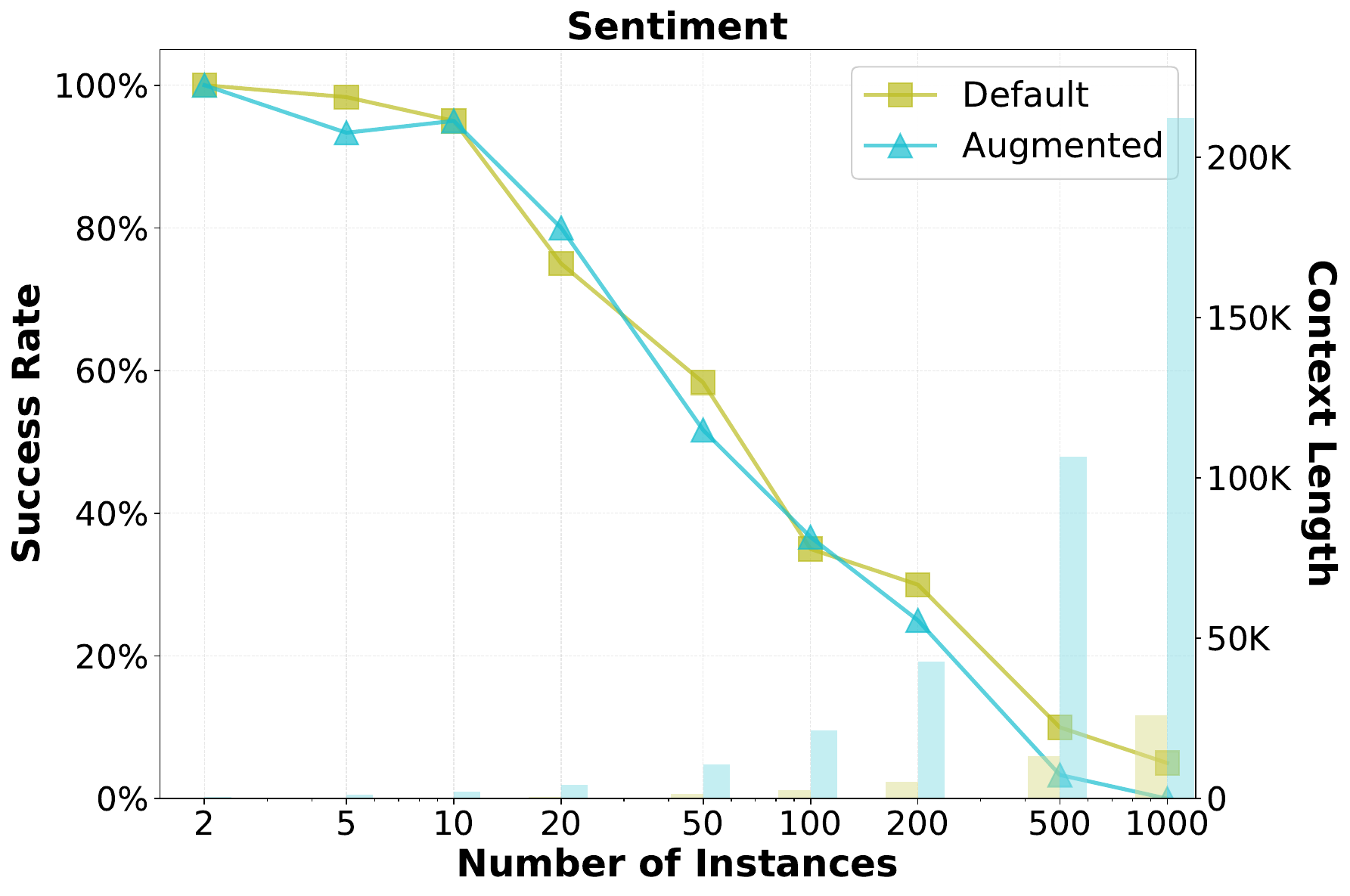}
  \includegraphics[width=0.9\linewidth]{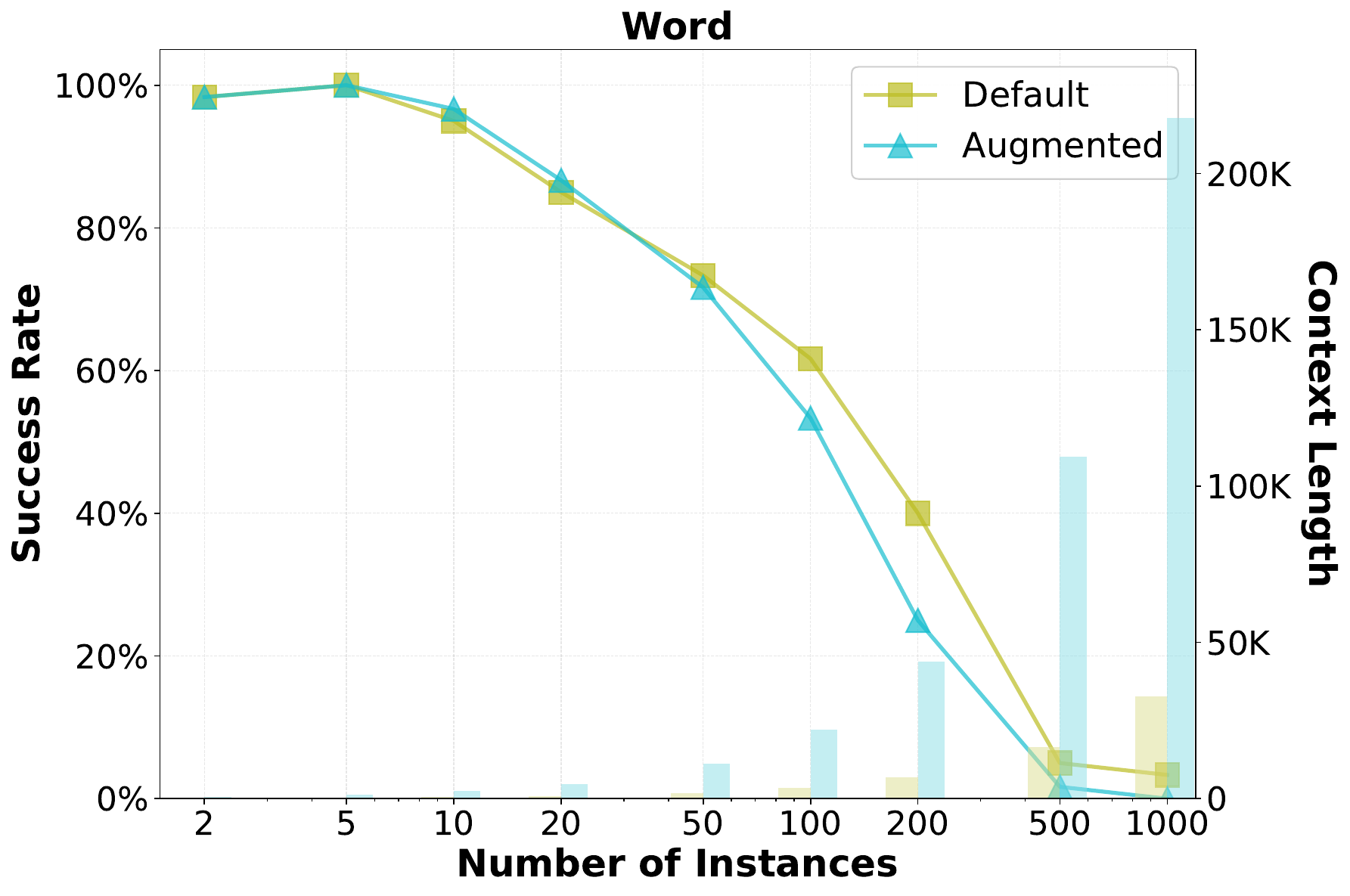}
  \includegraphics[width=0.9\linewidth]{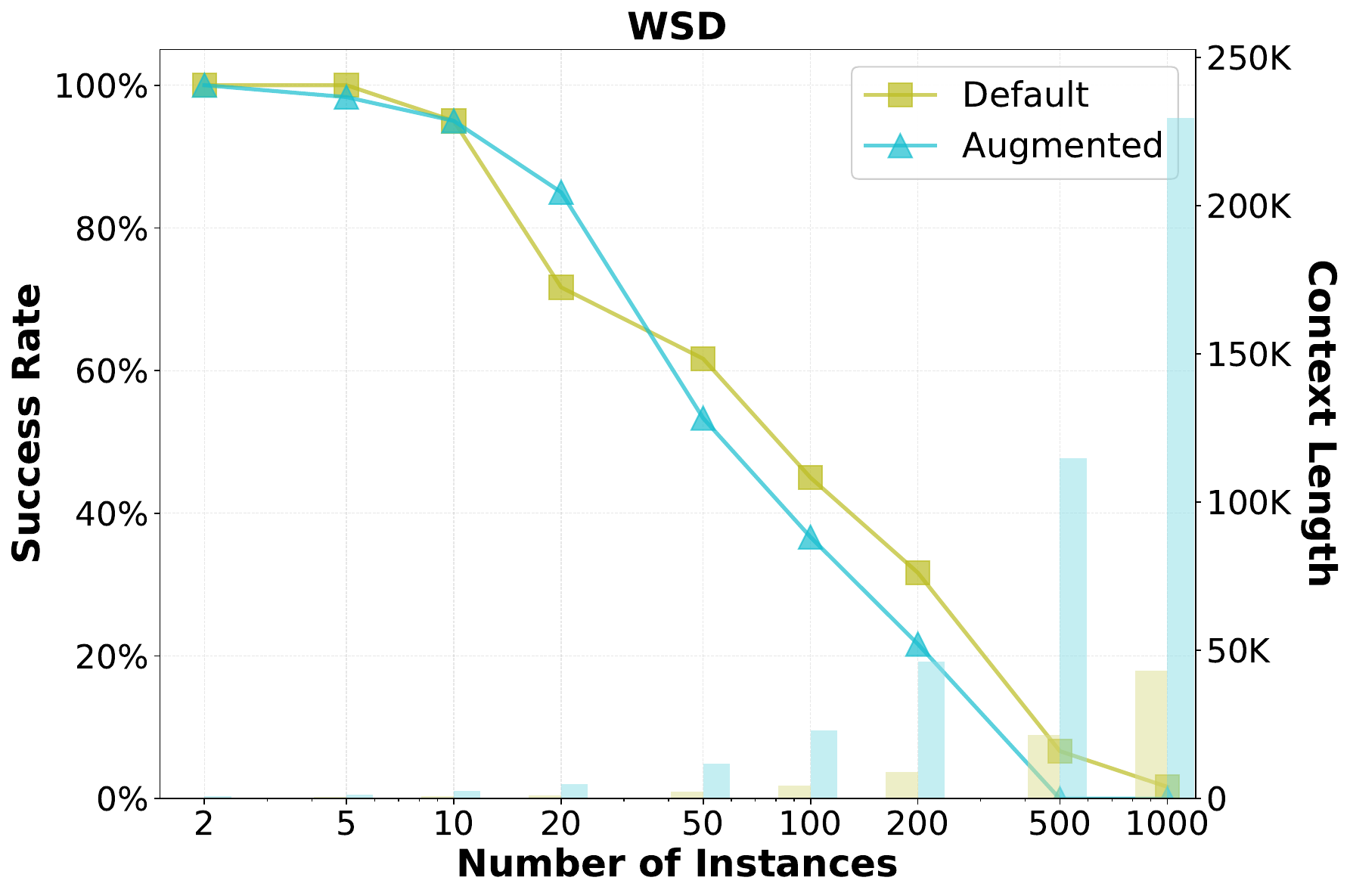}

  \caption{Success rate for tasks.}
  \label{fig:context_length_tasks_2}
\end{figure}

\camera{\subsubsection{Default and Augmented Success Rate for Tasks}
\label{appendix:context_length_tasks}
Figure~\ref{fig:context_length_tasks_1} and Figure~\ref{fig:context_length_tasks_2} show model success rate for each task.}

\end{document}